\documentclass{article}

\usepackage{iclr2026_conference,times}
\iclrfinalcopy

\usepackage[utf8]{inputenc}
\usepackage[T1]{fontenc}
\usepackage{CJKutf8}  
\usepackage{hyperref}
\hypersetup{
  colorlinks=true,
  linkcolor=[HTML]{2166ac},   
  citecolor=[HTML]{2166ac},   
  urlcolor=[HTML]{2166ac},    
}
\usepackage{url}
\usepackage{booktabs}
\usepackage{tabularx}
\usepackage{etoc}
\usepackage{amsfonts}
\usepackage{amsmath}
\usepackage{amssymb}
\usepackage{empheq}
\usepackage{nicefrac}
\usepackage{microtype}
\usepackage{xcolor}
\usepackage{graphicx}
\usepackage{tikz}
\usetikzlibrary{arrows.meta,positioning,calc}
\usepackage{subcaption}
\usepackage{enumitem}
\usepackage{wrapfig}
\usepackage{float}
\usepackage{listings}
\usepackage{fvextra}
\usepackage{multirow}

\lstdefinestyle{json}{
  basicstyle=\ttfamily\scriptsize,
  breaklines=true,
  frame=single,
  backgroundcolor=\color{gray!5},
  columns=fullflexible,
  keepspaces=true,
}

\newcommand{\ours}{Ours}

\newcommand{\zilongEmail}{\href{mailto:jaysonabcchen@gmail.com}{jaysonabcchen@gmail.com}}
\newcommand{\codeLink}{\href{https://github.com/heheyas/context-scaling}{\ttfamily https://github.com/heheyas/context-scaling}}
\newcommand{\websiteLink}{\href{https://heheyas.github.io/context-scaling}{\ttfamily https://heheyas.github.io/context-scaling}}
\newcommand{\modelLink}{\href{https://huggingface.co/collections/heheyas/context-scaling}{\ttfamily https://huggingface.co/collections/heheyas/context-scaling}}
\newcommand{\demoLink}{\href{https://heheyas-context-scaling.hf.space/}{\ttfamily https://heheyas-context-scaling.hf.space/}}

\title{Scaling Properties of Text Conditioning\\in Visual Generation}

\author{Zilong Chen, Chaorui Deng, Kunchang Li, Hongyi Yuan, Haoqi Fan\\
\mdseries ByteDance Seed\\
{\ttfamily \zilongEmail}\\
\textbf{Code:} \codeLink\\
\textbf{Models:} \modelLink\\
\textbf{Demo:} \demoLink\\
\textbf{Project page:} \websiteLink}

\begin{document}

\etocdepthtag.toc{mainbody}

\maketitle
\vspace{-0.22in}

\begin{figure}[H]
\vspace{-4pt}
\begin{center}
  \includegraphics[width=\linewidth]{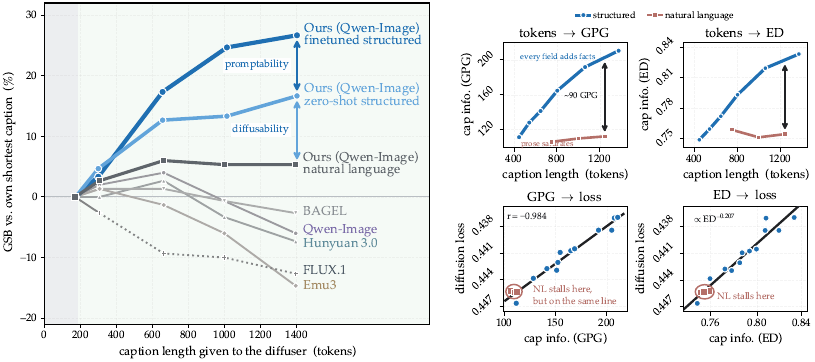}
\end{center}
\vspace{-6pt}
\caption{\small
\textbf{Information, not token count, is what scales in prompt enhancement for image generation.}
Naively increasing prompt length degrades performance across all evaluated open-weight models (Qwen-Image, HunyuanImage 3.0, BAGEL, FLUX.1 Dev, and Emu3), every one of them ending below its own shortest caption; a control of ours \emph{trained} on that same prose ladder improves only slightly before saturating. Under our structured-prompt schema, however, the GSB net preference rises monotonically with caption length, because structured prompts add new image-grounded information rather than more words. Finetuning the \emph{prompter} that writes them (\emph{finetuned structured}) yields a further large margin over the \emph{zero-shot structured} prompter.
\emph{(Left)} Each system is judged against its \emph{own} shortest-caption output, so the axis measures gain from lengthening rather than absolute quality; GSB is a VLM good/same/bad net preference over $150$ prompts, with every prompt enhancer off. FLUX.1 Dev's text encoder truncates past $512$ tokens, so its curve is dotted beyond that rung.
\emph{(Right)} Why: prose saturates in information while structure keeps gaining \emph{(top)}, yet at matched information both land on one fit \emph{(bottom)}.}
\label{fig:overview}
\vspace{-6pt}
\end{figure}

\begin{figure}[!tp]
\vspace{-15pt}
\begin{center}
  \includegraphics[width=\linewidth]{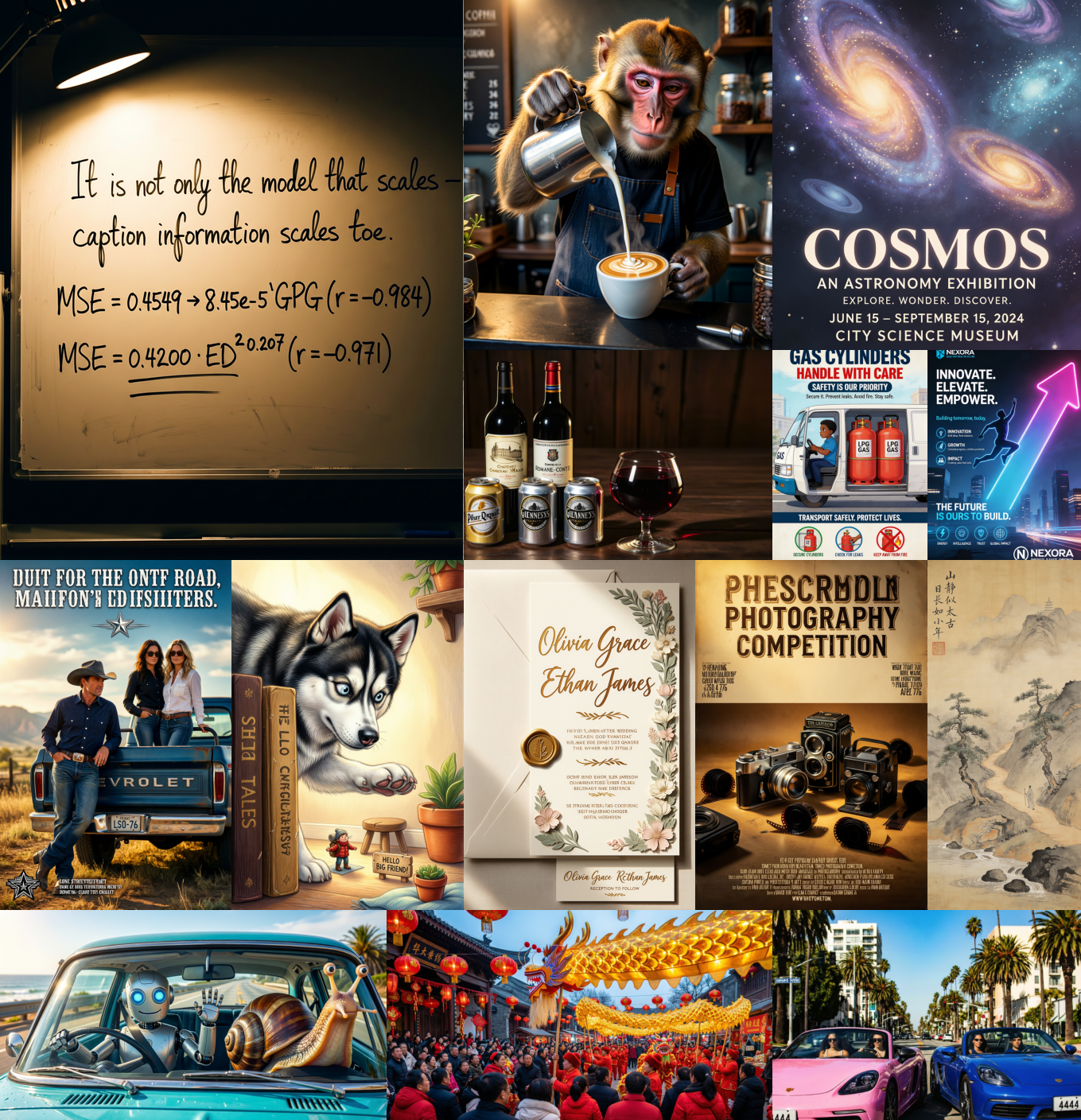}\\[0.5pt]
  \includegraphics[width=\linewidth]{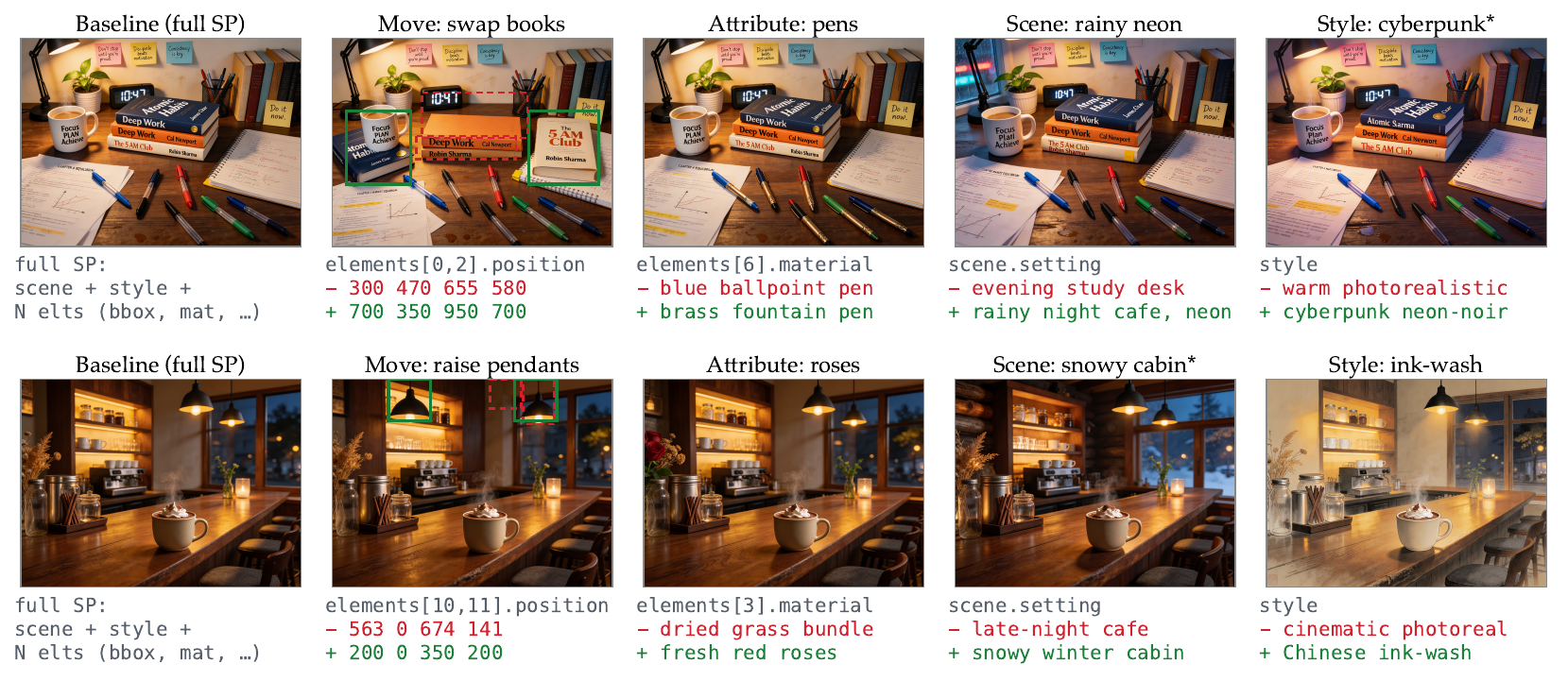}
\end{center}
\vspace{-7pt}
\caption{\small
\textbf{Qualitative results and zero-shot editing.}
\emph{(Top)} Outputs from our final system across multi-object scenes, dense text, complex layouts, and branded illustration. The trained LLM \emph{prompter} expands each user request into an information-dense structured prompt (SP), which the SP-trained \emph{diffuser} renders. Appendix~\ref{app:gallery_prompts} lists the corresponding prompts.
	\emph{(Bottom)} Because the SP exposes image factors as editable fields, a targeted field edit and regeneration can change a specific aspect---object position, material, scene, or global style---while preserving much of the remaining composition; diffs show \textcolor[HTML]{1a7f37}{added}\,/\,\textcolor[HTML]{cf222e}{removed} values, and \emph{Move} updates the relevant old/new bboxes.}
\label{fig:gallery}
\end{figure}

\begin{figure}[!t]
\vspace{-10pt}
\begin{center}
  \includegraphics[width=0.97\linewidth]{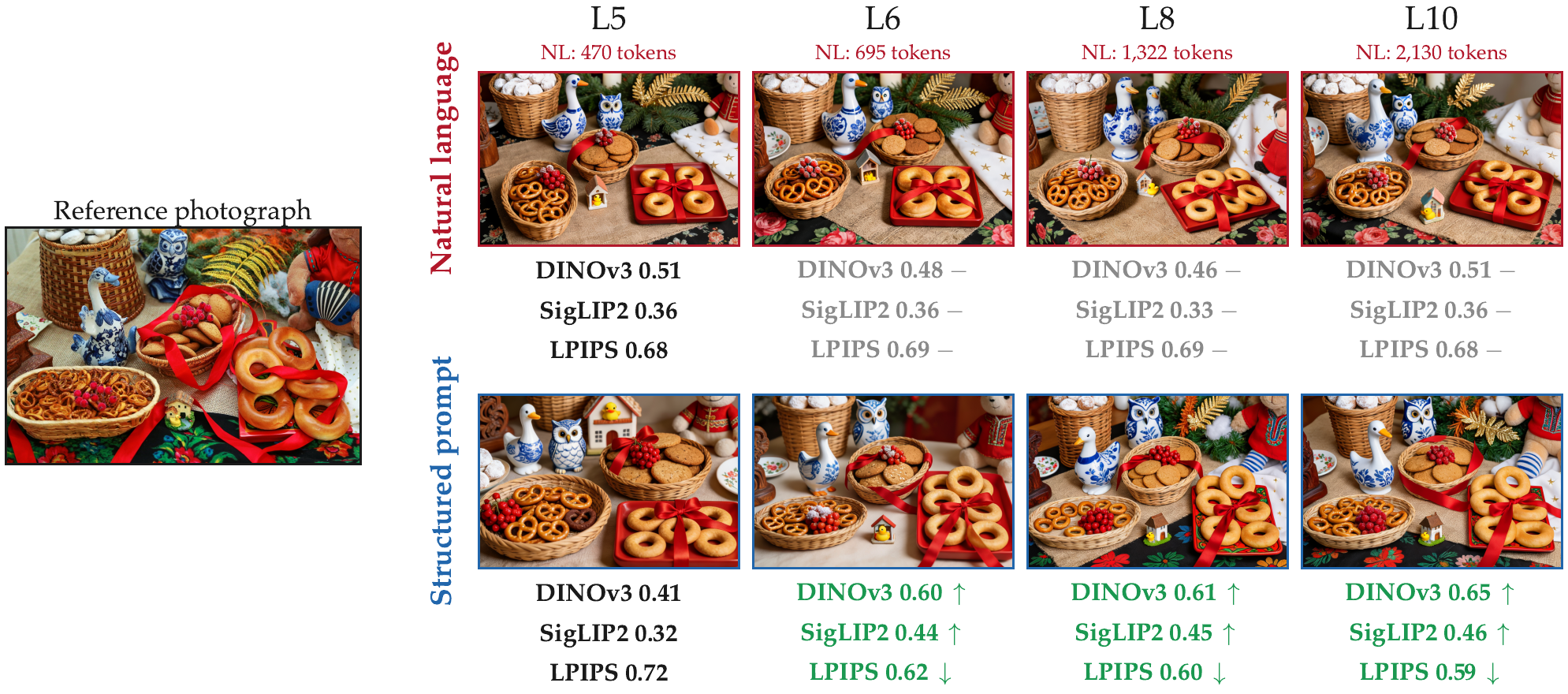}
\end{center}
\vspace{-7pt}
\caption{\small
\textbf{A fixed-backbone probe separates caption information from caption length.}
Using the same Qwen-Image backbone and seed, we reconstruct a held-out reference (\emph{left}) from natural-language (\textcolor[HTML]{b2182b}{NL}, top) and structured-prompt (\textcolor[HTML]{2166ac}{SP}, bottom) captions at four levels.
The NL captions preserve the same entities and relationships while increasing in length, yet reconstruction remains flat; progressively restoring SP fields improves all three metrics.
The displayed NL lengths are measured with the Qwen2.5-VL tokenizer.
Metrics are DINOv3/SigLIP2 cosine ($\uparrow$) and LPIPS distance ($\downarrow$); colors mark changes from L5.
Appendices~\ref{app:impl_annotation} and~\ref{app:prompts_teaser} provide the full controls and prompts.}
\label{fig:teaser}
\vspace{-5pt}
\end{figure}

\begin{abstract}
We study empirical scaling properties for text conditioning in visual generation. Such properties have rarely been measured because diffusion loss does not scale with the number of tokens in natural-language prompts. Surprisingly, we find that the converged diffusion loss scales with the amount of structured language in the prompt. To quantify structured language, we adapt two complementary measures: a white-box likelihood metric (GPG) and a black-box attribute metric (ED). Across controlled training runs, the converged diffusion loss decreases approximately linearly with GPG and follows a power law with ED. Guided by these scaling properties, we improve \emph{diffusability} by constructing structured prompts with semantic and geometric annotations derived from images, and improve \emph{promptability} by training a prompter through supervised fine-tuning, cold-start, and verifier-gated on-policy distillation. The resulting system outperforms all evaluated open-weight models on nearly every compositional, reasoning, and world-knowledge benchmark, while matching or surpassing the strongest closed-weight models on most evaluations.
\end{abstract}

\section{Introduction}
\label{sec:introduction}

Large language models have advanced through scaling model size, training data, and compute~\citep{kaplan2020,chinchilla}. Text-to-image generation has followed the same recipe, with larger diffusion backbones, heavier training runs~\citep{sdxl,esser2024sd3}, and larger captioned-image corpora~\citep{laion}. Yet this analogy hides a basic asymmetry. A language model receives its training signal from the text stream itself, whereas a text-to-image model learns text-conditioned generation through image--caption pairs. Visual content that a caption binds ambiguously reaches the model only weakly as conditioning supervision, and content the caption omits does not reach it at all, though both remain present in the pixels. The image-grounded information in a caption may therefore limit what a generator can learn to recover from text, but it has rarely been treated as an explicit training variable. We ask whether increasing this information can improve visual generation, particularly on information-dense prompts where current systems struggle with objects, layouts, relations, and visual coherence~\citep{jiao2025detailmaster,longt2ibench,tiif}.
Figure~\ref{fig:overview} previews the answer. Given progressively longer natural-language captions, every existing system we evaluate peaks early if at all and ends below its own shortest caption, and a diffuser of ours trained on that same ladder improves only slightly before saturating: prose does not scale, whether a system is merely prompted with it or trained on it. Quality keeps rising only when the added tokens carry more image-grounded information, which structured captions supply and which also predicts converged diffusion loss.

We begin with the fixed-backbone reconstruction probe shown in Figure~\ref{fig:teaser}. Starting from one reference image, we annotate it in detail, then verbalize that annotation as natural-language (NL) captions of increasing length and condition the same trained diffusion model on each to reconstruct the reference. Because they share one source annotation, these NL captions grow longer by elaborating the same annotated entities and relationships rather than adding new ones, and reconstruction quickly saturates. This suggests that verbosity alone does not improve what the diffuser can recover, motivating a representation that makes visual variables explicit. We therefore introduce \emph{structured prompts} (SPs), a typed semantic representation serialized as JSON. Dedicated fields for global scene context, per-element properties and geometry, and cross-element relations expose the same annotation in a precise, consistently addressable form. Reconstruction improves steadily as schema coverage increases. To quantify caption information beyond this illustrative probe, we adapt two complementary metrics from prior work: \emph{Grounded Perplexity Gain} (GPG), which measures how much revealing the paired image raises a caption's likelihood under a frozen vision--language model~\citep{m3id}, and \emph{Effective Detailness} (ED), which measures the precision and recall of caption attributes against image-grounded references~\citep{caption_detailness2025,liang2024precision}.

Using GPG and ED as complementary measures, we systematically study how caption informativeness influences diffusion training, with converged diffusion loss as the training-side readout. We sweep NL and SP caption formats and detail levels, running a separate diffusion training run for each condition while holding the image data, architecture, initialization, and compute fixed. Across this sweep, converged diffusion loss is well fit by a linear function of GPG and follows a power-law trend in ED. We call these two relations the \emph{scaling properties of text conditioning} under the controlled recipe. Once calibrated, they can rank candidate caption conditions within the tested range before another diffusion training run. They therefore isolate a format-side capability, which we call \emph{diffusability}: how effectively a caption representation exposes and organizes image-grounded information for diffusion learning. To raise it at scale, we first use an image-to-SP annotation pipeline that combines a VLM with frozen domain experts for human pose, depth, and segmentation to generate full-schema SPs, and then train the diffuser on the resulting organized supervision.

At test time, however, no paired image or oracle annotation is available to fill in an SP, so an LLM prompter must infer plausible visual details left unspecified by the user while preserving the explicit request. We call this capability \emph{promptability}. In a zero-shot sweep, generated-image quality improves with prompter scale and, except at the smallest scale, with chain-of-thought inference. This associates progress in general-purpose LLMs with better generation through the caption interface, without task-specific training. We further raise promptability through supervised fine-tuning (SFT), cold-start distillation, and verifier-gated reinforcement fine-tuning (RFT); the final stage distills an image-conditioned teacher on verified on-policy rollouts. At inference, an agentic refine--render--judge loop further improves generation by revising the SP fields responsible for failed visual decisions.

Together, the two factors yield an end-to-end system that leads every evaluated open-weight model on all but one reported metric and matches or surpasses the strongest closed systems on most, with the widest margins on the composition- and reasoning-heavy benchmarks. These gains extend beyond the training-loss relation to prompt fidelity, visual coherence, and compositional detail in generated images.

In summary, we show how to scale text conditioning in visual generation. Our contributions are:
\begin{itemize}[leftmargin=1.25em, nosep]
    \item \textbf{Scaling properties of text conditioning.} We adapt GPG and ED to quantify caption informativeness and show that both predict converged diffusion loss under a fixed training recipe.
    \item \textbf{Raising diffusability through structured prompts.} SPs organize image-grounded content into named fields, raising measured informativeness and lowering diffusion loss without architectural changes.
    \item \textbf{Raising promptability through training and inference-time refinement.} Our SFT--cold-start--RFT pipeline culminates in verifier-gated on-policy self-distillation (OPSD), while field-level agentic refinement further improves generation at inference time.
    \item \textbf{Scaling text conditioning end to end.} Combining the two factors yields broad gains across compositional, reasoning, and world-knowledge evaluations.
\end{itemize}

\section{Related Work}
\label{sec:related_work}

\paragraph{Text-to-image scaling, long captions, and structured control.}
Diffusion models~\citep{ho2020ddpm,song2021scorebased}, latent diffusion~\citep{rombach2022ldm}, and transformer backbones~\citep{peebles2023dit,esser2024sd3} have advanced T2I generation largely through model-side scaling. Conditioning has also mattered: Imagen~\citep{saharia2022imagen} found that a larger text encoder improves fidelity, while LLM conditioners~\citep{hu2024ella}, long-context encoders~\citep{zhang2024longclip}, and dense-caption corpora~\citep{urbanek2024dci,onoe2024docci} extend the information available in a prompt. Existing scaling laws establish a compute--loss relation for DiTs~\citep{mei2024scaling}; we instead hold model, data, and compute fixed and show that converged diffusion loss is predicted by caption information, measured by white-box GPG or black-box ED, rather than length. This distinction is consistent with long-prompt benchmarks: DetailMaster~\citep{jiao2025detailmaster} bins prompts by token count and reports a consistent negative correlation between prompt length and accuracy on character attributes, character locations, and entity relationships, while LongT2IBench~\citep{longt2ibench} reports that graph-structured alignment over entities, attributes, and relations decreases steadily across word-count intervals. TIIF-Bench~\citep{tiif} instead pairs each prompt with a semantically equivalent long version and finds that robustness to this change tracks overall instruction-following ability, with the strongest models remaining stable across both settings. Our measurements suggest that this behavior reflects an information plateau rather than length alone.
A separate line of work intervenes on the diffuser rather than on the caption. Layout-conditioned generation (e.g., GLIGEN~\citep{gligen}) and attention-manipulation methods (e.g., Attend-and-Excite~\citep{chefer2023attendexcite}) steer a fixed diffuser through auxiliary spatial conditions or model-internal intervention; our schema instead places boxes, depth, and relations in ordinary text fields, scored on the same GPG/ED axes as free-form captions and aimed at caption information rather than layout control.

\paragraph{Recaptioning and caption quality.}
DALL-E~3~\citep{betker2023dalle3} showed that detailed synthetic recaptioning can markedly improve prompt following; PixArt-$\alpha$~\citep{chen2023pixartalpha}, PixArt-$\Sigma$~\citep{chen2024pixartsigma}, CogView3~\citep{zheng2024cogview3}, and RECAP~\citep{segalis2023recap} likewise demonstrate the value of richer or more principled captions. Recent systems go beyond free-form prose. FIBO~\citep{gutflaish2025fibo} trains on long JSON captions, compares them with short captions under matched training, and learns a VLM translator from short requests to its schema. Cosmos~3~\citep{nvidia2026cosmos3} uses structured JSON annotations and a prompt upsampler, showing that predefined fields improve annotation recall over dense prose. Reve 2.0~\citep{reve2026layoutbet} uses a hierarchical layout intermediary and reports gains over text-only generation as well as improved reconstruction with more regions; concurrent work by \citet{merchant2025structuredcaptions} studies a fixed four-part caption template. These results establish structured representation as a useful design choice. Our focus is complementary: we measure image-grounded information across NL lengths, nested SP levels, spatial serializations, and field ablations, then calibrate that common variable against matched-budget converged diffusion loss. We further isolate prompt production by varying prompter scale, reasoning, and training while holding the schema and trained diffuser fixed. GPG adapts the per-token grounding signal of \citet{m3id} from decoding-time hallucination localization to corpus-level caption informativeness; ED adapts caption-detailness evaluation~\citep{caption_detailness2025} to black-box, image-grounded caption scoring; our precision-weighted matcher is motivated by \citet{liang2024precision}, who find caption precision to matter more than recall when training text-to-image models.

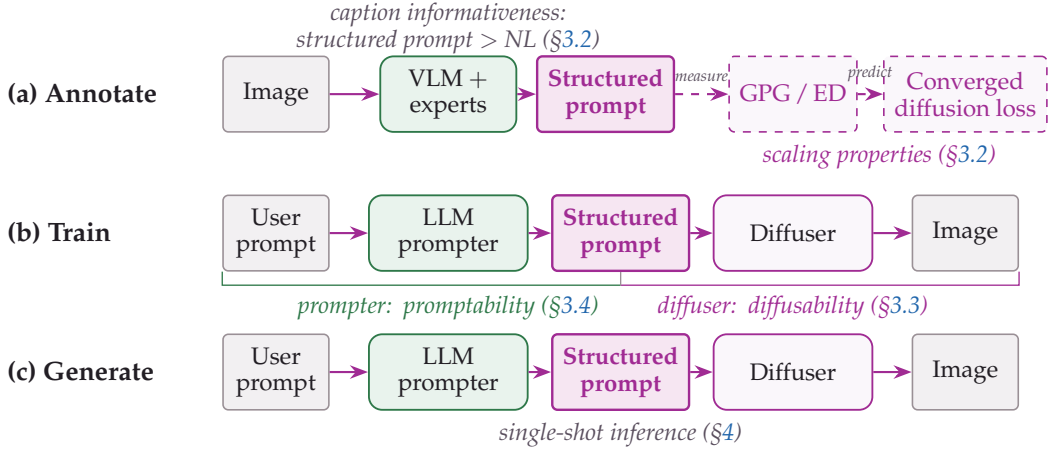
\begin{figure*}[!t]
\centering
\definecolor{cvMag}{HTML}{A02B93}
\definecolor{cvMagFill}{HTML}{F6E6F2}
\definecolor{cvMagTint}{HTML}{FAF7FB}
\definecolor{cvGrn}{HTML}{2E7849}
\definecolor{cvGrnFill}{HTML}{E9F3ED}
\definecolor{cvDark}{HTML}{221E24}
\definecolor{cvMuted}{HTML}{5A4F5E}
\definecolor{cvGray}{HTML}{9A9096}
\definecolor{cvGrayFill}{HTML}{F2F0F2}
{\fontfamily{ppl}\selectfont
\begin{tikzpicture}[
  x=0.80cm, y=0.92cm,
  font=\small,
  >={Stealth[length=2.4mm]},
  every node/.style={align=center},
  data/.style={rounded corners=2pt, draw=cvGray, line width=0.6pt, fill=cvGrayFill, text=cvDark, inner sep=4pt, minimum height=10mm, minimum width=14mm},
  sp/.style={rounded corners=2pt, draw=cvMag, line width=0.9pt, fill=cvMagFill, text=cvMag, font=\small\bfseries, inner sep=4pt, minimum height=10mm, minimum width=15mm},
  model/.style={rounded corners=5pt, draw=cvGrn, line width=0.7pt, fill=cvGrnFill, text=cvDark, inner sep=4pt, minimum height=10mm, minimum width=21mm},
  diff/.style={rounded corners=5pt, draw=cvMag, line width=0.7pt, fill=cvMagTint, text=cvDark, inner sep=4pt, minimum height=10mm, minimum width=21mm},
  lab/.style={font=\bfseries, text=cvDark, anchor=west},
  flow/.style={->, thick, cvMag},
  note/.style={font=\footnotesize\itshape, text=cvMuted},
  meas/.style={rounded corners=2pt, draw=cvMag, dashed, line width=0.6pt, fill=cvMagTint, text=cvMag, inner sep=4pt, minimum height=10mm, minimum width=15mm},
  mflow/.style={->, thick, cvMag, dashed},
]
\def\ca{0.0}\def\cb{2.85}\def\cc{5.7}\def\cd{8.55}\def\ce{11.4}
\def\rA{4.0}\def\rB{2.0}\def\rC{0.0}
\node[lab] at (-4.6,\rA) {(a) Annotate};
\node[data]  (imgA) at (\ca,\rA) {Image};
\node[model, minimum width=18mm] (vlmA) at (\cb,\rA) {VLM $+$\\experts};
\node[sp]    (spA)  at (5.45,\rA) {Structured\\prompt};
\draw[flow] (imgA)--(vlmA);\draw[flow] (vlmA)--(spA);
\node[note] at (\cb,\rA+0.95) {caption informativeness:\\ structured prompt $>$ NL~(\S\ref{sec:metrics})};
\node[meas] (gpgA) at (\cd,\rA) {GPG\,/\,ED};
\draw[mflow] (spA)--(gpgA) node[midway, above=1pt, font=\tiny\itshape, text=cvMuted] {measure};
\node[meas] (lossA) at (\ce,\rA) {Converged\\diffusion loss};
\draw[mflow] (gpgA)--(lossA) node[midway, above=1pt, font=\tiny\itshape, text=cvMuted] {predict};
\node[note, text=cvMag] at (9.98,\rA-0.9) {scaling properties~(\S\ref{sec:metrics})};
\node[lab] at (-4.6,\rB) {(b) Train};
\node[data]  (prB) at (\ca,\rB) {User\\prompt};
\node[model] (pmB) at (\cb,\rB) {LLM\\prompter};
\node[sp]    (spB) at (\cc,\rB) {Structured\\prompt};
\node[diff]  (dfB) at (\cd,\rB) {Diffuser};
\node[data]  (imB) at (\ce,\rB) {Image};
\draw[flow] (prB)--(pmB);\draw[flow] (pmB)--(spB);\draw[flow] (spB)--(dfB);\draw[flow] (dfB)--(imB);
\def\by{1.25}
\draw[cvGrn, line width=0.5pt]  (prB.south west) -- (prB.west|-0,\by) -- (spB.south|-0,\by);
\draw[cvMag, line width=0.5pt] (spB.south|-0,\by) -- (imB.east|-0,\by) -- (imB.south east);
\draw[cvGray, line width=0.5pt] (spB.south) -- (spB.south|-0,\by); 
\node[note, text=cvGrn]  at (\cb,0.92) {prompter:\, promptability~(\S\ref{sec:prompter})};
\node[note, text=cvMag] at (\cd,0.92) {diffuser:\, diffusability~(\S\ref{sec:sp})};
\node[lab] at (-4.6,\rC) {(c) Generate};
\node[data]  (prC) at (\ca,\rC) {User\\prompt};
\node[model] (pmC) at (\cb,\rC) {LLM\\prompter};
\node[sp]    (spC) at (\cc,\rC) {Structured\\prompt};
\node[diff]  (dfC) at (\cd,\rC) {Diffuser};
\node[data]  (imC) at (\ce,\rC) {Image};
\draw[flow] (prC)--(pmC);\draw[flow] (pmC)--(spC);\draw[flow] (spC)--(dfC);\draw[flow] (dfC)--(imC);
\node[note] at (\cc,\rC-0.9) {single-shot inference~(\S\ref{sec:experiments})};
\end{tikzpicture}%
}
\caption{\textbf{Method overview: annotate and measure, train, generate.}
\emph{(a)}~A VLM and frozen domain experts map each training image to an SP (\S\ref{sec:sp}); across controlled caption configurations, GPG and ED predict converged diffusion loss, yielding the scaling properties of text conditioning (\S\ref{sec:metrics}).
\emph{(b)}~The diffuser learns SP $\to$ image; the prompter learns user prompt $\to$ SP, and prompter-side comparisons hold the schema and trained diffuser fixed.
\emph{(c)}~Single-shot inference composes the two with one prompter call followed by one diffuser call, user prompt $\to$ SP $\to$ image.}
\label{fig:system_overview}
\end{figure*}

\paragraph{LLM prompters, RFT, and inference-time methods.}
LLMs have become effective inference-time prompt enhancers: RPG~\citep{yang2024rpg} decomposes prompts into regional sub-prompts with multimodal-LLM reasoning; PromptEnhancer~\citep{wang2025promptenhancer} applies RFT to a chain-of-thought rewriter with a dedicated reward model; and input-side inference-time scaling~\citep{chen2025inputside} trains a rewriter with iterative DPO. These methods improve prompts for a fixed generator. Our structured format additionally raises the caption representation's training-time \emph{diffusability}; with that schema and backbone fixed, we study prompter \emph{promptability} under LLM scale, chain-of-thought, verifier-gated self-distillation from an image-conditioned teacher, and an agentic refine--render--judge loop. Reward models such as ImageReward~\citep{xu2024imagereward} and model-side preference optimization such as Diffusion-DPO~\citep{wallace2024diffusiondpo} optimize the generator, whereas our verifier filters prompter rollouts while RFT holds the already SP-trained diffuser fixed. Finally, CLIPScore~\citep{clipscore} and VQAScore~\citep{lin2024genai} measure image--text alignment; GPG measures caption informativeness relative to its paired image, while ED uses a one-time image-grounded annotation pass followed by text-only scoring.

\section{Method}
\label{sec:method}

Our goal is to understand and exploit the image-grounded supervision that captions provide for visual generation.
We first test the standard caption-side intervention of lengthening free-form natural-language (NL) captions and find that it does not reliably increase useful supervision (\S\ref{sec:format}).
This limitation motivates the \emph{structured prompt} (SP), a typed JSON caption that organizes image-grounded variables into named fields.
Figure~\ref{fig:system_overview} summarizes how the SP becomes the shared interface of our method.
Using GPG and ED, we quantify the information exposed by each caption and relate it to converged diffusion loss in a controlled training sweep (\S\ref{sec:metrics}).
The resulting scaling properties isolate the diffusion-side role of the caption interface; realizing that interface from a user request introduces a second, LLM-side factor.
We summarize the two as \emph{Diffusability}$\times$\emph{Promptability}: diffusability captures how effectively the caption representation exposes and organizes the supervision the diffuser learns from, while promptability captures how effectively an LLM prompter instantiates that interface from a user request.
We raise diffusability by annotating training images as SPs and incorporating them into diffusion training (\S\ref{sec:sp}), and raise promptability by scaling and training the prompter (\S\ref{sec:prompter}).

\subsection{From natural-language captions to structured prompts}
\label{sec:format}

Most modern T2I systems condition their diffusers on free-form natural-language (NL) captions, whether collected from web-scale corpora, recaptioned, or produced by LLMs~\citep{rombach2022ldm,sdxl,bfl2024flux,betker2023dalle3}.
A common caption-side intervention is to make this NL condition longer and more detailed.
The NL row of Figure~\ref{fig:teaser} tests whether this helps using a fixed-backbone reconstruction probe.
A single Qwen-Image~\citep{qwenimage} backbone trained on captions spanning different lengths and richness levels attempts to regenerate one held-out image from each of four NL descriptions of the same annotated entities and relationships.
To our surprise, reconstruction remains essentially flat as the captions become substantially longer, with the backbone and sampling seed fixed.
In this probe, the added prose elaborates existing content without improving what the diffuser recovers.
This saturation motivates testing whether an organized representation can expose image-grounded variables more effectively.
We therefore introduce the \emph{structured prompt} (SP), a typed JSON caption that assigns each represented visual variable to a named field.

\begin{figure*}[!t]
  \centering
  \includegraphics[width=\textwidth]{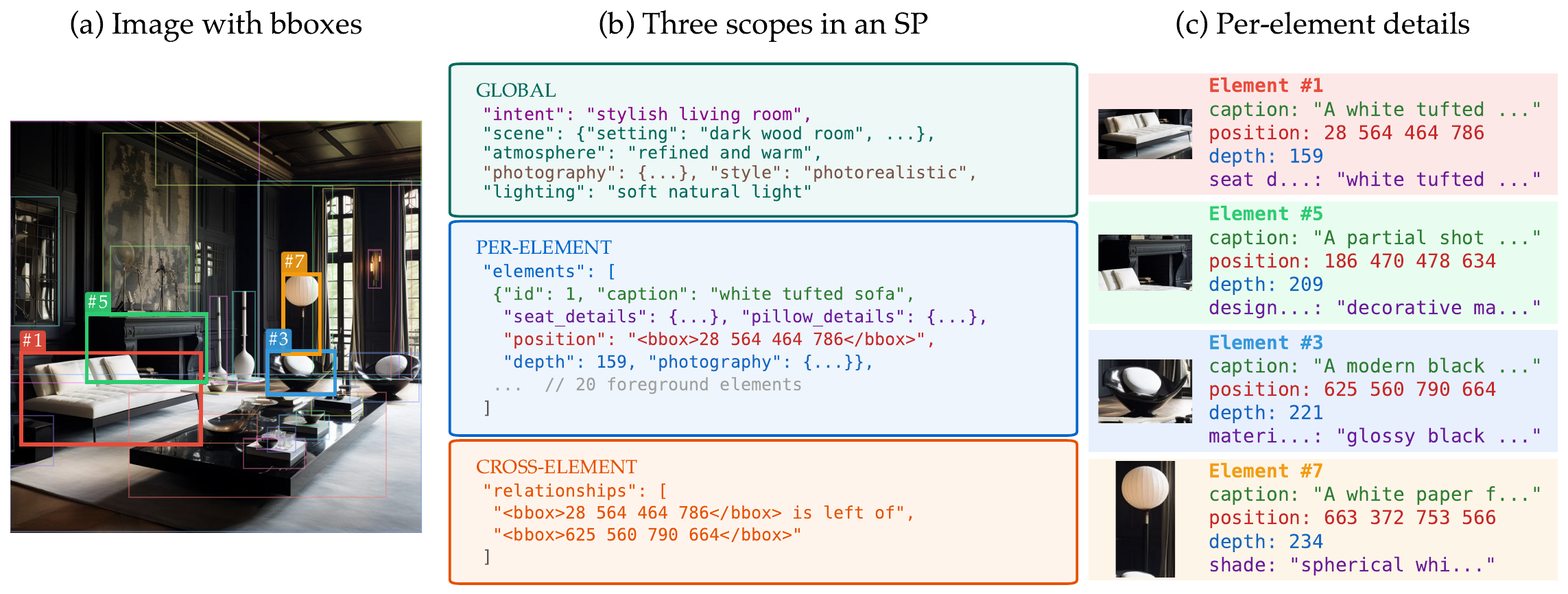}
  \caption{\textbf{Structured-prompt schema by example.}
  \emph{Left:} an image annotated with bounding boxes and element IDs.
  \emph{Center:} an abridged pseudo-JSON view grouping global, per-element, and cross-element fields; ellipses and comments are explanatory and are not part of the serialized training record, and the displayed count refers to foreground elements.
  \emph{Right:} selected per-element fields.
  The schema organizes global scene context, per-element properties and geometry, and cross-element relations into named conditioning fields; \S\ref{sec:metrics} measures how the information they carry relates to diffusion training loss.}
\label{fig:schema_example}
\end{figure*}

An SP organizes image-grounded information at three scopes, as illustrated in Figure~\ref{fig:schema_example}.
\emph{Global} fields describe the overall intent, scene, atmosphere, photography, style, and lighting.
Each foreground object receives a \emph{per-element} entry for its identity, attributes, actions, bounding-box position, optional depth, and photography, while \emph{cross-element} relationships bind these entries to one another.
The SP row of Figure~\ref{fig:teaser} shows a different trend: reconstruction improves steadily as additional field groups are included, unlike the flat NL length sweep.
The probe motivates this representation, but one image cannot establish whether caption informativeness predicts diffusion learning across training configurations.
We therefore next introduce caption-side information measures and a controlled training sweep.

\subsection{Caption informativeness predicts training loss}
\label{sec:metrics}

\begin{figure*}[!t]
  \centering
  \includegraphics[height=1.78in]{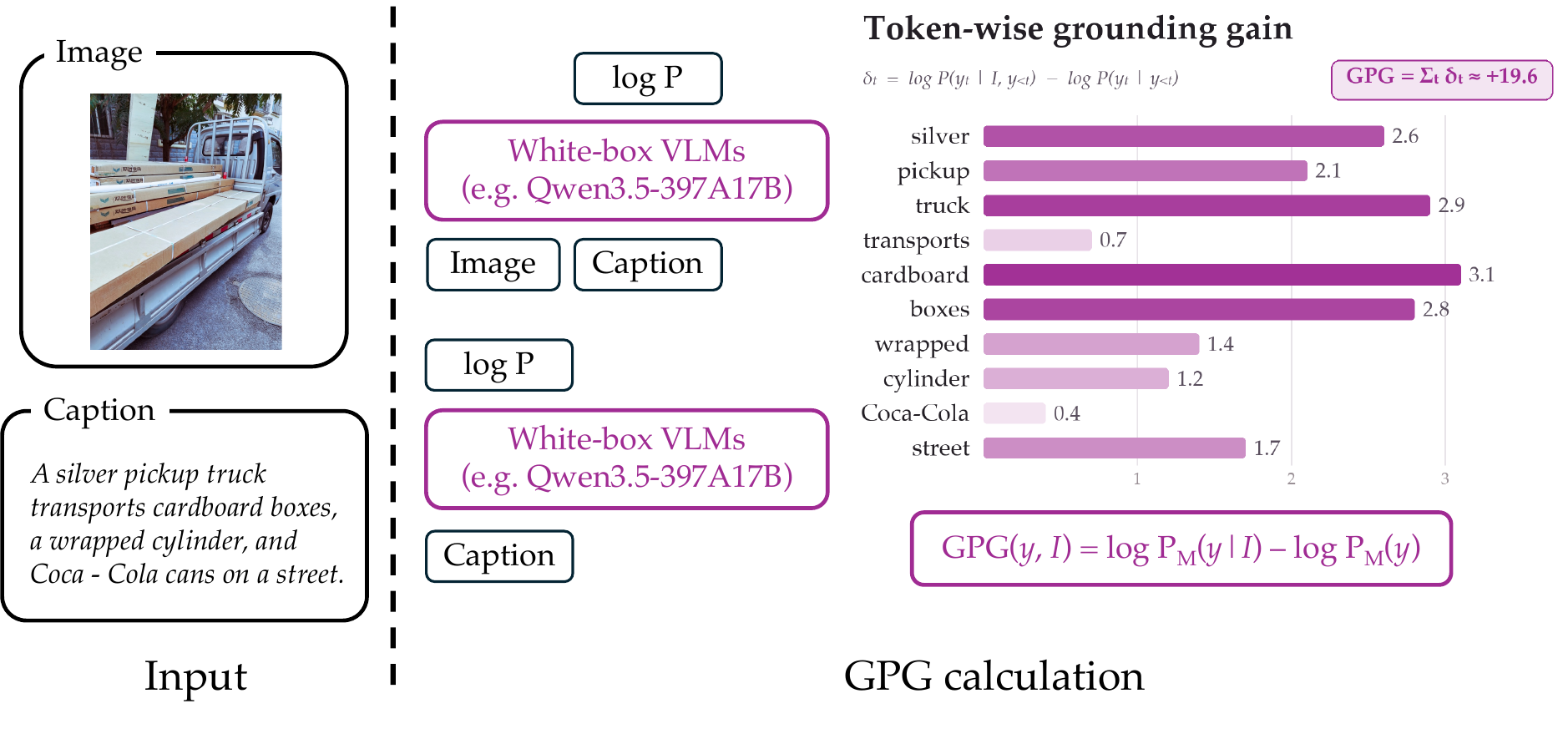}\hspace{8pt}%
  \includegraphics[height=1.78in]{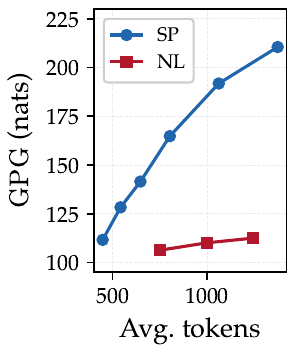}\\[6pt]
  \includegraphics[height=1.78in]{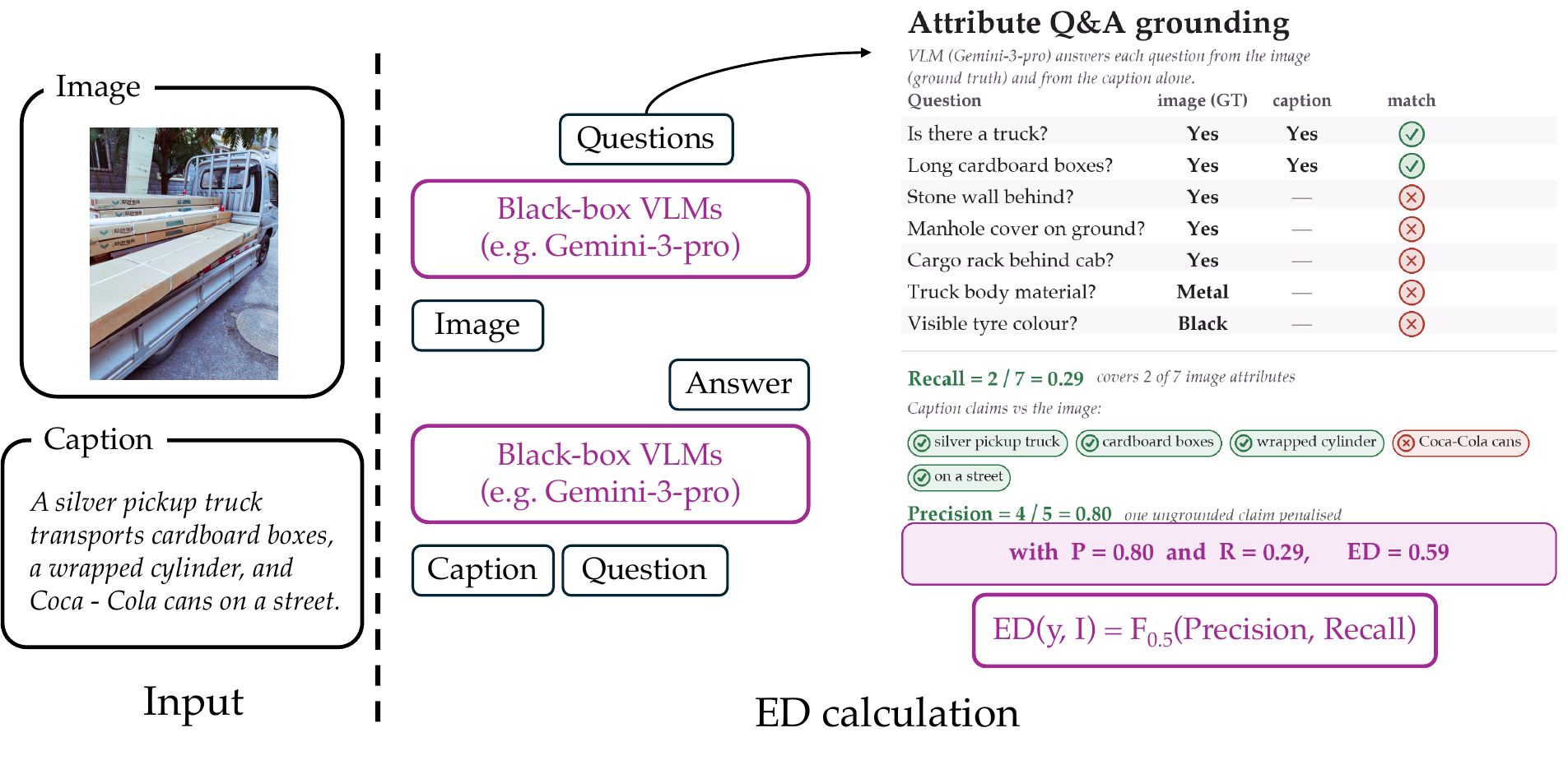}\hspace{8pt}%
  \includegraphics[height=1.78in]{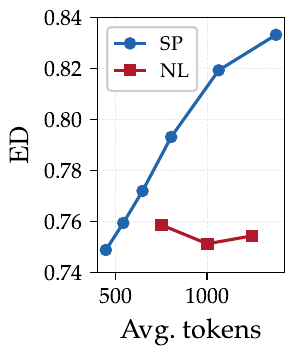}
  \caption{\textbf{The two informativeness measures: definition (left) and measured trends (right).}
  \emph{Top:} GPG sums the scored content-token log-likelihood gain from revealing the image (Eq.~\eqref{eq:gpg}).
  \emph{Bottom:} ED matches caption attributes against an image-grounded reference set and reports $F_{0.5}(P_A,R_A)$ (Eq.~\eqref{eq:ed}).
  Both remain nearly flat as NL captions grow longer, but increase across the nested SP configurations described below.}
  \label{fig:gpg_ed_demo}
\end{figure*}

We use two complementary caption-side metrics adapted from prior work---\emph{Grounded Perplexity Gain} (GPG; white-box)~\citep{m3id} and \emph{Effective Detailness} (ED; black-box)~\citep{caption_detailness2025,liang2024precision}---and study their relationship with converged diffusion training loss in a controlled scaling sweep.\footnote{Here, white-box and black-box refer to whether scoring requires access to a VLM's token log-probabilities.}

\paragraph{Grounded Perplexity Gain (GPG).}
\label{sec:gpg}
For GPG, we measure how much revealing the paired image increases a caption's likelihood under a frozen VLM.
We fix a VLM judge $M$ (here Qwen3.5-397B-A17B~\citep{qwen35report}) and apply the caption canonicalization and content-mask protocol of Appendix~\ref{app:gpg_protocol}; for image $I$, let $y_{1{:}T}$ denote the resulting sequence tokens and $m_{1{:}T}\!\in\!\{0,1\}^{T}$ the corresponding content mask.
The \emph{Grounded Perplexity Gain} of $y$ on $I$ is the log-likelihood gain when the image is revealed, summed over the scored content positions:
\begin{equation}
\mathrm{GPG}(y, I)
\;\triangleq\;
\sum_{t=1}^{T}
m_t\Bigl[\,\log p_M(y_t \mid I, y_{<t}) \;-\; \log p_M(y_t \mid \varnothing, y_{<t})\,\Bigr],
\label{eq:gpg}
\end{equation}
Here $\varnothing$ denotes the matched no-image pass.
This masked conditional log-likelihood gain is an operational estimate of caption--image mutual information under $M$; the image-free pass supplies a model-based prior rather than the exact dataset marginal.
GPG is thus a total rather than a per-token rate, so it can grow with caption length; we report its mean over the shared $30{,}000$-image evaluation pool (the same images used for ED) for each caption configuration.
We hold the judge, scoring template, image preprocessing, and paired no-image pass fixed across all caption configurations.
\begin{figure*}[t]
  \centering
  \includegraphics[width=\textwidth]{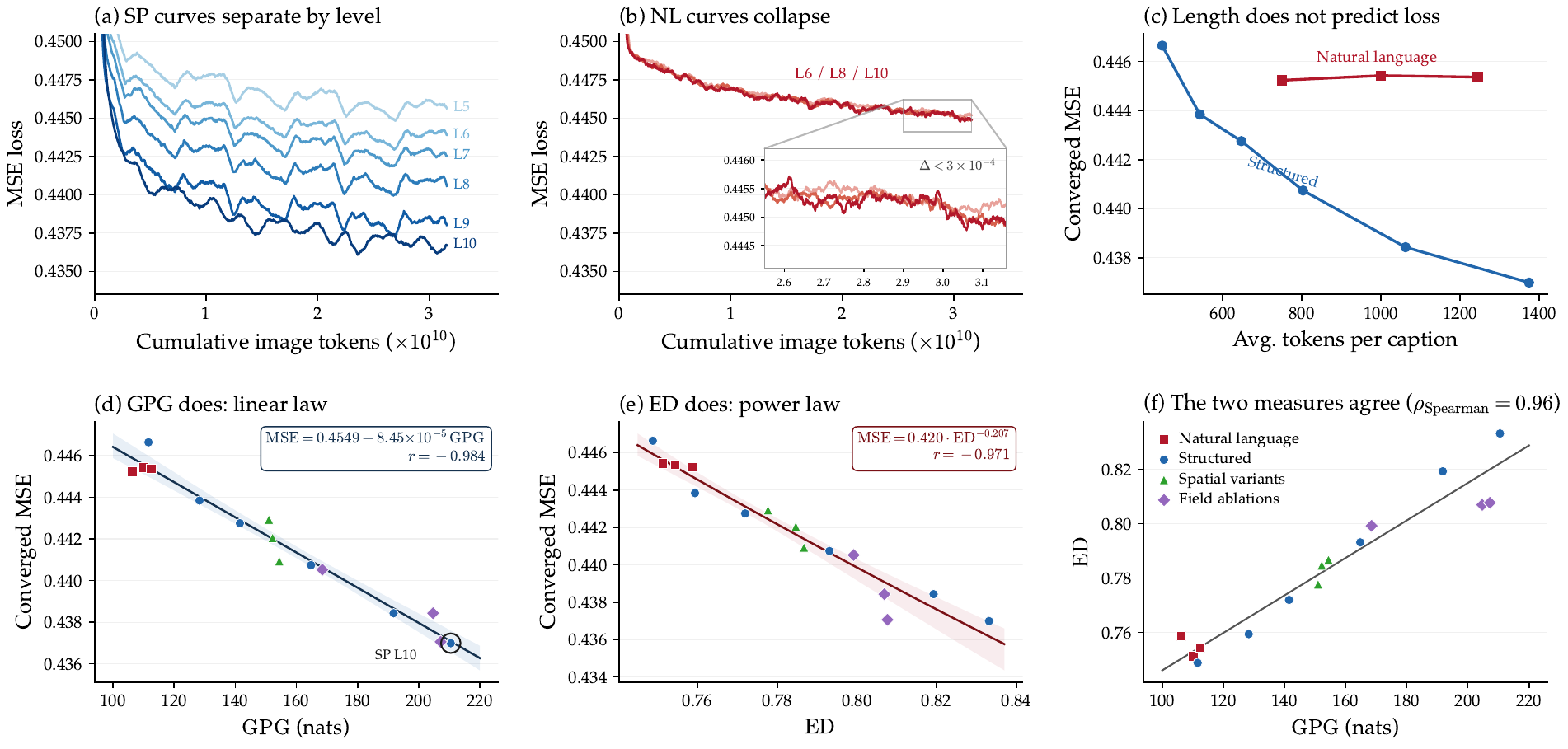}
  \caption{\textbf{The scaling properties of text conditioning: loss follows information.}
  \emph{(a--b)}~Training curves separate across SP levels but overlap across NL richness levels.
  \emph{(c)}~Caption length does not place NL and SP on a common loss trend.
  \emph{(d--e)}~Across all $15$ sweep points, converged MSE is approximately linear in GPG and follows a power law in ED; bands show the central $95\%$ range across resamples of the designed sweep settings, not training-run confidence intervals.
  \emph{(f)}~GPG and ED closely agree on the caption-configuration ranking despite using different scoring interfaces.}
  \label{fig:scaling_overview}
\end{figure*}

\paragraph{Effective Detailness (ED).}
\label{sec:ed}
ED offers a complementary semantic measure of caption information without requiring token log-probabilities.
Adapting caption-detailness evaluation~\citep{caption_detailness2025}, ED uses independent image-side and caption-side proposers to extract object--attribute--relationship--grounding tuples from pixels and text, respectively, followed by a paraphrase-tolerant matcher (Gemini 3 Pro for image proposal; separate GPT-5.4 calls for caption proposal and matching).
The matcher returns two symmetric binary masks, indicating which image-side attributes are covered by the caption and which caption-side attributes are supported by the image.
We denote the resulting caption-side attribute precision and image-side attribute recall by $P_A$ and $R_A$, respectively:
\begin{equation}
\mathrm{ED}(y, I)
\;\triangleq\;
F_{0.5}\bigl(P_A,\,R_A\bigr),
\label{eq:ed}
\end{equation}
where $F_{0.5}$ emphasizes precision over recall, penalizing unsupported caption attributes more heavily than omissions~\citep{vanrijsbergen1979,liang2024precision}.
Appendix~\ref{app:ed_beta} motivates this choice and tests its sensitivity.
Only the attribute subset enters the reported score; the broader tuple decomposition supplies entity context for matching, while object, relation, and grounding terms are not included in the reported ED value.
To avoid caption-derived leakage, the image-side proposer never sees the candidate caption; after this one-time image-side pass, caption proposal and matching are text-only and require no log-probabilities.
We aggregate ED over $30{,}000$ paired images per caption configuration using a two-sided $10\%$-trimmed mean, discarding the lowest and highest $10\%$ of pair-level scores to reduce sensitivity to occasional extraction or matching failures~\citep{hampel1974}.
Appendix~\ref{app:ed_validation} documents the extraction and matching protocol and its backend checks.

\paragraph{The scaling properties.}
Using GPG and ED, we ask whether caption informativeness predicts the converged diffusion loss reached at a common training budget.
The controlled sweep comprises $15$ caption configurations constructed from the same full image annotations.
Three NL controls verbalize the same core facts under increasing length budgets.
Six SP configurations expose progressively richer subsets of the schema; Section~\ref{sec:sp} defines the corresponding field ladder.
All six are deterministic projections of the same full-schema record, obtained by masking predefined field groups and cumulatively restoring them as detail increases.
The remaining six probe representation choices more directly: three replace bounding boxes with locations on $3{\times}3$, $5{\times}5$, or $9{\times}9$ grids, and three mask the scene, bounding-box, or relationship fields from the full-schema SP.
Appendix~\ref{app:all_cells} reports the complete sweep and its measured values.

For each configuration, we compute GPG and ED on fixed evaluation pools and train a separate diffuser from the same in-house BAGEL continued-training checkpoint~\citep{bagel} to a common budget of $2.84\times10^{10}$ cumulative image tokens.
The images, initialization, architecture, and optimization recipe are identical across runs, leaving the caption configuration as the only changing variable.
We use BAGEL because repeating all $15$ runs with the substantially larger Qwen-Image backbone would be prohibitively expensive; Appendix~\ref{app:training} and Table~\ref{tab:hyperparams_bagel} provide the complete setup.

The sweep first shows that caption length does not explain the outcomes.
As Figure~\ref{fig:scaling_overview}c shows, NL and SP captions do not share a common length--loss trend: making NL captions longer nudges GPG upward (partly a length effect, since GPG is a token sum) yet leaves ED and converged loss essentially unchanged, whereas restoring SP fields raises both measures substantially and lowers loss, as shown in Figures~\ref{fig:gpg_ed_demo} and~\ref{fig:scaling_overview}a--b.
In contrast, measured informativeness provides the common axis.
Figure~\ref{fig:scaling_overview}d--e shows that, across all $15$ configurations, converged training MSE is well fit by a linear function of $\mathrm{GPG}$ and follows a power-law trend in $\mathrm{ED}$:
\begin{empheq}[box=\fcolorbox{black!55}{black!4}]{align}
\mathrm{MSE} &\;=\; 0.4549 \;-\; 8.45\times 10^{-5}\,\cdot\,\mathrm{GPG},
& r &= -0.984, \label{eq:scaling_law}\\[2pt]
\mathrm{MSE} &\;=\; 0.4200 \,\cdot\, \mathrm{ED}^{-0.2073},
& r &= -0.971. \label{eq:scaling_law_ed}
\end{empheq}
Here $r$ is Pearson correlation in raw space for GPG and log--log space for ED.
We call these two relations the \emph{scaling properties of text conditioning}: within the calibrated architecture, training recipe, and measured ranges, caption information predicts matched-budget converged diffusion loss.
Both fits are tight: their residual standard deviations are approximately $6\!\times\!10^{-4}$ for GPG and $7.8\!\times\!10^{-4}$ for the likelihood-free ED measure.
Figure~\ref{fig:scaling_overview}f further shows that the two measures agree closely on the configuration ranking despite their different scoring interfaces, with Spearman $\rho_{\mathrm{GPG,ED}}\!=\!0.96$.

This calibration has both scientific and practical consequences.
Scientifically, it elevates caption information from a descriptive property to a controlled training variable: with architecture, images, and compute fixed, it orders the loss reached across caption formats.
Practically, after one calibration it can screen candidate caption configurations before training another diffuser under the same recipe.
A configuration-level holdout directly tests this use: fits learned only from the NL and nested-SP families predict the converged MSE of all six spatial and field variants excluded from fitting.
The GPG- and ED-based fits achieve mean absolute MSE errors of $5.0\times10^{-4}$ and $8.1\times10^{-4}$, respectively, showing that the relation predicts configurations beyond those used to estimate it; at this error level, screening separates configurations whose loss gaps exceed the fit error rather than near-identical ones.

The remaining checks delimit the claim's scope.
Each configuration is trained once, so the resampling ranges in Figure~\ref{fig:scaling_overview}d--e measure sensitivity to the selected sweep settings, not run-to-run uncertainty or classical confidence intervals.
Changing the GPG judge or ED image proposer largely preserves the rankings, and budget-wise refits preserve both relations.
Appendices~\ref{app:judge} and~\ref{app:ed_extractors} report checks of the GPG judge and ED image proposer, respectively; Appendix~\ref{app:scaling_supplement} reports setting-resampling, budget-refit, and trailing-window analyses.
The calibration is also backbone-specific. It is fit on BAGEL, whereas the end-to-end system of \S\ref{sec:experiments} uses Qwen-Image; the matched control there shows that the structured interface still outperforms a free-form one on that backbone, but the quantitative GPG--loss relation is not re-fit for it. Like parameter, data, and compute laws~\citep{kaplan2020,chinchilla}, this is a recipe-specific empirical calibration rather than a universal theorem; Appendix~\ref{app:why_linear} gives its mutual-information motivation.

\paragraph{From the properties to two factors.}
When captions are annotated from paired images, the scaling properties isolate a diffusion-side capability: how effectively the caption interface exposes and organizes image-grounded supervision under a fixed training recipe.
We call this capability \emph{diffusability}, adapting the terminology of~\citet{skorokhodov2025diffusability}.
The NL/SP contrast identifies organization as a practical intervention, but it tests richer content and organization jointly rather than JSON syntax in isolation.
At inference, however, no paired image is available to supply that content; an LLM prompter must infer the visual variables from the user request.
We call this LLM-side capability \emph{promptability} and compare it through the quality of images generated under a fixed schema and diffuser.
End-to-end generation therefore depends on both the representation available to the diffuser and the prompter's ability to instantiate it.
Let $f$ denote the caption interface, including its represented field groups, and $\pi$ the LLM prompter; we summarize their joint role schematically, rather than as a fitted multiplicative quality law:
\begin{equation}
\boxed{
\underbrace{\mathrm{Quality}(f,\pi)}_{\text{system output}}
\;=\;
\underbrace{\mathrm{Diffusability}(f)}_{\text{diffusion side}}
\;\times\;
\underbrace{\mathrm{Promptability}(f,\pi)}_{\text{LLM side}}
}
\label{eq:factorization}
\end{equation}

\subsection{Raising Diffusability: constructing structured supervision}
\label{sec:sp}

\begin{figure}[!t]
  \centering
  \includegraphics[width=\linewidth]{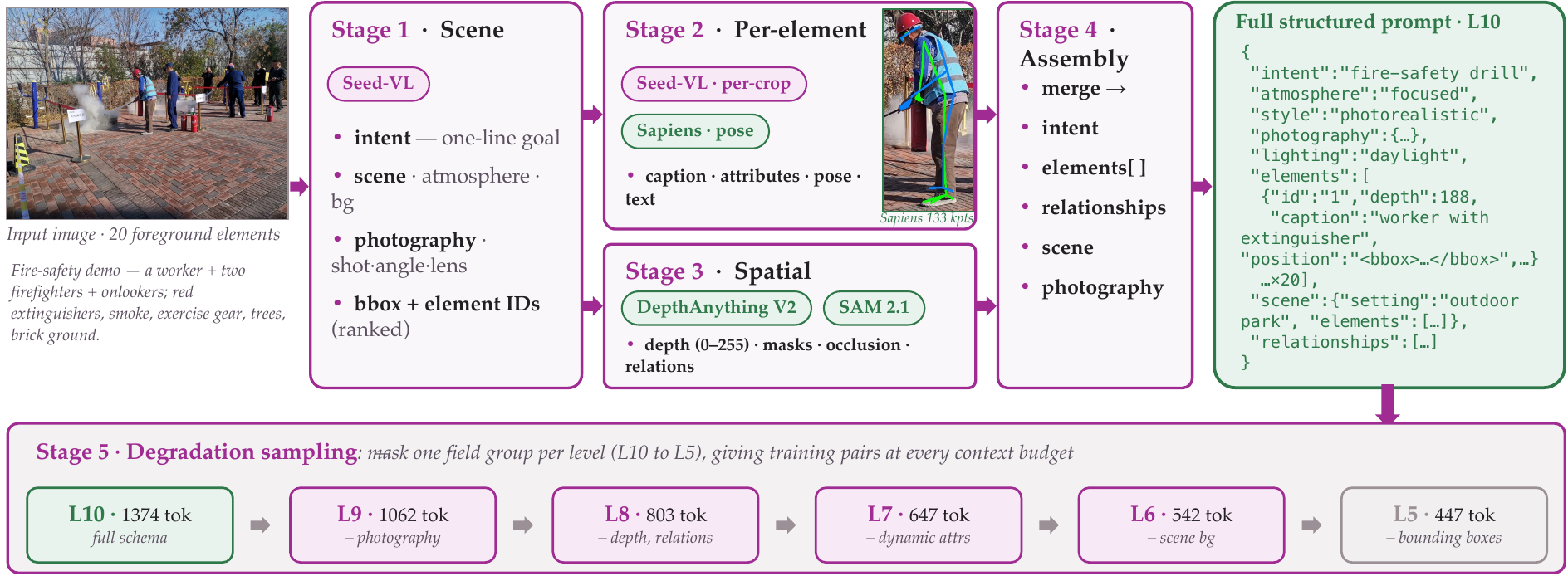}
  \caption{\textbf{Image-to-SP annotation with domain experts.}
  A VLM recovers global and element-level semantics; Sapiens, DepthAnything~V2, and SAM~2.1 provide complementary pose and geometric evidence; a final VLM pass assembles the full L10 SP, from which L5--L9 are derived by deterministically masking field groups.}
\label{fig:pipeline}
\end{figure}

\paragraph{Image-to-SP annotation with domain experts.}
To raise diffusability at corpus scale, we annotate each training image with a faithful, full-schema SP and use the resulting SP levels as structured supervision in diffusion training.
Constructing that supervision is a heterogeneous perception problem: it combines global semantics and local appearance and actions with specialized evidence for human pose, depth, extent, occlusion, and cross-element relations.
A general-purpose VLM handles the semantic content well but remains less reliable on body-side orientation and precise geometry; if serialized directly, these errors become explicit conditioning variables.
To combine these complementary signals, we propose the five-stage image-to-SP annotation pipeline shown in Figure~\ref{fig:pipeline}.
Frozen specialists extract pose and geometry evidence, and a final VLM reconciles it with the scene semantics into one coherent SP.
Appendix~\ref{app:impl_annotation} provides the implementation details.

The pipeline first establishes a shared semantic frame.
Stage~1 reads the full image to recover its intent, scene and atmosphere, style, lighting, and camera setup, while creating an ordered element inventory with identifiers and bounding boxes.
Stage~2 revisits each element crop so that the VLM can resolve local descriptions, attributes, actions, and photography with less interference from the surrounding scene.
Human elements additionally pass through Sapiens~\citep{sapiens}, whose $133$ pose keypoints are rendered as an overlay for the VLM; this evidence helps it disambiguate body-side orientation and joint geometry rather than infer them from appearance alone.
Stage~3 supplies complementary geometric evidence: DepthAnything~V2~\citep{depthanything2} estimates relative element depth, while SAM~2.1~\citep{sam21} provides masks and occlusion cues.
Bounding boxes, masks, and depth support geometric relations such as overlap, containment, relative position, and depth order.
Stage~4 reconciles this evidence with the global and crop-level semantics, infers semantic relations such as support and interaction, and serializes a well-formed L10 SP.
The expert outputs constrain this annotation rather than being copied as raw predictions: inferred human pose may be expressed textually under the element's \texttt{pose} and \texttt{action} keys, while raw keypoints and masks remain intermediate evidence.
Stage~5 then projects the full annotation into the controlled field ladder described below.

\begin{wraptable}{r}{0.63\linewidth}
  \centering
  \caption{\textbf{SP field ladder and downstream transfer.}
  L5--L9 are deterministic projections of L10. All levels are rendered by the BAGEL diffuser trained at that level, so these values are not comparable with the Qwen-Image systems of Table~\ref{tab:sota}. Tokens are averages per SP; GSB is measured against L5 over $N=150$ prompt pairs.}
  \label{tab:degradation}
  \footnotesize
  \renewcommand{\arraystretch}{1.08}
  \setlength{\tabcolsep}{3.2pt}
  \begin{tabular}{@{}crl@{\hspace{3pt}}rr@{}}
    \toprule
    Level & \shortstack{Avg.\\tokens} & Added fields & \shortstack{GenEval2\\GM $\uparrow$} & \shortstack{GSB vs.\\L5 $\uparrow$} \\
    \midrule
    L5  & 447  & Base fields             & 46.79 & --- \\
    L6  & 542  & Bounding boxes          & 48.65 & 8.0 \\
    L7  & 647  & Scene context           & 49.72 & 17.3 \\
    L8  & 803  & Dynamic attributes     & 52.94 & 22.7 \\
    L9  & 1062 & Depth \& relationships  & 55.16 & 24.7 \\
    \textbf{L10} & \textbf{1374} & photography & \textbf{57.70} & \textbf{26.0} \\
    \bottomrule
  \end{tabular}
\end{wraptable}

\paragraph{Controlled field ladder.}
To separate schema richness from annotation quality, stage 5 derives L5--L9 from each full L10 SP by deterministically masking predefined field groups.
Table~\ref{tab:degradation} presents the equivalent ascending view: moving from L5 to L10 successively restores bounding boxes, scene context, dynamic attributes, depth and relations, and element-level photography without re-annotating the image.
The SP levels are projections of one L10 annotation, and the NL controls are verbalized independently from the same underlying annotation evidence rather than converted from the SP JSON; the images and source annotations are fixed, and only the caption content and organization exposed to the learner vary.

Across the SP ladder, exposing more field groups raises GPG and ED and lowers loss along the calibrated relations of \S\ref{sec:metrics}.
Because every level is a deterministic projection of the same L10 annotation, the comparison varies the information and organization exposed by the schema without level-specific re-annotation.
To test whether this training-side gain transfers to generated images, a frozen Gemini~3~Pro generates one L10 SP zero-shot for each user prompt; L5--L9 are derived from that same output and rendered by the BAGEL diffuser trained at the corresponding level.
Table~\ref{tab:degradation} shows that GenEval2 GM and order-swapped GSB against L5 both improve monotonically with schema richness, demonstrating that higher diffusability benefits generated images rather than training loss alone.
A complementary field-wise ablation in Appendix~\ref{app:schema_field_ablation}, measured on converged training loss rather than the generation benchmarks above, identifies global scene context as the largest individual contributor, followed by bounding-box conditioning.
The verbal spatial variants follow the same relations, showing that the result is not tied to the default coordinate serialization.
The annotation pipeline thus supplies structured supervision at scale, while the controlled ladder operationalizes the diffusion-side intervention.

\subsection{Raising Promptability: scaling and training the LLM prompter}
\label{sec:prompter}

The annotation pipeline produces SPs from paired images during training, but inference begins only from a user request.
An LLM prompter must therefore translate that request into an SP.
This is more than a formatting task: the prompter must infer plausible visual details left unspecified by the user and fill in the schema with them without violating explicit constraints.
We therefore assess promptability through the quality of images rendered from its SPs, rather than through JSON validity or schema completeness alone.
We train one Qwen-Image diffuser on a mixture of SP levels and NL captions so that the same backbone can accommodate conditioning inputs with different structures and degrees of richness.
For the controlled promptability sweeps below, we fix this diffuser and the L10 schema and vary only the prompter $\pi$.
Differences in the resulting images therefore isolate its model, reasoning mode, or training configuration (training-pipeline ablations in \S\ref{sec:ablations}).

\paragraph{Off-the-shelf prompter scaling.}
Before applying task-specific training, we first test whether progress in general-purpose LLMs transfers to promptability through model scale and inference-time reasoning.
We use six frozen Qwen3.5 checkpoints, spanning $0.8$B to $397$B total parameters, as zero-shot prompters.
Each checkpoint maps user requests to L10 SPs in both non-thinking and chain-of-thought modes, and the same fixed Qwen-Image diffuser renders the resulting SPs.
Within each benchmark, the user prompts and image-generation settings are held fixed across checkpoints and reasoning modes, so only the prompter changes.
We evaluate basic compositional alignment with GenEval++, world-knowledge-conditioned generation with WISE, and open-ended generation with GPT-5.4 structure scores and good/same/bad (GSB) net preference against the smallest, $0.8$B prompter.
The broader promptability experiments additionally report GPT-5.4 alignment scores.

\begin{figure}[!t]
  \centering
  \includegraphics[width=\linewidth]{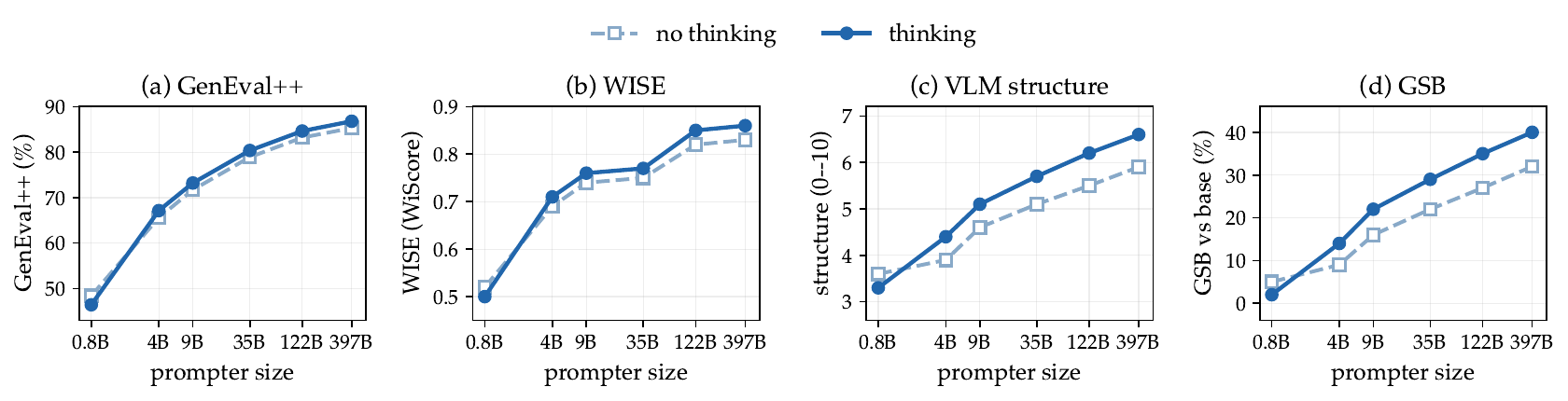}
  \caption{\textbf{Prompter scaling: generated-image quality improves with prompter size and reasoning mode.}
  Qwen3.5 prompters from $0.8$B to $397$B are evaluated in non-thinking and chain-of-thought modes with the L10 schema and Qwen-Image diffuser fixed.
  \emph{(a--b)}~GenEval++~\citep{ye2025echo4o} and WISE~\citep{niu2025wise} measure basic alignment and world-knowledge generation, respectively.
  \emph{(c--d)}~The offline GPT-5.4 structure scores and good/same/bad (GSB) net preferences (against the $0.8$B prompter) continue to improve with model scale and reasoning mode.
  Appendices~\ref{app:judge_prompts} and~\ref{app:judge_gsb} provide the structure/alignment rubrics and the pairwise rubric with order-swapped GSB aggregation, respectively.}
\label{fig:prompter_scaling}
\end{figure}

The results in Figure~\ref{fig:prompter_scaling} show that generated-image quality improves with prompter scale: in thinking mode, GenEval++ rises from $46.4\%$ at $0.8$B to $86.8\%$ at $397$B.
Chain-of-thought inference provides a further gain at every scale except $0.8$B, where reasoning often enters repetitive loops before producing valid JSON and therefore underperforms non-thinking inference.
Because none of these prompters receives task-specific training, the trend associates advances in general-purpose LLMs and reasoning mode with higher image quality through the SP interface.
GenEval++ and WISE capture broad gains with scale, while structure and GSB distinguish high-capacity prompters and reasoning modes more finely.

\begin{figure}[!t]
  \centering
  \includegraphics[width=\linewidth]{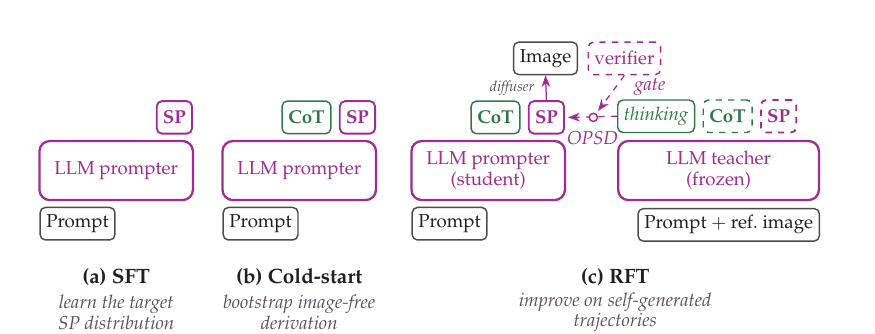}
  \caption{\textbf{The three-stage prompter-training pipeline (\S\ref{sec:prompter}).}
  \emph{(a)}~Supervised fine-tuning (SFT) learns the target SP distribution from (user prompt, SP) pairs.
  \emph{(b)}~Cold-start bootstraps image-free prompt-to-chain-of-thought (CoT)-to-SP derivation from privileged image-conditioned traces.
  \emph{(c)}~Reinforcement fine-tuning (RFT) renders the student's own rollouts; a QA verifier selects accepted trajectories, and on-policy self-distillation (OPSD) supplies targets from an image-conditioned teacher. Only the student is updated.}
  \label{fig:prompter_training}
\end{figure}

\paragraph{Training pipeline.}
The zero-shot sweep establishes transfer from general-purpose LLM progress, but even Qwen3.5-397B-A17B produces schema-valid SPs that carry insufficient visual detail.
The resulting images often appear overly simple and less realistic, especially for complex scenes and infographics, as illustrated in Figure~\ref{fig:abl_training_stages}.
Following the staged post-training paradigm used for reasoning LLMs, we improve the prompter through SFT, cold-start distillation, and RFT; Figure~\ref{fig:prompter_training} diagrams the pipeline.
\begin{itemize}[leftmargin=1.25em, topsep=0.2em, itemsep=0.15em]
  \item \textbf{Supervised fine-tuning (SFT)} first teaches the SP content distribution expected by the diffuser. Its core task pairs an image's original caption, treated as the user prompt, with the SP produced by our image-to-SP annotation pipeline. Across the corpus, these pairs teach a conditional prior over plausible SP completions, including objects, attributes, geometry, depth, and relations, rather than a deterministic prompt-to-layout mapping; Figure~\ref{fig:abl_sft} illustrates the resulting layout prior. A replay mixture of general reasoning and instruction examples preserves the base model's broader abilities. Appendix~\ref{app:training_prompter} details the replay composition.
  \item \textbf{Cold-start} teaches the prompter how to reason from a user prompt to a detailed SP. For each original image--caption pair, an image-conditioned VLM writes a trace showing how an image-free prompter can infer the visual decisions needed to construct the target SP. Because privileged image access can leak instance-specific observations into the trace, a Gemini judge rejects candidates that state such details without a prompt-grounded, common-sense, or explicit design rationale, as well as traces inconsistent with the caption, image, or SP. The prompter then learns from the accepted (user prompt $\to$ thinking trace $\to$ SP) examples without receiving the image. Appendix~\ref{app:impl_prompter} details trace construction, filtering, and training.
  \item \textbf{Reinforcement fine-tuning (RFT)} moves beyond offline teacher traces to the prompter's own rollouts from original image--caption pairs. The prompter generates a thinking trace and SP, which the fixed diffuser renders. A verifier filters these trajectories, and on-policy self-distillation (OPSD) trains the prompter on the accepted ones; we describe both components below.
\end{itemize}

Within RFT, the verifier observes only the user request and rendered image.
Using its score directly as a reward is unreliable: VLM judgments and stochastic rendering introduce both false acceptances and false rejections, so a single operating threshold cannot be assumed to provide both high precision and high recall.
We therefore impose a conservative threshold on alignment, structure, and aesthetics and use the verifier only as a high-precision acceptance gate.
This sacrifices rollout coverage but limits training to trajectories whose renders are judged prompt-faithful, structurally coherent, and visually acceptable.
Because the verifier still cannot supervise useful visual details left unspecified by the request, OPSD supplies the token-level update on the accepted trajectories.

OPSD is motivated by the same privileged-image versus image-free conditioning asymmetry measured by GPG. On each accepted rollout $\tau$, a frozen teacher $\pi^{\star}$ observes the paired reference image $I$, while the prompter $\pi_\theta$ remains image-free. OPSD minimizes the divergence between their next-token distributions along that rollout:
\begin{equation}
\mathcal{L}_{\text{OPSD}}(\theta) \;=\; \mathbb{E}_{\tau \sim \pi_\theta,\; \tau \,\text{accepted}}\!\left[\frac{1}{|\mathcal{T}_\tau|}\sum_{t\in\mathcal{T}_\tau} D\!\left(\pi^{\star}(\cdot \mid \tau_{<t}, \text{prompt}, I)\,,\; \pi_\theta(\cdot \mid \tau_{<t}, \text{prompt})\right)\right],
\label{eq:opsd}
\end{equation}
where the expectation is also over image--caption training pairs $(\text{prompt},I)\sim\mathcal{D}$, $\mathcal{T}_\tau$ indexes the rollout's response tokens, and $D$ is the token-level Kullback--Leibler (KL) divergence from the teacher to the prompter.
Appendix~\ref{app:opsd} details its implementation.
The verifier therefore determines which trajectories contribute to training, while OPSD transfers image-grounded token preferences from the teacher to the image-free prompter on those trajectories.

Together, the diffuser trained with structured supervision and the trained LLM prompter raise diffusability and promptability, respectively; \S\ref{sec:experiments} evaluates their combined system and ablates the corresponding diffusion-side and prompter-side interventions.

\section{Experiments}
\label{sec:experiments}

\begin{figure*}[!t]
  \centering
  \includegraphics[width=\textwidth]{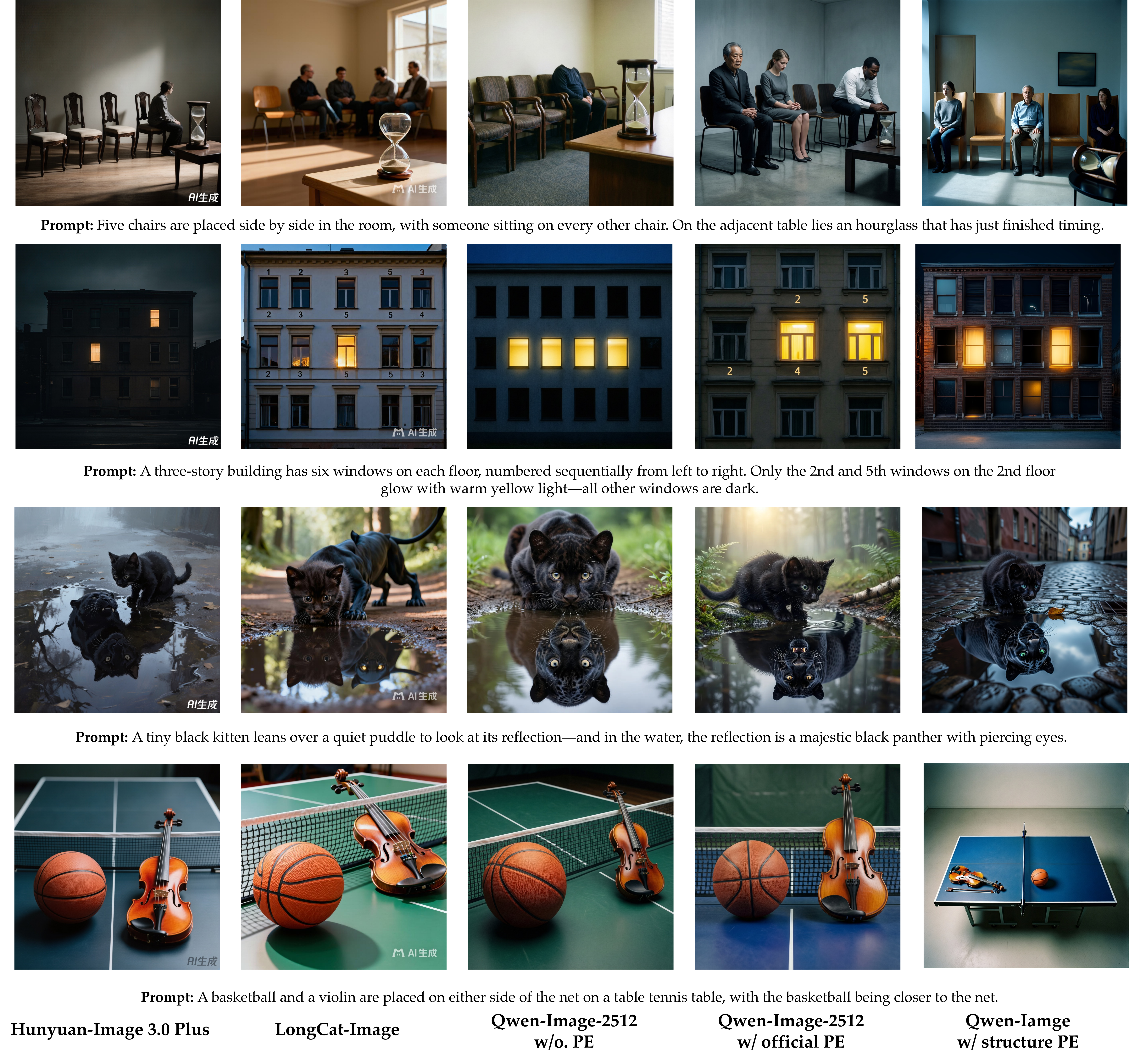}
  \caption{\textbf{Qualitative comparison.}
  Complex prompts requiring spatial reasoning, attribute binding, and compositional understanding.
  Columns from left to right: HunyuanImage~3.0, LongCat-Image, Qwen-Image without prompt enhancement (PE), Qwen-Image with official PE, and Ours (Qwen-Image with structured PE).
  In these examples, the structured-prompt system improves spatial layout, object count, and attribute binding without architectural changes to the diffusion backbone.}
  \label{fig:qualitative_comparison}
  \vspace{-5mm}
\end{figure*}

We first compare the complete prompter--diffuser system with representative text-to-image models (\S\ref{sec:main_results}). Holding the schema and diffuser fixed, we then isolate promptability, the LLM-side factor of Eq.~\eqref{eq:factorization}, through backend and training ablations (\S\ref{sec:ablations}). Finally, we test whether allocating inference-time compute to iterative refinement yields further gains (\S\ref{sec:agentic}).

\subsection{End-to-end performance and matched control}
\label{sec:main_results}

\textbf{Comparison with existing systems.}
Table~\ref{tab:sota} compares the resulting system with representative open-weight and closed-source generators. Among the evaluated open-weight systems, \ours{} leads or ties almost every reported metric and matches or surpasses the evaluated closed systems on most. The benchmarks probe complementary capabilities: GenEval and GenEval2 measure object-centric and compositional alignment~\citep{ghosh2023geneval,kamath2025geneval2}; DPG-Bench and TIIF stress dense and information-intensive prompt following~\citep{hu2024ella,tiif}; WISE evaluates world knowledge~\citep{niu2025wise}; and CoReBench targets composition and reasoning~\citep{li2026corebench}. Consistent gains across these settings show that the benefit is not confined to a single prompt regime or evaluator, but extends from basic alignment to knowledge-dependent and reasoning-intensive generation. Figures~\ref{fig:gallery} and~\ref{fig:qualitative_comparison} reflect the same breadth qualitatively: on complex prompts, the structured-prompt system more faithfully realizes spatial layouts, object counts, and attribute bindings.

\textbf{Matched NL control.}
To verify that the gains arise from using structured prompts as the shared caption interface between the prompter and diffuser, rather than from a larger backbone or additional training alone, we compare against two controls built on the same Qwen-Image architecture. The official Qwen-Image prompt enhancer reaches $52.8$ GenEval2 GM and $74.7$ CoReBench, compared with $72.5$ and $85.2$ for our system. More stringently, matched NL retrains the same Qwen-Image diffuser and prompter on the same images, stages, and budgets while retaining a free-form caption interface. This additional training improves the two scores to $56.2$ and $76.1$, but remains well below the matched end-to-end SP system. Thus, retraining the same architecture and data with an NL interface does not reproduce the SP system's gains. Appendix~\ref{app:impl_annotation} details the matched control, and Appendix~\ref{app:additional_results} provides category-level results.

\begin{table*}[!t]
  \centering
  \caption{\textbf{Comparison with representative text-to-image systems.}
  Higher is better. GenEval2 reports Soft-TIFA arithmetic/geometric means (AM/GM), and TIIF reports short/long accuracy on the \emph{testmini} subset.
  ``PE'' denotes prompt enhancement; Qwen-Image$^{*}$ uses its official PE, and $^{\dagger}$ marks our Qwen-Image-2512 re-evaluation.
  \ours{} uses single-shot inference. Evaluation protocols and score provenance are detailed in Appendix~\ref{app:additional_results}.}
  \label{tab:sota}
  \small
  \resizebox{\linewidth}{!}{%
  \begin{tabular}{@{}lcccccc@{}}
    \toprule
    Model & GenEval & GenEval2 (AM/GM) & DPG & TIIF (s/l) & WISE & CoReBench \\
    \midrule
    \multicolumn{7}{@{}l}{\textcolor{gray}{\emph{w/o PE}}} \\
    FLUX.1 Dev~\citep{bfl2024flux}       & 0.67 & 67.1/21.1 & 83.84 & 71.1/71.8 & 0.50 & 42.2 \\
    OmniGen2~\citep{wu2025omnigen2}      & 0.80 & --        & 83.57 & --        & --   & 42.9 \\
    Emu3.5~\citep{cui2025emu35}          & 0.86 & --        & 87.46 & \textbf{89.5}/88.2 & 0.57 & --   \\
    BAGEL~\citep{bagel}                  & 0.82 & --        & 85.07 & 71.5/71.7 & 0.52 & 38.2 \\
    Qwen-Image~\citep{qwenimage}         & 0.87 & 80.8/33.8 & 88.32 & 86.1/86.8 & 0.62 & 58.9 \\
    \midrule
    \multicolumn{7}{@{}l}{\textcolor{gray}{\emph{w/\ PE}}} \\
    BAGEL + CoT~\citep{bagel}                 & 0.88 & 70.9/23.1 & --    & --        & 0.70 & 41.1 \\
    GPT-Image-1~\citep{openai2025gptimage1} & 0.84 & -- & 85.15 & 89.2/88.3 & 0.80 & 72.6 \\
    Nano Banana~\citep{google2025nanobanana} & 0.89 & 82.8/44.6 & 85.23 & --        & \textbf{0.89} & 75.9 \\
    LongCat-Image~\citep{longcatimage2025}       & 0.87 & --        & 86.80 & --        & 0.65 & 59.6 \\
    HunyuanImage 3.0~\citep{hunyuanimage2025}    & 0.72 & --        & 86.10 & --        & 0.57 & 58.7 \\
    Qwen-Image$^{*}$                          & 0.91 & 82.4/52.8$^{\dagger}$ & 87.20$^{\dagger}$ & 88.3/88.4$^{\dagger}$ & 0.83 & 74.7$^{\dagger}$ \\
    Matched NL + Qwen-Image                    & 0.91 & 84.5/56.2 & 87.80 & 88.5/88.0 & 0.84 & 76.1 \\
    \midrule
    \ours{} (Qwen-Image)  & \textbf{0.94} & \textbf{90.6}/\textbf{72.5} & \textbf{90.71} & 89.1/\textbf{89.2} & \textbf{0.89} & \textbf{85.2} \\
    \bottomrule
\end{tabular}}
\end{table*}

\subsection{Isolating promptability}
\label{sec:ablations}

\begin{wrapfigure}{r}{0.40\linewidth}
  \vspace{-3mm}
  \centering
  \includegraphics[width=\linewidth]{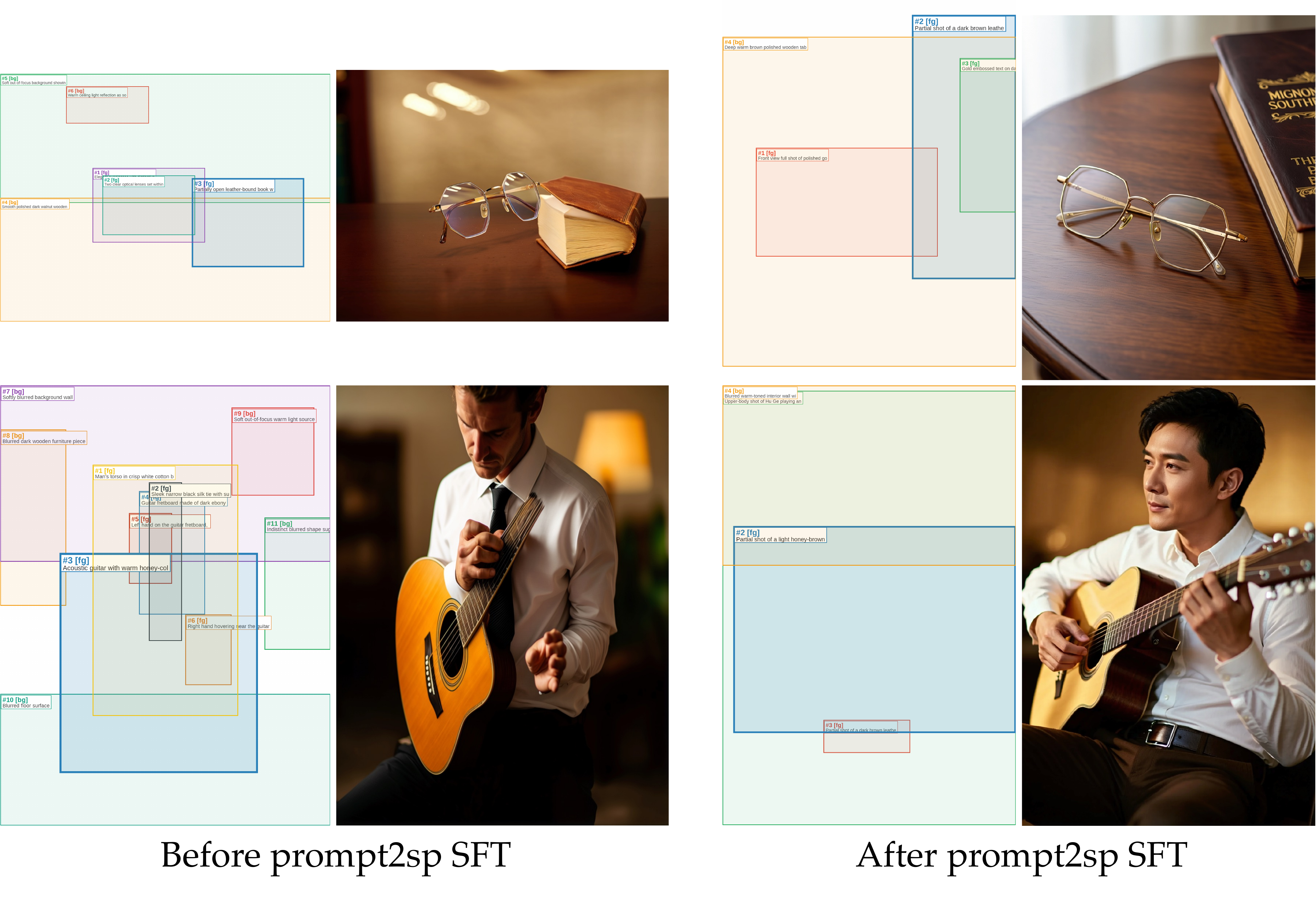}
  \caption{\textbf{Stage-1 SFT recovers a non-canonical crop.}}
  \label{fig:abl_sft}
  \vspace{-2mm}
\end{wrapfigure}
\textbf{Promptability: transferring LLM progress.}
We next ask whether advances in general-purpose LLMs can transfer through the SP interface into better text-to-image generation. To isolate this question, Table~\ref{tab:abl_backend} holds the schema and Qwen-Image diffuser fixed while varying the model and inference procedure used to produce the prompt. Within the single-turn rows, zero-shot SP filling improves alignment and GSB over the same LLM's NL rewrite, but consistently lowers DPG-Bench and structure. The LLM therefore provides genuine system-level headroom, but zero-shot schema filling alone does not fully realize it. Training the prompter closes this gap; our model is the strongest single-turn backend on all four metrics.
The coding-agent rows use each agent's native multi-turn procedure, so they change both the model and the inference process rather than providing a model-only backend control. Their strong results nevertheless suggest that iterative revision adds value, motivating the controlled common-harness experiment of Section~\ref{sec:agentic}.

\begin{table}[t]
  \centering
  \caption{\textbf{Rewriting backend comparison.}
  Schema and Qwen-Image are fixed; rows vary the prompt-producing backend and, where marked, its inference mode.
  ``NL rewrite'' is the matched free-form baseline, ``agentic'' uses each coding agent's native loop, and ``Ours'' reproduces the final trained-prompter row of Table~\ref{tab:abl_training}.
  DPG-Bench and offline GPT-5.4 evaluations follow Table~\ref{tab:abl_training}. GSB is the net pairwise preference $100(n_{\mathrm{Good}}-n_{\mathrm{Bad}})/N$ against the zero-shot, single-shot Qwen3.5-397B-A17B Base prompter; each pair is judged in both image orders, with inconsistent orderings counted as Same, over $N=150$ prompt pairs (Appendix~\ref{app:judge_gsb}). Bold marks the best in each column.}
  \label{tab:abl_backend}
  \small
  \setlength{\tabcolsep}{4pt}
  \resizebox{\linewidth}{!}{%
  \begin{tabular}{@{}llcccc@{}}
    \toprule
    & & & \multicolumn{2}{c}{Offline Judge (GPT-5.4, 0--10)} & \\
    \cmidrule(lr){4-5}
    Rewriting Backend & Mode & DPG-Bench & Structure & Alignment & GSB vs Base (\%) \\
    \midrule
    GPT-5.5            & NL rewrite  & 90.18 & 6.913 & 8.747 & 36.0 \\
    Claude Opus 4.8    & NL rewrite  & 89.93 & 6.733 & 8.617 & 32.7 \\
    GLM-5.2            & NL rewrite  & 89.47 & 6.653 & 8.493 & 30.0 \\
    Gemini 3 Pro       & NL rewrite  & 88.72 & 6.183 & 8.107 & 19.3 \\
    \midrule
    GPT-5.5            & single-turn & 89.84 & 6.840 & 8.907 & 37.3 \\
    Claude Opus 4.8    & single-turn & 89.21 & 6.653 & 8.793 & 34.7 \\
    GLM-5.2            & single-turn & 88.93 & 6.557 & 8.746 & 32.0 \\
    Gemini 3 Pro       & single-turn & 87.86 & 6.020 & 8.687 & 21.3 \\
    \midrule
    Codex              & agentic     & 90.32 & 7.360 & 8.980 & 38.7 \\
    Claude Code        & agentic     & 90.63 & 7.560 & \textbf{9.087} & \textbf{44.7} \\
    \midrule
    Ours (LoRA-on-Qwen3.5-397B-A17B + RFT) & single-turn & \textbf{90.71} & \textbf{7.600} & 9.047 & 42.0 \\
    \bottomrule
  \end{tabular}}
\end{table}

\begin{figure}[t]
  \centering
  \includegraphics[width=0.90\linewidth]{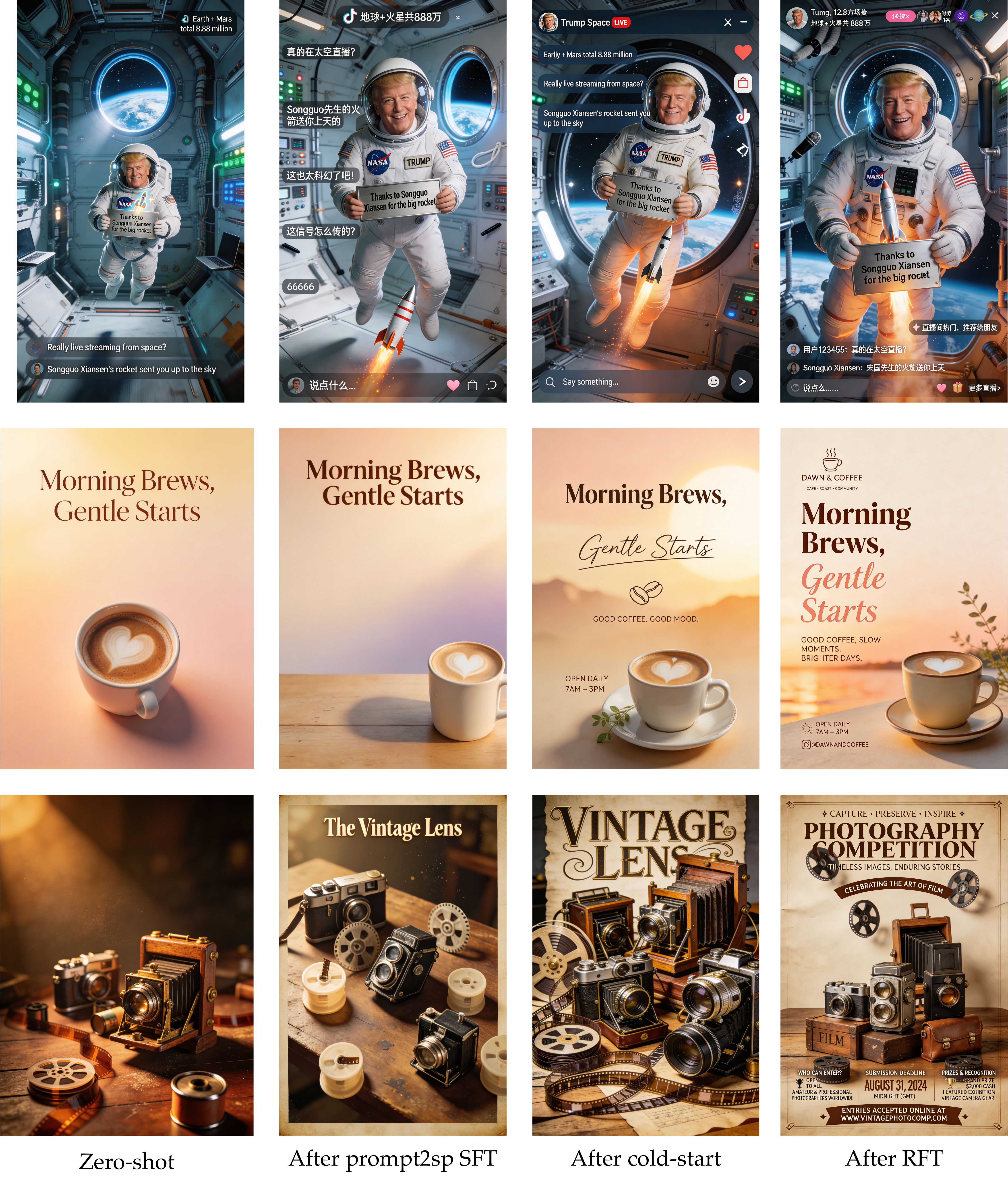}
  \caption{\textbf{Qualitative progression across prompter training stages.}
  Each row uses one user prompt and the same Qwen-Image backbone; across these multilingual, dense-text, and poster-design examples, text fidelity, compositional density, and layout structure improve stage by stage.}
  \label{fig:abl_training_stages}
\end{figure}

\noindent\textbf{Promptability: training the prompter.}
We next study how task-specific training improves promptability by examining the contribution of each training stage and supervision signal. Table~\ref{tab:abl_training} holds the schema and diffuser fixed, cumulatively adds SFT and Cold-start, and then varies the RFT signal. SFT produces the largest single-stage gain in structure ($4.860\!\rightarrow\!6.273$) and a $23.3\%$ GSB preference, consistent with learning plausible SP content rather than JSON syntax alone; Figure~\ref{fig:abl_sft} illustrates the acquired layout prior. Cold-start further improves structure and GSB, although alignment decreases slightly, showing that its privileged traces do not improve every criterion uniformly. Verifier-reward GRPO and ungated OPSD both improve structure, alignment, and GSB over Cold-start; combining high-confidence rollout selection with dense image-conditioned OPSD targets gives the strongest endpoint. Across the full pipeline, DPG-Bench changes by only $1.29$ points, whereas structure rises to $7.600$ and GSB to $42.0\%$. DPG-Bench is therefore comparatively insensitive to the structural and compositional differences visible in Figure~\ref{fig:abl_training_stages}. Overall, the ablation supports the intended division of labor: SFT learns the target SP distribution, Cold-start teaches image-free derivation, and RFT improves the prompter on its own rollouts.

\begin{table}[h]
  \centering
  \caption{\textbf{Prompter training-pipeline ablation.}
  Rows add SFT and Cold-start, then compare Group Relative Policy Optimization (GRPO) with the QA verifier score as reward~\citep{shao2024deepseekmath}, ungated OPSD, and the full verifier-gated OPSD rule.
  Schema (L10) and Qwen-Image are fixed, so only prompter training changes.
  Structure/alignment ($0$--$10$) and GSB are evaluated offline by GPT-5.4; GSB is the order-swapped net preference against the zero-shot, single-shot Qwen3.5-397B-A17B Base prompter over $N=150$ prompt pairs (Appendix~\ref{app:judge_gsb}). The training verifier's aesthetic score is used only for rollout acceptance.}
  \label{tab:abl_training}
  \small
  \setlength{\tabcolsep}{5pt}
  \begin{tabular}{@{}lcccc@{}}
    \toprule
    & & \multicolumn{2}{c}{Offline Judge (GPT-5.4, $0$--$10$)} & \\
    \cmidrule(lr){3-4}
    Training stage & DPG-Bench & Structure & Alignment & GSB (\%) \\
    \midrule
    Base                              & 89.42 & 4.860 & 8.307 & --- \\
    + SFT                             & 89.61 & 6.273 & 8.473 & $23.3$ \\
    + Cold-start                      & 89.84 & 6.753 & 8.360 & $28.7$ \\
    + RFT: verifier-reward GRPO        & 89.73 & 7.113 & 8.907 & $36.7$ \\
    + RFT: ungated OPSD                & 89.68 & 6.993 & 8.747 & $33.3$ \\
    + RFT: verifier-gated OPSD         & \textbf{90.71} & \textbf{7.600} & \textbf{9.047} & \textbf{42.0} \\
    \bottomrule
  \end{tabular}
\end{table}

\subsection{Inference-time scaling of promptability}
\label{sec:agentic}

The strong coding-agent results in Table~\ref{tab:abl_backend} suggest that their native multi-turn harnesses contribute beyond the underlying rewriting backend. Motivated by this observation, we isolate iterative inference from backend choice by placing both the Base and trained prompters in the same refine--render--judge loop. Prior work scales inference on the input side by training a prompt rewriter offline~\citep{chen2025inputside}; here the loop runs online at generation time and the prompter weights are held fixed. The schema, diffuser, online judge, and prompter weights remain fixed within each comparison; only the available refinement budget changes. Because SPs expose visual decisions in named fields, critiques can be translated into targeted revisions instead of rewriting the entire caption.

\begin{figure}[!t]
  \centering
  \includegraphics[width=\linewidth]{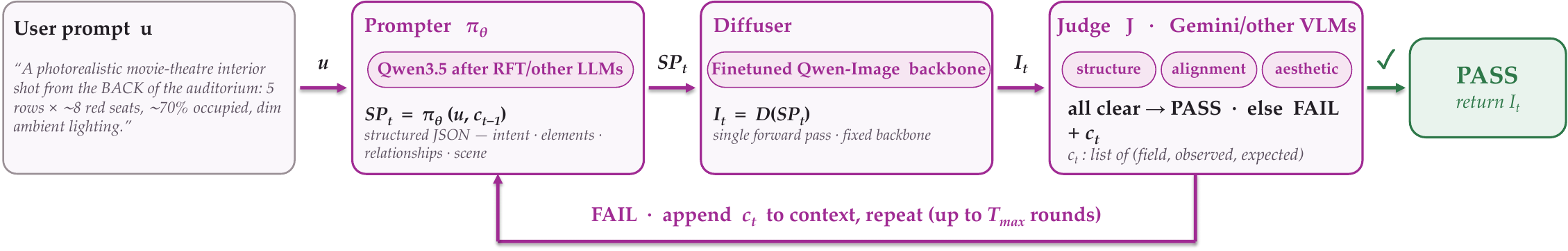}
  \caption{\textbf{The agentic inference-time loop.}
  At round $t$, the prompter emits $\mathrm{SP}_t$ from the user prompt and accumulated critique; the fixed diffuser renders $I_t$; and the Gemini judge returns PASS or field-level critique over structure, alignment, and aesthetics.
  Failed rounds edit targeted schema slots and stop at PASS or $T_{\max}$.}
  \label{fig:agentic_loop}
\end{figure}

\begin{figure}[t]
  \centering
  \includegraphics[width=\linewidth]{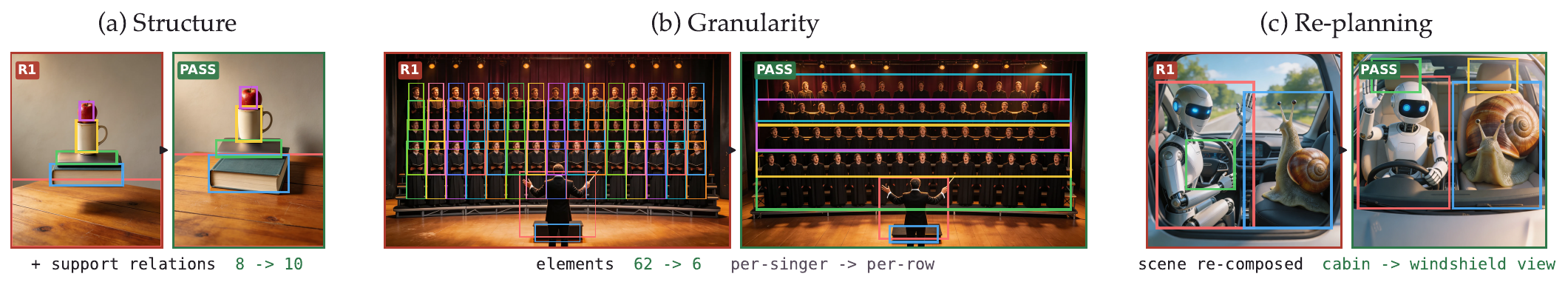}
  \caption{\textbf{Case types the agentic loop resolves.}
  Each pair shows the round-$1$ defect (\textcolor[HTML]{B03A2E}{R1}) and the accepted result (\textcolor[HTML]{2E7849}{PASS}), with the structured edit beneath.
  Structure fixes add missing support relations; granularity fixes collapse over-enumerated elements to renderable groups; larger layout fixes re-compose the scene when local edits are insufficient.}
  \label{fig:agentic_cases}
\end{figure}

\paragraph{Refine--render--judge loop.}
As shown in Figure~\ref{fig:agentic_loop}, the loop maintains the previous SPs and the critique history as explicit state. At round $t$, the prompter uses this state together with the original user request to produce $\mathrm{SP}_t$, and the fixed diffuser renders $I_t$. The online Gemini judge sees only the user request and $I_t$, not the SP. It checks prompt-derived requirements alongside structural and aesthetic quality, then returns per-axis scores, a PASS/FAIL decision, and a structured list of observed failures. A PASS immediately returns $I_t$; after a FAIL, the critique is appended to the prompter context so that the next round can revise the corresponding objects, attributes, relations, or layout fields. Repeated failures permit progressively broader changes, from local field edits to element regrouping and full scene re-planning, as illustrated in Figure~\ref{fig:agentic_cases}. The loop terminates at PASS or $T_{\max}$. Reported results are evaluated by GPT-5.4 in a separate offline pass rather than by the online Gemini judge, so the control signal and the final evaluation use different models, though both apply the same rubrics (Appendix~\ref{app:judge_prompts}); Appendix~\ref{app:impl_inference} specifies the formal loop, $6/10$ PASS threshold, and per-round cost.

\paragraph{Returns from additional rounds.}
Table~\ref{tab:agentic} shows that both prompters improve with $T_{\max}$ when the schema, diffusion backbone, judge, and prompter weights are held fixed. The larger change is in structure: from one to eight rounds it rises by $1.247$ for Base and $0.660$ for the trained prompter, compared with alignment gains of $0.366$ and $0.266$. Together with Figure~\ref{fig:agentic_cases}, this indicates that feedback primarily repairs object decomposition, relations, and layout rather than basic prompt alignment.

Placing the trained prompter in this loop also answers the agentic backends of Table~\ref{tab:abl_backend}: at $T_{\max}\!=\!8$ it reaches $54.7\%$ GSB, above the $44.7\%$ of the strongest coding agent, which is itself already multi-turn. Training absorbs much of the work that would otherwise require iterative correction. Even after eight rounds, Base remains below the trained prompter in a single shot on both structure ($6.107$ vs.\ $7.600$) and GSB ($24.0\%$ vs.\ $42.0\%$). Under the same eight-round limit, the trained prompter also reaches PASS after only $2.31$ rounds on average, compared with $3.41$ for Base. Thus, prompter training not only raises the starting point but also shortens the subsequent refinement trajectory; agentic inference complements training rather than replacing it.

\begin{table}[h]
  \centering
  \caption{\textbf{Agentic inference-time scaling.}
  At fixed schema, Qwen-Image backbone, online Gemini judge, and prompter weights, increasing $T_{\max}$ allocates more refine--render--judge rounds.
  Final outputs are evaluated offline by GPT-5.4 for structure/alignment and order-swapped GSB net preference against the zero-shot Qwen3.5-397B-A17B Base prompter at $T_{\max}\!=\!1$, over the $150$-prompt evaluation pool ($N=150$, Appendix~\ref{app:judge_gsb}); average rounds measures refinement rounds consumed, each of which issues three Gemini judge calls, and $T_{\max}\!=\!1$ is single-shot inference.}
  \label{tab:agentic}
  \small
  \setlength{\tabcolsep}{5pt}
  \begin{tabular}{@{}llcccc@{}}
    \toprule
    & & \multicolumn{2}{c}{Offline Judge (GPT-5.4, $0$--$10$)} & & \\
    \cmidrule(lr){3-4}
    Prompter & $T_{\max}$ & Structure & Alignment & GSB (\%) & Avg.\ rounds \\
    \midrule
    Base (zero-shot)             & $1$ & $4.860$ & $8.307$ & ---     & $1.00$ \\
    Base (zero-shot)             & $2$ & $5.387$ & $8.493$ & $14.7$  & $1.74$ \\
    Base (zero-shot)             & $4$ & $6.027$ & $8.640$ & $22.7$  & $2.83$ \\
    Base (zero-shot)             & $8$ & $6.107$ & $8.673$ & $24.0$  & $3.41$ \\
    \midrule
    Trained (SFT + Cold-start + RFT) & $1$ & $7.600$         & $9.047$         & $42.0$         & $1.00$ \\
    Trained (SFT + Cold-start + RFT) & $2$ & $7.940$         & $9.173$         & $49.3$         & $1.51$ \\
    Trained (SFT + Cold-start + RFT) & $4$ & $8.213$         & $9.293$         & $54.0$         & $2.04$ \\
    Trained (SFT + Cold-start + RFT) & $8$ & $\mathbf{8.260}$ & $\mathbf{9.313}$ & $\mathbf{54.7}$ & $2.31$ \\
    \bottomrule
  \end{tabular}
\end{table}

The effective refinement horizon is also short. For the trained prompter, increasing $T_{\max}$ from four to eight raises the average rounds only from $2.04$ to $2.31$, while structure changes from $8.213$ to $8.260$ and GSB from $54.0\%$ to $54.7\%$. The average therefore stays close to two rounds, and little is gained by permitting a substantially longer trajectory. Within the tested schema, judge, refinement policy, and diffuser, text-to-image generation benefits from iterative correction but does not exhibit a strong need for long-horizon prompt-side reasoning: once the main specification errors are repaired, further render--feedback rounds quickly saturate.

\section{Conclusion}
\label{sec:conclusion}

We show that caption \emph{information content}, rather than caption length, is a measurable and scalable axis in text-to-image learning. Across controlled caption configurations, GPG and ED predict converged diffusion loss, defining empirical scaling properties for text conditioning that can be used to compare caption representations after calibration. This supports a \emph{Diffusability}$\times$\emph{Promptability} view of the caption interface: structured prompts raise diffusability by exposing and organizing image-grounded variables in addressable fields, while scaling and training the LLM prompter raises promptability by translating user requests into detailed, coherent instances of that representation. Matched natural-language retraining shows that the structured-representation gains are not explained by additional training alone, and the resulting end-to-end system improves visual coherence, prompt fidelity, and compositional generation without changing the diffusion architecture. At inference, short refine--render--judge loops provide further gains, but the trained prompter internalizes much of the correction and additional rounds yield rapidly diminishing returns; prompt-side inference compute is therefore a complement to training rather than a substitute for it.

\paragraph{Limitations.}
Both metrics depend on image-conditioned measurement: GPG queries a vision--language judge at scoring time, while ED requires a one-time offline extraction of attribute tuples from each image.
The fitted relations may shift under different judges or extractors. Agreement between the two measures reduces, but does not eliminate, the risk that the shared trend reflects a particular scoring interface.
The schema is hand-designed; automatic schema discovery and extension to video and three-dimensional generation remain open.
The structured-prompt path also adds prompter latency, so deployment must weigh the generation gains against this additional inference cost.

\bibliographystyle{iclr2026_conference}
\bibliography{Bibliography_base}

@article{bagel,
  title       = {Emerging Properties in Unified Multimodal Pretraining},
  author      = {Deng, Chaorui and Zhu, Deyao and Li, Kunchang and Gou, Chenhui and Li, Feng and Wang, Zeyu and Zhong, Shu and Yu, Weihao and Nie, Xiaonan and Song, Ziang and Shi, Guang and Fan, Haoqi},
  journal     = {arXiv preprint arXiv:2505.14683},
  year        = {2025},
  eprint      = {2505.14683},
  archivePrefix = {arXiv},
  primaryClass = {cs.CV}
}

@techreport{betker2023dalle3,
  title       = {Improving Image Generation with Better Captions},
  author      = {Betker, James and Goh, Gabriel and Jing, Li and Brooks, Tim and Wang, Jianfeng and Li, Linjie and Ouyang, Long and Zhuang, Juntang and Lee, Joyce and Guo, Yufei and Manassra, Wesam and Dhariwal, Prafulla and Chu, Casey and Jiao, Yunxin and Ramesh, Aditya},
  institution = {OpenAI},
  year        = {2023},
  url         = {https://cdn.openai.com/papers/dall-e-3.pdf}
}

@misc{bfl2024flux,
  title        = {Announcing {Black Forest Labs}},
  author       = {{Black Forest Labs}},
  year         = {2024},
  howpublished = {\url{https://blackforestlabs.ai/announcing-black-forest-labs/}},
  note         = {Model release}
}

@article{caption_detailness2025,
  title       = {Harnessing Caption Detailness for Data-Efficient Text-to-Image Generation},
  author      = {Wang, Xinran and Diao, Muxi and Liu, Yuanzhi and Wang, Chunyu and Liang, Kongming and Ma, Zhanyu and Guo, Jun},
  journal     = {arXiv preprint arXiv:2505.15172},
  year        = {2025},
  eprint      = {2505.15172},
  archivePrefix = {arXiv},
  primaryClass = {cs.CV}
}

@article{chefer2023attendexcite,
  title       = {Attend-and-Excite: Attention-Based Semantic Guidance for Text-to-Image Diffusion Models},
  author      = {Chefer, Hila and Alaluf, Yuval and Vinker, Yael and Wolf, Lior and Cohen-Or, Daniel},
  journal     = {ACM Transactions on Graphics (SIGGRAPH)},
  year        = {2023},
  volume      = {42},
  number      = {4},
  eprint      = {2301.13826},
  archivePrefix = {arXiv},
  primaryClass = {cs.CV}
}

@inproceedings{chen2023pixartalpha,
  title       = {{PixArt-$\alpha$}: Fast Training of Diffusion Transformer for Photorealistic Text-to-Image Synthesis},
  author      = {Chen, Junsong and Yu, Jincheng and Ge, Chongjian and Yao, Lewei and Xie, Enze and Wu, Yue and Wang, Zhongdao and Kwok, James and Luo, Ping and Lu, Huchuan and Li, Zhenguo},
  booktitle   = {International Conference on Learning Representations (ICLR)},
  year        = {2024},
  eprint      = {2310.00426},
  archivePrefix = {arXiv},
  primaryClass = {cs.CV}
}

@inproceedings{chen2024pixartsigma,
  title       = {{PixArt-$\Sigma$}: Weak-to-Strong Training of Diffusion Transformer for {4K} Text-to-Image Generation},
  author      = {Chen, Junsong and Ge, Chongjian and Xie, Enze and Wu, Yue and Yao, Lewei and Ren, Xiaozhe and Wang, Zhongdao and Luo, Ping and Lu, Huchuan and Li, Zhenguo},
  booktitle   = {European Conference on Computer Vision (ECCV)},
  year        = {2024},
  eprint      = {2403.04692},
  archivePrefix = {arXiv},
  primaryClass = {cs.CV}
}

@inproceedings{chinchilla,
  title       = {Training Compute-Optimal Large Language Models},
  author      = {Hoffmann, Jordan and Borgeaud, Sebastian and Mensch, Arthur and Buchatskaya, Elena and Cai, Trevor and Rutherford, Eliza and Casas, Diego de Las and Hendricks, Lisa Anne and Welbl, Johannes and Clark, Aidan and others},
  booktitle   = {Advances in Neural Information Processing Systems (NeurIPS)},
  year        = {2022},
  eprint      = {2203.15556},
  archivePrefix = {arXiv},
  primaryClass = {cs.CL}
}

@inproceedings{clipscore,
  title       = {{CLIPScore}: A Reference-free Evaluation Metric for Image Captioning},
  author      = {Hessel, Jack and Holtzman, Ari and Forbes, Maxwell and Le Bras, Ronan and Choi, Yejin},
  booktitle   = {Empirical Methods in Natural Language Processing (EMNLP)},
  year        = {2021},
  eprint      = {2104.08718},
  archivePrefix = {arXiv},
  primaryClass = {cs.CV}
}

@article{cui2025emu35,
  title       = {{Emu3.5}: Native Multimodal Models are World Learners},
  author      = {Cui, Yufeng and Chen, Honghao and Deng, Haoge and Huang, Xu and Li, Xinghang and
                 Liu, Jirong and Liu, Yang and Luo, Zhuoyan and Wang, Jinsheng and others},
  journal     = {arXiv preprint arXiv:2510.26583},
  year        = {2025},
  eprint      = {2510.26583},
  archivePrefix = {arXiv},
  primaryClass = {cs.CV}
}

@inproceedings{depthanything2,
  title       = {Depth Anything {V2}},
  author      = {Yang, Lihe and Kang, Bingyi and Huang, Zilong and Zhao, Zhen and Xu, Xiaogang and Feng, Jiashi and Zhao, Hengshuang},
  booktitle   = {Advances in Neural Information Processing Systems (NeurIPS)},
  year        = {2024},
  eprint      = {2406.09414},
  archivePrefix = {arXiv},
  primaryClass = {cs.CV}
}

@inproceedings{esser2024sd3,
  title       = {Scaling Rectified Flow Transformers for High-Resolution Image Synthesis},
  author      = {Esser, Patrick and Kulal, Sumith and Blattmann, Andreas and Entezari, Rahim and M{\"u}ller, Jonas and Saini, Harry and Levi, Yam and Lorenz, Dominik and Sauer, Axel and Boesel, Frederic and Podell, Dustin and Dockhorn, Tim and English, Zion and Lacey, Kyle and Goodwin, Alex and Marek, Yannik and Rombach, Robin},
  booktitle   = {International Conference on Machine Learning (ICML)},
  year        = {2024},
  eprint      = {2403.03206},
  archivePrefix = {arXiv},
  primaryClass = {cs.CV}
}

@inproceedings{gligen,
  title       = {{GLIGEN}: Open-Set Grounded Text-to-Image Generation},
  author      = {Li, Yuheng and Liu, Haotian and Wu, Qingyang and Mu, Fangzhou and Yang, Jianwei and Gao, Jianfeng and Li, Chunyuan and Lee, Yong Jae},
  booktitle   = {Conference on Computer Vision and Pattern Recognition (CVPR)},
  year        = {2023},
  eprint      = {2301.07093},
  archivePrefix = {arXiv},
  primaryClass = {cs.CV}
}

@misc{google2025nanobanana,
  title        = {Nano Banana (Gemini 2.5 Flash Image)},
  author       = {{Google}},
  year         = {2025},
  howpublished = {\url{https://blog.google/products/gemini/updated-image-editing-model/}},
  note         = {Model release}
}

@article{hampel1974,
  title     = {The Influence Curve and its Role in Robust Estimation},
  author    = {Hampel, Frank R.},
  journal   = {Journal of the American Statistical Association},
  volume    = {69},
  number    = {346},
  pages     = {383--393},
  year      = {1974},
  publisher = {Taylor \& Francis}
}

@inproceedings{ho2020ddpm,
  title     = {Denoising Diffusion Probabilistic Models},
  author    = {Ho, Jonathan and Jain, Ajay and Abbeel, Pieter},
  booktitle = {Advances in Neural Information Processing Systems (NeurIPS)},
  year      = {2020},
  eprint    = {2006.11239},
  archivePrefix = {arXiv},
  primaryClass  = {cs.LG}
}

@article{kaplan2020,
  title   = {Scaling Laws for Neural Language Models},
  author  = {Kaplan, Jared and McCandlish, Sam and Henighan, Tom and Brown, Tom B. and Chess, Benjamin and Child, Rewon and Gray, Scott and Radford, Alec and Wu, Jeffrey and Amodei, Dario},
  journal = {arXiv preprint arXiv:2001.08361},
  year    = {2020},
  eprint  = {2001.08361},
  archivePrefix = {arXiv},
  primaryClass  = {cs.LG}
}

@inproceedings{laion,
  title     = {{LAION-5B}: An Open Large-Scale Dataset for Training Next Generation Image-Text Models},
  author    = {Schuhmann, Christoph and Beaumont, Romain and Vencu, Richard and Gordon, Cade and Wightman, Ross and Cherti, Mehdi and Coombes, Theo and Katta, Aarush and Mullis, Clayton and Wortsman, Mitchell and Schramowski, Patrick and Kundurthy, Srivatsa and Crowson, Katherine and Schmidt, Ludwig and Kaczmarczyk, Robert and Jitsev, Jenia},
  booktitle = {Advances in Neural Information Processing Systems (NeurIPS) Datasets \& Benchmarks Track},
  year      = {2022},
  eprint    = {2210.08402},
  archivePrefix = {arXiv},
  primaryClass  = {cs.CV}
}

@inproceedings{lin2024genai,
  title     = {Evaluating Text-to-Visual Generation with Image-to-Text Generation},
  author    = {Lin, Zhiqiu and Pathak, Deepak and Li, Baiqi and Li, Jiayao and Xia, Xide and Neubig, Graham and Zhang, Pengchuan and Ramanan, Deva},
  booktitle = {European Conference on Computer Vision (ECCV)},
  year      = {2024},
  eprint    = {2404.01291},
  archivePrefix = {arXiv},
  primaryClass  = {cs.CV}
}

@inproceedings{longt2ibench,
  title     = {{LongT2IBench}: A Benchmark for Evaluating Long Text-to-Image Generation with Graph-structured Annotations},
  author    = {Yang, Zhichao and Gu, Tianjiao and Wang, Jianjie and Lin, Feiyu and others},
  booktitle = {AAAI Conference on Artificial Intelligence},
  year      = {2026},
  eprint    = {2512.09271},
  archivePrefix = {arXiv},
  primaryClass  = {cs.CV}
}

@inproceedings{m3id,
  title     = {Multi-Modal Hallucination Control by Visual Information Grounding},
  author    = {Favero, Alessandro and Zancato, Luca and Trager, Matthew and Choudhary, Siddharth and Perera, Pramuditha and Achille, Alessandro and Swaminathan, Ashwin and Soatto, Stefano},
  booktitle = {Conference on Computer Vision and Pattern Recognition (CVPR)},
  year      = {2024},
  eprint    = {2403.14003},
  archivePrefix = {arXiv},
  primaryClass  = {cs.CV}
}

@article{chen2025inputside,
  title       = {Improving Text-to-Image Generation with Input-Side Inference-Time Scaling},
  author      = {Chen, Ruibo and Pan, Jiacheng and Huang, Heng and Yang, Zhenheng},
  journal     = {arXiv preprint arXiv:2510.12041},
  year        = {2025},
  eprint      = {2510.12041},
  archivePrefix = {arXiv},
  primaryClass  = {cs.CV}
}

@article{mei2024scaling,
  title   = {Scaling Laws for Diffusion Transformers},
  author  = {Zhengyang Liang and Hao He and Ceyuan Yang and Bo Dai},
  journal = {arXiv preprint arXiv:2410.08184},
  year    = {2024},
  eprint  = {2410.08184},
  archivePrefix = {arXiv},
  primaryClass  = {cs.CV}
}

@misc{openai2025gptimage1,
  title        = {{GPT-Image-1}},
  author       = {{OpenAI}},
  year         = {2025},
  howpublished = {\url{https://openai.com/index/image-generation-api/}},
  note         = {Model release}
}

@inproceedings{peebles2023dit,
  title     = {Scalable Diffusion Models with Transformers},
  author    = {Peebles, William and Xie, Saining},
  booktitle = {International Conference on Computer Vision (ICCV)},
  year      = {2023},
  eprint    = {2212.09748},
  archivePrefix = {arXiv},
  primaryClass  = {cs.CV}
}

@article{qwenimage,
  title   = {{Qwen-Image} Technical Report},
  author  = {Wu, Chenfei and Li, Jiahao and Zhou, Jingren and Lin, Junyang and Gao, Kaiyuan and
             Yan, Kun and Yin, Sheng-ming and Bai, Shuai and Xu, Xiao and others},
  journal = {arXiv preprint arXiv:2508.02324},
  year    = {2025},
  eprint  = {2508.02324},
  archivePrefix = {arXiv},
  primaryClass  = {cs.CV}
}

@inproceedings{rombach2022ldm,
  title     = {High-Resolution Image Synthesis with Latent Diffusion Models},
  author    = {Rombach, Robin and Blattmann, Andreas and Lorenz, Dominik and Esser, Patrick and Ommer, Bj{\"o}rn},
  booktitle = {Conference on Computer Vision and Pattern Recognition (CVPR)},
  year      = {2022},
  eprint    = {2112.10752},
  archivePrefix = {arXiv},
  primaryClass  = {cs.CV}
}

@article{skorokhodov2025diffusability,
  title   = {Improving the Diffusability of Autoencoders},
  author  = {Skorokhodov, Ivan and Girish, Sharath and Hu, Benran and Menapace, Willi and Li, Yanyu and Abdal, Rameen and Tulyakov, Sergey and Siarohin, Aliaksandr},
  journal = {arXiv preprint arXiv:2502.14831},
  year    = {2025},
  eprint  = {2502.14831},
  archivePrefix = {arXiv},
  primaryClass  = {cs.CV}
}

@inproceedings{saharia2022imagen,
  title     = {Photorealistic Text-to-Image Diffusion Models with Deep Language Understanding},
  author    = {Saharia, Chitwan and Chan, William and Saxena, Saurabh and Li, Lala and Whang, Jay and Denton, Emily L. and Seyed Ghasemipour, Seyed Kamyar and Karagol Ayan, Burcu and Mahdavi, S. Sara and
               Gontijo Lopes, Raphael and Salimans, Tim and Ho, Jonathan and Fleet, David J. and Norouzi, Mohammad},
  booktitle = {Advances in Neural Information Processing Systems (NeurIPS)},
  year      = {2022},
  eprint    = {2205.11487},
  archivePrefix = {arXiv},
  primaryClass  = {cs.CV}
}

@article{sam21,
  title   = {{SAM 2}: Segment Anything in Images and Videos},
  author  = {Ravi, Nikhila and Gabeur, Valentin and Hu, Yuan-Ting and Hu, Ronghang and Ryali, Chaitanya and Ma, Tengyu and Khedr, Haitham and R{\"a}dle, Roman and Rolland, Chloe and Gustafson, Laura and Mintun, Eric and Pan, Junting and Alwala, Kalyan Vasudev and Carion, Nicolas and Wu, Chao-Yuan and Girshick, Ross and Doll{\'a}r, Piotr and Feichtenhofer, Christoph},
  journal = {arXiv preprint arXiv:2408.00714},
  year    = {2024},
  eprint  = {2408.00714},
  archivePrefix = {arXiv},
  primaryClass  = {cs.CV}
}

@inproceedings{sapiens,
  title     = {Sapiens: Foundation for Human Vision Models},
  author    = {Khirodkar, Rawal and Bagautdinov, Timur and Martinez, Julieta and Zhaoen, Su and James, Austin and Selednik, Peter and Anderson, Stuart and Saito, Shunsuke},
  booktitle = {European Conference on Computer Vision (ECCV)},
  year      = {2024},
  eprint    = {2408.12569},
  archivePrefix = {arXiv},
  primaryClass  = {cs.CV}
}

@article{sdxl,
  title   = {{SDXL}: Improving Latent Diffusion Models for High-Resolution Image Synthesis},
  author  = {Podell, Dustin and English, Zion and Lacey, Kyle and Blattmann, Andreas and Dockhorn, Tim and M{\"u}ller, Jonas and Penna, Joe and Rombach, Robin},
  journal = {arXiv preprint arXiv:2307.01952},
  year    = {2023},
  eprint  = {2307.01952},
  archivePrefix = {arXiv},
  primaryClass  = {cs.CV}
}

@inproceedings{song2021scorebased,
  title     = {Score-Based Generative Modeling through Stochastic Differential Equations},
  author    = {Song, Yang and Sohl-Dickstein, Jascha and Kingma, Diederik P. and Kumar, Abhishek and Ermon, Stefano and Poole, Ben},
  booktitle = {International Conference on Learning Representations (ICLR)},
  year      = {2021},
  eprint    = {2011.13456},
  archivePrefix = {arXiv},
  primaryClass  = {cs.LG}
}

@article{tiif,
  title   = {{TIIF-Bench}: How Does Your {T2I} Model Follow Your Instructions?},
  author  = {Wei, Xinyu and Zhang, Jinrui and Wang, Zeqing and Wei, Hongyang and Guo, Zhen and Li, Bairui and Zhang, Lei},
  journal = {arXiv preprint arXiv:2506.02161},
  year    = {2025},
  eprint  = {2506.02161},
  archivePrefix = {arXiv},
  primaryClass  = {cs.CV}
}

@book{vanrijsbergen1979,
  title     = {Information Retrieval},
  author    = {van Rijsbergen, C. J.},
  edition   = {2nd},
  publisher = {Butterworths},
  address   = {London},
  year      = {1979}
}

@inproceedings{wallace2024diffusiondpo,
  title     = {Diffusion Model Alignment Using Direct Preference Optimization},
  author    = {Wallace, Bram and Dang, Meihua and Rafailov, Rafael and Zhou, Linqi and Lou, Aaron and Purushwalkam, Senthil and Ermon, Stefano and Xiong, Caiming and Joty, Shafiq and Naik, Nikhil},
  booktitle = {Conference on Computer Vision and Pattern Recognition (CVPR)},
  year      = {2024},
  eprint    = {2311.12908},
  archivePrefix = {arXiv},
  primaryClass  = {cs.CV}
}

@article{wang2025promptenhancer,
  title   = {{PromptEnhancer}: A Simple Approach to Enhance Text-to-Image Models via Chain-of-Thought Prompt Rewriting},
  author  = {Wang, Linqing and Xing, Ximing and Cheng, Yiji and Zhao, Zhiyuan and Li, Donghao and
             Hang, Tiankai and Tao, Jiale and Wang, Qixun and Li, Ruihuang and others},
  journal = {arXiv preprint arXiv:2509.04545},
  year    = {2025},
  eprint  = {2509.04545},
  archivePrefix = {arXiv},
  primaryClass  = {cs.CV}
}

@article{wu2025omnigen2,
  title   = {{OmniGen2}: Towards Instruction-Aligned Multimodal Generation},
  author  = {Wu, Chenyuan and Zheng, Pengfei and Yan, Ruiran and Xiao, Shitao and Luo, Xin and
             Wang, Yueze and Li, Wanli and Jiang, Xiyan and Liu, Yexin and others},
  journal = {arXiv preprint arXiv:2506.18871},
  year    = {2025},
  eprint  = {2506.18871},
  archivePrefix = {arXiv},
  primaryClass  = {cs.CV}
}

@inproceedings{xu2024imagereward,
  title     = {{ImageReward}: Learning and Evaluating Human Preferences for Text-to-Image Generation},
  author    = {Xu, Jiazheng and Liu, Xiao and Wu, Yuchen and Tong, Yuxuan and Li, Qinkai and Ding, Ming and Tang, Jie and Dong, Yuxiao},
  booktitle = {Advances in Neural Information Processing Systems (NeurIPS)},
  year      = {2023},
  eprint    = {2304.05977},
  archivePrefix = {arXiv},
  primaryClass  = {cs.CV}
}

@inproceedings{yang2024rpg,
  title     = {Mastering Text-to-Image Diffusion: Recaptioning, Planning, and Generating with Multimodal {LLMs}},
  author    = {Yang, Ling and Yu, Zhaochen and Meng, Chenlin and Xu, Minkai and Ermon, Stefano and Cui, Bin},
  booktitle = {International Conference on Machine Learning (ICML)},
  year      = {2024},
  eprint    = {2401.11708},
  archivePrefix = {arXiv},
  primaryClass  = {cs.CV}
}

@article{jiao2025detailmaster,
  title   = {{DetailMaster}: Can Your Text-to-Image Model Handle Long Prompts?},
  author  = {Jiao, Qirui and Chen, Daoyuan and Huang, Yilun and Lin, Xika and Shen, Ying and Li, Yaliang},
  journal = {arXiv preprint arXiv:2505.16915},
  year    = {2025},
  eprint  = {2505.16915},
  archivePrefix = {arXiv},
  primaryClass  = {cs.CV}
}

@inproceedings{zheng2024cogview3,
  title     = {{CogView3}: Finer and Faster Text-to-Image Generation via Relay Diffusion},
  author    = {Zheng, Wendi and Teng, Jiayan and Yang, Zhuoyi and Wang, Weihan and Chen, Jidong and Gu, Xiaotao and Dong, Yuxiao and Ding, Ming and Tang, Jie},
  booktitle = {European Conference on Computer Vision (ECCV)},
  year      = {2024},
  eprint    = {2403.05121},
  archivePrefix = {arXiv},
  primaryClass  = {cs.CV}
}

@inproceedings{liang2024precision,
  title     = {Precision or Recall? An Analysis of Image Captions for Training Text-to-Image Generation Model},
  author    = {Sheng Cheng and Maitreya Patel and Yezhou Yang},
  booktitle = {Findings of the Association for Computational Linguistics: EMNLP 2024},
  year      = {2024},
  eprint    = {2411.05079},
  archivePrefix = {arXiv},
  primaryClass  = {cs.CV}
}

@article{segalis2023recap,
  title={A Picture is Worth a Thousand Words: Principled Recaptioning Improves Image Generation},
  author={Segalis, Eyal and Valevski, Dani and Lumen, Danny and Matias, Yossi and Leviathan, Yaniv},
  journal={arXiv preprint arXiv:2310.16656}, year={2023}
}

@article{hu2024ella,
  title={ELLA: Equip Diffusion Models with LLM for Enhanced Semantic Alignment},
  author={Hu, Xiwei and Wang, Rui and Fang, Yixiao and Fu, Bin and Cheng, Pei and Yu, Gang},
  journal={arXiv preprint arXiv:2403.05135}, year={2024}
}

@inproceedings{zhang2024longclip,
  title={Long-{CLIP}: Unlocking the Long-Text Capability of {CLIP}},
  author={Zhang, Beichen and Zhang, Pan and Dong, Xiaoyi and Zang, Yuhang and Wang, Jiaqi},
  booktitle={European Conference on Computer Vision (ECCV)}, year={2024}
}

@inproceedings{urbanek2024dci,
  title={A Picture is Worth More Than 77 Text Tokens: Evaluating {CLIP}-Style Models on Dense Captions},
  author={Urbanek, Jack and Bordes, Florian and Astolfi, Pietro and Williamson, Mary and Sharma, Vasu and Romero-Soriano, Adriana},
  booktitle={Conference on Computer Vision and Pattern Recognition (CVPR)}, year={2024}
}

@inproceedings{onoe2024docci,
  title={{DOCCI}: Descriptions of Connected and Contrasting Images},
  author={Onoe, Yasumasa and Rane, Sunayana and Berger, Zachary and Bitton, Yonatan and Cho, Jaemin and Garg, Roopal and Ku, Alexander and Parekh, Zarana and Pont-Tuset, Jordi and Tanzer, Garrett and Wang, Su and Baldridge, Jason},
  booktitle={European Conference on Computer Vision (ECCV)}, year={2024}
}

@article{merchant2025structuredcaptions,
  title={Structured Captions Improve Prompt Adherence in Text-to-Image Models (Re-LAION-Caption 19M)},
  author={Merchant, Nicholas and Borde, Haitz S{\'a}ez de Oc{\'a}riz and Popescu, Andrei Cristian and Suarez, Carlos Garcia Jurado},
  journal={arXiv preprint arXiv:2507.05300}, year={2025}
}

@article{gutflaish2025fibo,
  title={{Generating an Image From 1,000 Words: Enhancing Text-to-Image With Structured Captions}},
  author={Gutflaish, Eyal and Kachlon, Eliran and Zisman, Hezi and Hacham, Tal and Sarid, Nimrod and Visheratin, Alexander and Huberman, Saar and Davidi, Gal and Bukchin, Guy and Goldberg, Kfir and Mokady, Ron},
  journal={arXiv preprint arXiv:2511.06876},
  year={2025}
}

@article{nvidia2026cosmos3,
  title={{Cosmos 3: Omnimodal World Models for Physical AI}},
  author={{NVIDIA}},
  journal={arXiv preprint arXiv:2606.02800},
  year={2026},
  url={https://research.nvidia.com/labs/cosmos-lab/cosmos3/technical-report.pdf}
}

@misc{reve2026layoutbet,
  title={{The Layout Bet}},
  author={{Reve Team}},
  year={2026},
  month={June},
  url={https://blog.reve.com/posts/the-layout-bet/}
}

@article{longcatimage2025,
  title={{LongCat-Image} Technical Report},
  author={{Meituan LongCat Team} and Ma, Hanghang and Tan, Haoxian and Huang, Jiale and
          Wu, Junqiang and He, Jun-Yan and Gao, Lishuai and Xiao, Songlin and Wei, Xiaoming and
          Ma, Xiaoqi and Cai, Xunliang and Guan, Yayong and Hu, Jie},
  journal={arXiv preprint arXiv:2512.07584},
  year={2025}
}

@article{hunyuanimage2025,
  title={{HunyuanImage 3.0} Technical Report},
  author={{Tencent Hunyuan Foundation Model Team}},
  journal={arXiv preprint arXiv:2509.23951},
  year={2025}
}

@article{hu2021lora,
  title   = {{LoRA}: Low-Rank Adaptation of Large Language Models},
  author  = {Hu, Edward J. and Shen, Yelong and Wallis, Phillip and Allen-Zhu, Zeyuan and Li, Yuanzhi and Wang, Shean and Wang, Lu and Chen, Weizhu},
  journal = {arXiv preprint arXiv:2106.09685},
  year    = {2021}
}

@article{ye2025echo4o,
  title   = {{Echo-4o}: Harnessing the Power of {GPT-4o} Synthetic Images for Improved Image Generation},
  author  = {Ye, Junyan and Jiang, Dongzhi and Wang, Zihao and Zhu, Leqi and Hu, Zhenghao and Huang, Zilong and He, Jun and Yan, Zhiyuan and Yu, Jinghua and Li, Hongsheng and He, Conghui and Li, Weijia},
  journal = {arXiv preprint arXiv:2508.09987},
  year    = {2025}
}

@article{niu2025wise,
  title   = {{WISE}: A World Knowledge-Informed Semantic Evaluation for Text-to-Image Generation},
  author  = {Niu, Yuwei and Ning, Munan and Zheng, Mengren and Jin, Weiyang and Lin, Bin and Jin, Peng and Liao, Jiaqi and Feng, Chaoran and Meng, Fanqing and Ning, Kunpeng and Zhu, Bin and Yuan, Li},
  journal = {arXiv preprint arXiv:2503.07265},
  year    = {2025}
}

@article{shao2024deepseekmath,
  title   = {{DeepSeekMath}: Pushing the Limits of Mathematical Reasoning in Open Language Models},
  author  = {Shao, Zhihong and Wang, Peiyi and Zhu, Qihao and Xu, Runxin and Song, Junxiao and Bi, Xiao and Zhang, Haowei and Zhang, Mingchuan and Li, Y. K. and Wu, Y. and Guo, Daya},
  journal = {arXiv preprint arXiv:2402.03300},
  year    = {2024}
}

@inproceedings{ghosh2023geneval,
  title     = {{GenEval}: An Object-Focused Framework for Evaluating Text-to-Image Alignment},
  author    = {Ghosh, Dhruba and Hajishirzi, Hannaneh and Schmidt, Ludwig},
  booktitle = {Advances in Neural Information Processing Systems, Datasets and Benchmarks Track},
  year      = {2023},
  eprint    = {2310.11513},
  archivePrefix = {arXiv},
  primaryClass = {cs.CV}
}

@article{kamath2025geneval2,
  title   = {{GenEval 2}: Addressing Benchmark Drift in Text-to-Image Evaluation},
  author  = {Kamath, Amita and Chang, Kai-Wei and Krishna, Ranjay and Zettlemoyer, Luke and Hu, Yushi and Ghazvininejad, Marjan},
  journal = {arXiv preprint arXiv:2512.16853},
  year    = {2025},
  eprint  = {2512.16853},
  archivePrefix = {arXiv},
  primaryClass = {cs.CV}
}

@inproceedings{li2026corebench,
  title     = {Easier Painting Than Thinking: Can Text-to-Image Models Set the Stage, but Not Direct the Play?},
  author    = {Li, Ouxiang and Wang, Yuan and Hu, Xinting and Huang, Huijuan and Chen, Rui and Ou, Jiarong and Tao, Xin and Wan, Pengfei and Qi, Xiaojuan and Feng, Fuli},
  booktitle = {International Conference on Learning Representations},
  year      = {2026},
  eprint    = {2509.03516},
  archivePrefix = {arXiv},
  primaryClass = {cs.CV}
}

@inproceedings{guo2025comfymind,
  title     = {{ComfyMind}: Toward General-Purpose Generation via Tree-Based Planning and Reactive Feedback},
  author    = {Guo, Litao and Xu, Xinli and Wang, Luozhou and Lin, Jiantao and Zhou, Jinsong and Zhang, Zixin and Su, Bolan and Chen, Ying-Cong},
  booktitle = {Advances in Neural Information Processing Systems},
  year      = {2025},
  url       = {https://papers.nips.cc/paper_files/paper/2025/file/40168e00bf87869c5d153e934d8a3602-Paper-Conference.pdf}
}

@article{he2026gems,
  title         = {{GEMS}: Agent-Native Multimodal Generation with Memory and Skills},
  author        = {He, Zefeng and Huang, Siyuan and Qu, Xiaoye and Li, Yafu and Zhu, Tong and Cheng, Yu and Yang, Yang},
  journal       = {arXiv preprint arXiv:2603.28088},
  year          = {2026},
  eprint        = {2603.28088},
  archivePrefix = {arXiv},
  primaryClass  = {cs.CV}
}

@misc{qwen35report,
  title        = {Qwen3.5},
  author       = {{Qwen Team}},
  year         = {2025},
  howpublished = {\url{https://qwen.ai/blog?id=qwen3.5}},
  note         = {Blog post}
}

\clearpage
\appendix
\etocdepthtag.toc{appendix}
\begin{center}
  {\Large\bfseries Appendix}
\end{center}
\vspace{0.8em}

\etocsettagdepth{mainbody}{none}
\etocsettagdepth{appendix}{subsection}
\renewcommand{\contentsname}{Contents}
\etocsetnexttocdepth{subsection}
\tableofcontents
\vspace{1.2em}

\section{Grounded Perplexity Gain: measurement details}
\label{app:gpg_details}

\subsection{Default scoring protocol}
\label{app:gpg_protocol}

All headline GPG scores use Qwen3.5-397B-A17B frozen as the VLM judge.
Images are resized so that the longer side is $1024$ pixels with aspect ratio preserved.
Before either pass, captions are mapped to a deterministic scored sequence.
For SPs, canonicalization removes the global metadata keys \texttt{atmosphere}, \texttt{lighting}, \texttt{style}, and \texttt{photography}, including its \texttt{layout}, \texttt{shot\_type}, \texttt{camera\_angle}, and \texttt{lens\_and\_effect} subfields.
NL captions retain their text, with template-only spans marked separately.
The image-conditioned and no-image passes then use the same conversation template and canonical sequence; the latter replaces the image with an empty image slot.
All tokens in that sequence remain in the autoregressive context in both passes. The content mask excludes JSON syntax, stylistic boilerplate, and NL template-only positions only from the accumulated GPG sum.
The judge, preprocessing, and template are fixed across all caption conditions, so comparisons vary only the caption content presented for scoring.
The canonicalization and content-mask recipe was finalized during metric development on the controlled BAGEL sweep and then frozen. Consequently, the headline GPG--loss relation is a calibration for this fixed recipe, not an independent validation of these scoring choices; applying it to new caption families requires retaining the same recipe or recalibrating it once.

\subsection{GPG robustness across judges}
\label{app:judge}

\begin{figure}[t]
  \centering
  \includegraphics[width=\linewidth]{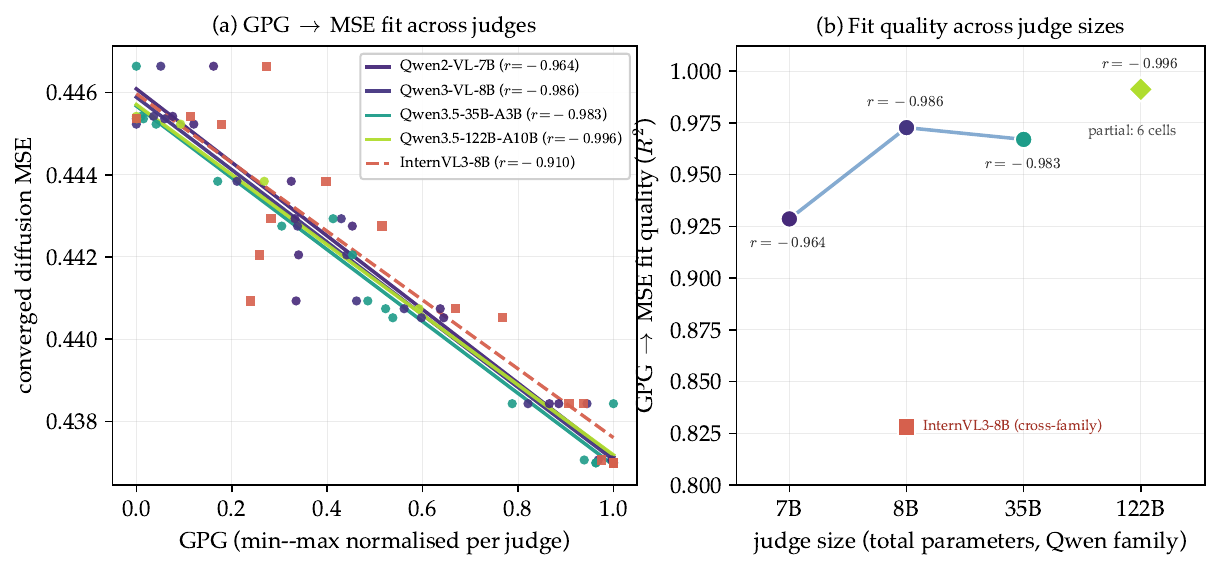}
  \caption{\textbf{GPG is robust across compatible judges.}
  \emph{(a)}~GPG\,$\to$\,MSE fits with GPG min--max normalized per judge.
  \emph{(b)}~Fit quality versus judge size; the 122B point uses only six cells and is shown separately from the full-cell fits.
  InternVL3-8B is the cross-family check.}
  \label{fig:gpg_judge_robustness}
\end{figure}

GPG depends on the complete judge interface---the VLM, tokenizer, conversation template, image processor, and coordinate convention---rather than on the likelihood query alone; we hold that interface fixed for every headline comparison.
Our default judge is Qwen3.5-397B-A17B, chosen because its normalized bbox convention is directly compatible with the caption format's $0$--$999$ coordinates, avoiding an additional coordinate conversion.
To verify that the linear scaling property (Eq.~\eqref{eq:scaling_law}) is not specific to this judge, we recompute GPG on a held-out pool of $1{,}000$ paired images across the scaling-property cells under six additional VL judges---spanning the Qwen2-VL / Qwen2.5-VL / Qwen3-VL / Qwen3.5-MoE lineage plus a cross-family judge (InternVL3-8B)---and refit the linear relation on each.
Table~\ref{tab:judge_selection} reports the resulting cross-judge comparison.
Every judge that grounds the caption's spatial vocabulary gives a strong negative fit ($r$ from $-0.91$ to $-0.99$, within $\pm 0.08$ of the main fit's $-0.984$), and their per-cell GPG rankings agree at Spearman $\rho \!\geq\! 0.86$ (identical top and bottom cells).
Across the fully evaluated Qwen-family judges, $R^2$ remains between $0.93$ and $0.97$; the 122B judge reaches $0.99$ on the six available cells, but this partial-cell fit is not directly comparable to the full-cell fits (Figure~\ref{fig:gpg_judge_robustness}).
The robustness conclusion is therefore agreement across judges supplied with a compatible coordinate representation, not a monotonic law in judge size; we use the largest available judge for the headline measurement and compare its ordering with smaller and cross-family alternatives.
The lone exception is instructive rather than a counterexample: without a bbox-format adapter, Qwen2.5-VL-7B cannot parse our $0$--$999$ \texttt{<bbox>} tokens and \emph{inverts} the ordering ($r\!=\!+0.23$); supplying the adapter (\texttt{--qwen-native-bbox}, which maps to its native pixel-space coordinates) restores $r\!=\!-0.97$.
GPG therefore remains judge-dependent, but its configuration ordering is stable across the tested judges once each can read the caption's spatial representation.

\begin{table}[h]
  \centering
  \caption{\textbf{GPG fit quality across alternative judges.}
  We refit the GPG\,$\to$\,MSE relation under each judge on a held-out $1{,}000$-image robustness pool; the selected main judge (bold) uses the full $30$k-UID pool.
  Qwen2.5-VL-7B reports raw\,/\,bbox-adapted scores; see text for the inverted raw case.}
  \label{tab:judge_selection}
  \small
  \begin{tabularx}{\linewidth}{@{}llrr>{\raggedright\arraybackslash}X@{}}
    \toprule
    Judge model & Size & Pearson $r$ & $R^2$ & Notes \\
    \midrule
    \multicolumn{5}{@{}l}{\textcolor{gray}{\emph{Robustness sweep (held-out $1{,}000$-image pool)}}}\\
    Qwen2-VL-7B & 7B & $-0.96$ & $0.93$ & Oldest Qwen-VL; narrower GPG range \\
    Qwen3-VL-8B & 8B & $-0.99$ & $0.97$ & Normalized bbox convention compatible with caption format \\
    Qwen3.5-35B-A3B & 35B & $-0.98$ & $0.97$ & \texttt{no\_bbox} cell is $3.3$ nats higher \\
    Qwen3.5-122B-A10B & 122B & $-0.996$ & $0.99$ & Partial fit ($6$ of $15$ cells); not directly comparable \\
    InternVL3-8B & 8B & $-0.91$ & $0.83$ & Cross-family; confirms not Qwen-specific \\
    Qwen2.5-VL-7B & 7B & $+0.23$\,/\,$-0.97$ & $0.05$\,/\,$0.94$ & Raw\,/\,with \texttt{--qwen-native-bbox} (see text) \\
    \midrule
    \textbf{Qwen3.5-397B-A17B} & \textbf{397B} & $\mathbf{-0.98}$ & $\mathbf{0.97}$ & Selected (main); full 30k-UID pool \\
    \bottomrule
  \end{tabularx}
\end{table}

\subsection{Full 15-setting data}
\label{app:all_cells}

Table~\ref{tab:all_cells} reports the complete per-setting GPG, ED, and converged diffusion loss underlying Eqs.~\eqref{eq:scaling_law} and~\eqref{eq:scaling_law_ed}.
All settings share the same $30{,}000$ paired image UIDs for both caption-side measurements.
GPG uses Qwen3.5-397B-A17B with the content-mask and canonicalize-JSON recipe of Appendix~\ref{app:gpg_protocol}, while ED follows the extraction and matching protocol of Appendix~\ref{app:ed_validation}; MSE is measured at the unified budget of $2.84{\times}10^{10}$ cumulative image tokens reached by every run.

\begin{table}[h]
  \centering
  \caption{\textbf{Full 15-setting scaling-property data}, sorted by GPG. The full-schema baseline of the ablation suite coincides with Structured~L10 and is counted once.}
  \label{tab:all_cells}
  \small
  \begin{tabular}{@{}llrrr@{}}
    \toprule
    Kind & Level & GPG & ED & MSE \\
    \midrule
    Dense (NL) & L6 & 106.3 & 0.759 & 0.44523\\
    Dense (NL) & L8 & 110.1 & 0.751 & 0.44542\\
    Structured & L5 & 111.6 & 0.749 & 0.44664\\
    Dense (NL) & L10 & 112.5 & 0.754 & 0.44536\\
    Structured & L6 & 128.3 & 0.759 & 0.44384\\
    Structured & L7 & 141.5 & 0.772 & 0.44275\\
    Spatial (coarse, $3{\times}3$) & L10 & 151.0 & 0.778 & 0.44293\\
    Spatial (fine, $5{\times}5$) & L10 & 152.2 & 0.785 & 0.44205\\
    Spatial (finer, $9{\times}9$) & L10 & 154.4 & 0.787 & 0.44093\\
    Structured & L8 & 164.8 & 0.793 & 0.44074\\
    Abl: $-$scene & L10 & 168.5 & 0.799 & 0.44052\\
    Structured & L9 & 191.8 & 0.819 & 0.43843\\
    Abl: $-$bbox & L10 & 204.7 & 0.807 & 0.43843\\
    Abl: $-$relationships & L10 & 207.2 & 0.808 & 0.43706\\
    Structured & L10 & 210.5 & 0.833 & 0.43699\\
    \bottomrule
  \end{tabular}
\end{table}

\subsection{Monotonicity analysis}
\label{app:monotonicity}

Sorted by GPG, MSE decreases overall across the 15 settings, with a few small local reversals.
Their total magnitude is approximately $0.0016$ MSE units, of which approximately $0.0012$ comes from structured L5 relative to the nearby NL settings; we report this reversal descriptively rather than assigning it to a specific cause.
The remaining violations are $|\Delta\text{MSE}| \!\leq\! 2{\times}10^{-4}$, the same order as trailing-window read-out variation, so we do not assign them to a specific cause.

\subsection{Fit sensitivity across settings and training budgets}
\label{app:scaling_supplement}

\begin{figure}[h]
  \centering
  \includegraphics[width=0.5\linewidth]{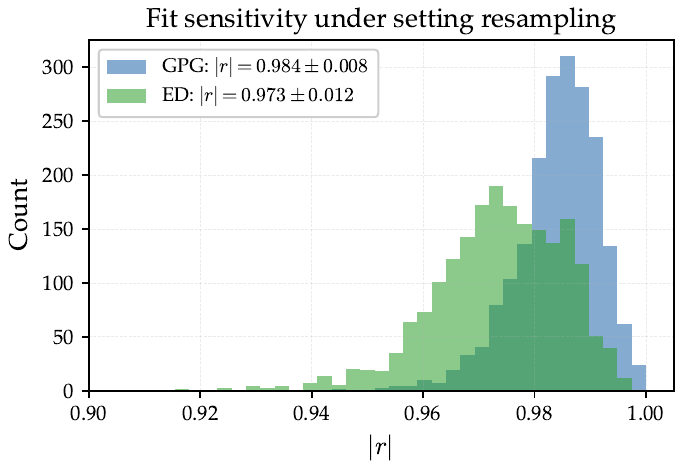}
  \caption{\textbf{Sensitivity of the scaling-property fit to resampling sweep settings.}
  Over $B\!=\!2{,}000$ resamples of the $15$ designed settings, the distributions of fit magnitude have mean$\pm$SD $|r|\!=\!0.984\pm0.008$ for GPG and $|r|\!=\!0.973\pm0.012$ for ED.}
  \label{fig:scaling_supplement}
\end{figure}

Figure~\ref{fig:scaling_supplement} shows how the fit correlations change when the $15$ designed sweep settings are resampled; the per-level training dynamics for both caption families appear in the main text (Figure~\ref{fig:scaling_overview}a--b).

\paragraph{Uncertainty scope.}
Each setting is trained once.
The analyses below quantify three operational sensitivities: resampling the deliberately designed sweep settings, refitting the same training trajectories at matched budget cuts, and measuring local variation within their trailing windows.
They are not population confidence intervals, independent training replications, or estimates of seed-level optimization uncertainty.

\begin{figure*}[h!]
  \centering
  \includegraphics[width=0.98\textwidth]{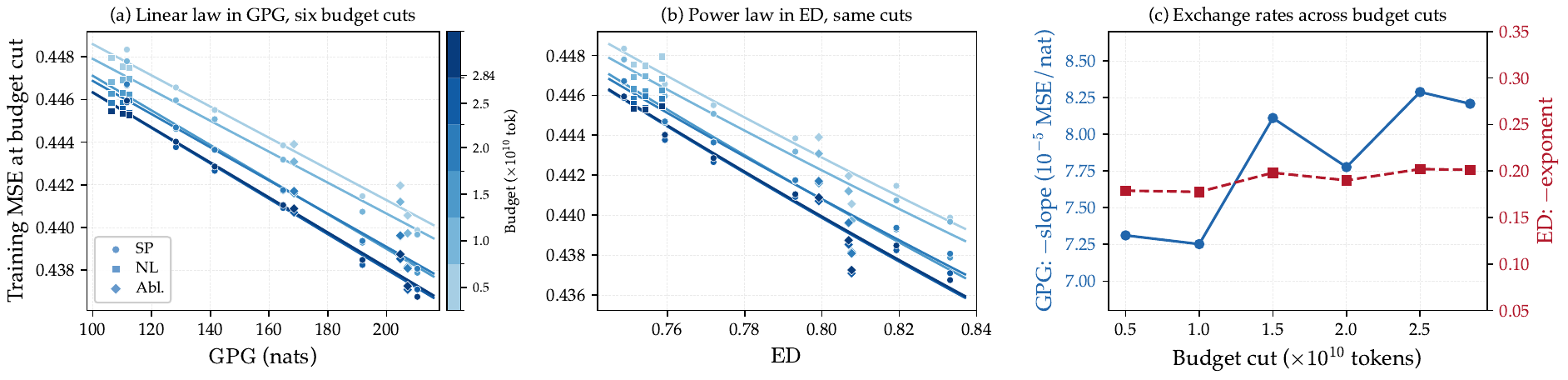}
  \caption{\textbf{The scaling properties persist across training budgets.}
  Refitting the same trajectories at six matched budget cuts preserves both the linear GPG relation and the ED power law; the fitted slope and exponent steepen mildly as training proceeds.}
  \label{fig:law_budget}
\end{figure*}

\paragraph{Read-out conventions for the budget refits.}
The budget refits of Figure~\ref{fig:law_budget} use the twelve settings whose full training curves were exported: six structured levels, three natural-language levels, and the three field ablations with measured GPG and ED; the three spatial variants were logged only at the final budget.
The read-out at each cut is the mean training MSE over the trailing $5\%$ of tokens before the cut; the two runs that restarted mid-training (L5, L9) have their token axes corrected for the restart's token-counter reset.
GPG and ED are caption-side quantities and do not vary with budget; at each cut the GPG relation is fit linearly and the ED relation in log--log space, mirroring Eqs.~\eqref{eq:scaling_law} and~\eqref{eq:scaling_law_ed}.
The curve-based read-out differs slightly from the per-setting converged values behind the headline fits, and the final-cut coefficients remain close to them.

\paragraph{Within-trajectory read-out variation.}
Splitting each observed trajectory's trailing window into five disjoint blocks, the standard deviation of block means spans $1.0$--$5.0\times10^{-4}$ across the twelve settings (median $3.5\times10^{-4}$).
This scale is comparable to the residuals of both fits: $\sigma_{\mathrm{resid}} \approx 6\times10^{-4}$ for GPG and $7.8\times10^{-4}$ for ED.
These values contextualize the residual scale but do not decompose it into lack-of-fit and run-level optimization variability.

\subsection{Mutual-information motivation and the empirical status of the relations}
\label{app:why_linear}

For image and caption random variables $I$ and $Y$, the identity
$H(I \mid Y)=H(I)-I(I;Y)$ states that caption--image mutual information reduces uncertainty about the image.
This intuition motivates asking whether captions with more image-grounded information are associated with a lower conditional training objective.
It does not, however, derive the form of that association in our experiments.
GPG is an operational, judge-dependent estimate rather than the dataset mutual information; the converged flow-matching velocity MSE is not measured in nats; and the entropy identity alone neither makes that objective a conditional-likelihood bound nor implies a linear GPG--MSE relation.

Equation~\eqref{eq:scaling_law} is therefore an empirical, recipe-specific calibration rather than a theorem.
Its fitted slope has no universal information-theoretic interpretation: it depends on the judge, caption construction, model, objective, optimization, and training budget held fixed in the sweep.
The empirical result is instead that GPG measured by a frozen VLM tracks the converged loss of a separately trained diffuser across the tested caption settings.
ED provides a complementary check: despite using no token probabilities, it closely agrees with the GPG ordering ($\rho_{\text{Spearman}}\!=\!0.96$) and follows its own negative power-law trend (Eq.~\eqref{eq:scaling_law_ed}).
Residuals from either fit should likewise be read as deviations from this calibration, not as direct estimates of information that the diffuser fails to exploit.

\section{Effective Detailness: validation details}
\label{app:ed_validation}

\subsection{Image-grounded source extraction}
\label{app:ed_source}

The image-grounded source set for ED is obtained by one offline Gemini 3 Pro call per image, prompted to enumerate object--attribute--relationship--grounding (OARG) tuples.
The extractor sees only the image and is asked for $80$--$150$ atomic tuples; it returns a mean of $102$ tuples over the shared $30{,}000$-image ED pool.
For each caption condition, GPT-5.4 independently extracts caption-side OARG tuples, and a separate GPT-5.4 call performs symmetric paraphrase-tolerant matching against the cached source tuples.
For each valid image--caption pair, $P_A$ is the fraction of caption-side attribute tuples supported by the image source, and $R_A$ is the fraction of image-side attribute tuples covered by the caption. The pair-level score is $F_{0.5}(P_A,R_A)$; pairs for which either side contains no attribute tuple are skipped, and the degenerate $P_A=R_A=0$ case is assigned zero. A caption configuration's ED is the two-sided $10\%$ trimmed mean of its pair-level scores: after sorting, we drop $\lfloor0.1n\rfloor$ values from each tail and average the remainder.
Only attributes enter the reported score; object, relation, and grounding tuples provide entity context for extraction and matching.
All image-side sources, caption-side tuples, and match masks are computed once and cached.

\subsection{Why precision-weighted $F_{0.5}$?}
\label{app:ed_beta}

The two ED errors have different consequences for conditioning: an unsupported caption attribute supplies contradictory supervision for its paired image, whereas an omitted attribute reduces conditioning bandwidth without introducing a false visual fact.
This asymmetry motivates the standard precision-oriented $F_{0.5}$ convention~\citep{vanrijsbergen1979,liang2024precision}.
We selected the precision-oriented aggregation over $F_1$ and $P\cdot R$ during metric development on the controlled BAGEL sweep, then froze $\beta=0.5$ for every caption family and subsequent analysis. As with the GPG scoring recipe of Appendix~\ref{app:gpg_protocol}, the headline ED--loss relation is therefore a calibration for this fixed choice rather than an independent validation of it.
The sensitivity study below varies $\beta$ post hoc to test how strongly the empirical relation depends on this precision-oriented choice.

\begin{table}[t]
  \centering
  \caption{\textbf{Sensitivity to the precision--recall weighting in ED} on $30{,}000$ paired images per caption configuration. We replace $F_{0.5}$ by $F_1$ or $F_2$ while keeping the tuple extraction, matching, aggregation, caption configurations, and converged-loss read-out fixed. $r$ and $R^2$ are measured in log--log space; variant MAE fits the three NL and six nested-SP configurations and evaluates the six spatial and field variants excluded from the fit.}
  \label{tab:ed_beta_sensitivity}
  \small
  \begin{tabular}{@{}lcrrrrr@{}}
    \toprule
    $\beta$ & Emphasis & exponent $b$ & Pearson $r$ & $R^2$ & Spearman $\rho$ & variant MAE \\
    \midrule
    $0.5$ & precision & $-0.207$ & $-0.971$ & $0.943$ & $-0.993$ & $0.00081$ \\
    $1.0$ & balanced  & $-0.098$ & $-0.736$ & $0.541$ & $-0.729$ & $0.00153$ \\
    $2.0$ & recall    & $-0.050$ & $-0.559$ & $0.313$ & $-0.607$ & $0.00191$ \\
    \bottomrule
  \end{tabular}
\end{table}
The fitted log--log slope remains negative as $\beta$ increases, and the nested SP ladder remains monotonic from L5 to L10 under all three settings. Predictive strength nevertheless falls substantially: $R^2$ decreases from $0.943$ for $F_{0.5}$ to $0.313$ for $F_2$, while variant MAE more than doubles. Thus the direction of the relation is robust to $\beta$, but the precision-oriented score is markedly more predictive.

\subsection{Shared-UID resampling sensitivity}
\label{app:ed_hp}

\paragraph{Conditional resampling analysis.}
A shared-UID bootstrap over the $30{,}000$ paired images ($B\!=\!10{,}000$) gives a resampling-distribution mean$\pm$SD Spearman $\rho=-0.979\pm0.010$ between $\mathrm{ED}$ and converged MSE, with central $95\%$ range $[-0.993,-0.957]$.
The corresponding ranges are $[-0.978,-0.946]$ for the log--log Pearson correlation and $[-0.220,-0.191]$ for the exponent.
These ranges condition on the cached outputs of the image proposer, caption extractor, and matcher; they measure sensitivity to which paired UIDs are included, but do not include API/model stochasticity or training-run uncertainty.

\subsection{Robustness across source extractors}
\label{app:ed_extractors}

ED's only image-conditioned step is the offline extraction of attribute tuples from each image.
A natural concern is that the resulting per-cell rankings might be an artifact of the specific extractor model.
To test this, we re-run the image-side source extraction with two additional vision--language backends---GPT-4o and GPT-5.5 (reasoning)---alongside the default Gemini 3 Pro, each given the same exhaustive-OARG instructions, on a held-out pool of $1{,}000$ paired images (GPT-5.5 on the first $150$).
Caption extraction and symmetric matching remain fixed at GPT-5.4, so Table~\ref{tab:ed_extractors} isolates the image-source backend.
ED proves \emph{rank}-robust: under every backend ED remains a strong negative predictor of converged MSE ($\rho(\mathrm{ED},\mathrm{MSE})$ from $-0.86$ to $-0.90$), the cells keep the same ordering (cross-backend cell-rank $\rho\!=\!0.80$--$0.92$; pairwise per-tuple Cohen $\kappa\!=\!0.75$--$0.77$), and the power-law slope keeps its sign and rough magnitude.
Absolute ED levels, however, are \emph{not} backend-invariant: the intraclass correlation for absolute agreement is only $0.11$ (consistency ICC $0.64$)---different extractors place the cells on shifted ED scales even while agreeing on their order.
We therefore use ED as a \emph{relative} ruler with a single fixed extractor (Gemini 3 Pro) throughout the paper, and compare only rankings and slopes across backends---never absolute ED between extractors.

\begin{table}[h]
  \centering
  \caption{\textbf{ED backend robustness} (held-out $1{,}000$-image pool; GPT-5.5 on $N\!=\!150$).
  ED remains a negative predictor of converged MSE under every extractor, with stable cell rankings but shifted absolute scales.
  Cross-backend rank and tuple-agreement statistics are discussed in the text.}
  \label{tab:ed_extractors}
  \small
  \begin{tabular}{@{}lccc@{}}
    \toprule
    Extractor & $\rho(\mathrm{ED},\mathrm{MSE})$ & power-law slope $b$ & cell-rank $\rho$ vs Gemini \\
    \midrule
    Gemini 3 Pro (reference) & $-0.87$ & $-0.221 \pm 0.020$ & --- \\
    GPT-4o                   & $-0.86$ & $-0.244 \pm 0.027$ & $0.80$ \\
    GPT-5.5 (reasoning)      & $-0.90$ & $-0.137 \pm 0.042$ & $0.88$ \\
    \bottomrule
  \end{tabular}
\end{table}

Complete per-setting ED values appear alongside GPG and converged MSE in Table~\ref{tab:all_cells}.
They underlie the power-law fit of Eq.~\eqref{eq:scaling_law_ed}; Grid-$k$ denotes verbal locations on a $k{\times}k$ grid, and the ablation rows mask the named L10 field.

\section{Structured-prompt schema and implementation details}
\label{app:training}

The structured prompt is the central interface of this paper: its definition in \S\ref{sec:format}, its construction in \S\ref{sec:sp}, and the scaling-property evidence in \S\ref{sec:metrics} depend on both the JSON layout and how its fields are populated from images.
This appendix first summarizes the schema (\S\ref{app:schema}), then reports the annotation, diffusion, prompter, inference, and dependency details needed to implement the full system.
Training covers three model components: the BAGEL and Qwen-Image diffusion backbones, and a rank-$128$ LoRA~\citep{hu2021lora} on Qwen3.5-397B-A17B as the prompter (\S\ref{sec:prompter}, three serial stages: SFT, Cold-start, and RFT).
BAGEL is trained once per caption configuration for the scaling-property experiments of \S\ref{sec:metrics}; Qwen-Image is trained once on the structured-prompt corpus (a mixture of SP levels with NL captions) and held fixed across all prompter ablations (\S\ref{sec:ablations}).
All trainings use packed sequences, so there is no fixed per-step batch size in samples; the tables below report the packed sequence length and global GPU count from which effective tokens per step can be derived.
Stage-by-stage annotation details appear in \S\ref{app:impl_annotation}, and the degradation levels used in the scaling-property experiments are defined in Table~\ref{tab:degradation}.

\subsection{Structured-prompt schema}
\label{app:sp}
\label{app:schema}

Table~\ref{tab:schema_fields} summarizes the structured-prompt schema. Three required fields define the scene skeleton---the overall \texttt{intent}, the macro \texttt{scene}, and the list of \texttt{elements}. Five optional fields add cross-element relationships (\texttt{relationships}) and global controls (\texttt{atmosphere}, \texttt{photography}, \texttt{style}, \texttt{lighting}). The prompter additionally emits a leading \texttt{ratio} control field. At inference, the generation wrapper uses it to select the output canvas and removes it before the remaining SP is passed to the diffuser. Image-to-SP annotation omits \texttt{ratio}, since the source image already fixes the canvas. The schema is extensible: the \texttt{elements} list uses dynamic per-element keys, so attributes and actions can be added without schema changes.

\begin{table}[h]
  \centering
  \caption{\textbf{Structured-prompt schema summary.} Required fields define the scene skeleton; optional fields add cross-element relations and global controls. $^{\ast}$\texttt{ratio} is an inference-only control emitted by the prompter, consumed by the generation wrapper, and removed before diffusion conditioning.}
  \label{tab:schema_fields}
  \small
  \setlength{\tabcolsep}{5pt}
  \begin{tabularx}{\linewidth}{@{}lll>{\raggedright\arraybackslash}X@{}}
    \toprule
    Field & Req. & Type & Description \\
    \midrule
    \texttt{ratio} & req.$^{\ast}$ & string & Output aspect ratio from a fixed menu (e.g.\ \texttt{16:9}); consumed by the inference wrapper and not passed to the diffuser. \\
    \texttt{intent} & req. & string & One high-level sentence describing the whole scene and its key entities/interactions; when it conflicts with a specific field, the specific field takes precedence. \\
    \texttt{scene} & req. & object & Macro environment: \texttt{setting} (indoor/outdoor, time, weather) and a background \texttt{elements} list with the same structure as foreground elements. \\
    \texttt{elements} & req. & array & All independently-editable entities. Each has \texttt{id}, a concise \texttt{caption} (identity plus the single most salient feature), \texttt{position} (language / point / bbox, with bbox coordinates normalized to $0$--$999$ per axis), optional \texttt{depth} ($0$ nearest, $255$ farthest), an optional per-element \texttt{photography} object, and one named key per visual dimension: attribute keys (\texttt{material\_and\_surface}, \texttt{color}, \texttt{lighting\_interaction}, \ldots) and action keys (\texttt{pose}, \texttt{gesture}, \texttt{expression}, \texttt{gaze}, \texttt{action}). Raw human keypoints are annotation evidence, not a schema field. \\
    \texttt{relationships} & opt. & array & Textual statements of interaction or spatial relation between entities, each referencing their bounding boxes. \\
    \texttt{atmosphere} & opt. & string & Overall mood, in a few words. \\
    \texttt{photography} & opt. & object & Global camera: \texttt{layout}, \texttt{shot\_type}, \texttt{camera\_angle}, \texttt{lens\_and\_effect}. \\
    \texttt{style} & opt. & string & Overall artistic style, e.g.\ ``photorealistic''. \\
    \texttt{lighting} & opt. & string & Overall lighting environment and its interaction with the scene. \\
    \bottomrule
  \end{tabularx}
\end{table}

\subsection{Annotation pipeline}
\label{app:impl_annotation}

Figure~\ref{fig:pipeline} traces the five stages that populate the L10 schema, while Table~\ref{tab:schema_fields} defines the resulting fields.
Every model in the annotation pipeline runs frozen at our inference settings.

\paragraph{(1) Global scene understanding.}
A VLM (Seed-VL) reads the whole image and emits the high-level \texttt{intent}, the global \texttt{atmosphere}, \texttt{style}, \texttt{lighting}, and \texttt{photography} fields, and the \texttt{scene} block (setting and background elements); it also lays out the JSON skeleton---splitting foreground from background and assigning each entity an \texttt{id} ordered by compositional importance.

\paragraph{(2) Per-element description.}
Each entity is cropped and re-captioned by the VLM into a concise \texttt{caption} plus one named key per visual dimension: attribute keys (color, material, surface) and action keys (pose, gesture, expression, gaze).
For each person, Sapiens~\citep{sapiens} predicts $133$ keypoints that are rendered as a pose overlay for a downstream VLM pass; the overlay helps resolve body-side orientation and joint geometry, but the raw keypoints are not written into the SP.

\paragraph{(3) Spatial annotation.}
DepthAnything~V2~\citep{depthanything2} supplies relative element depth, quantized from $0$ (nearest) to $255$ (farthest) when available; otherwise the \texttt{depth} field is omitted.
SAM~2.1~\citep{sam21} supplies per-element masks and occlusion cues.
Bounding-box, mask, and depth evidence supports geometric relations such as overlap, containment, relative position, and depth order; semantic relations such as support and interaction are inferred during the final VLM reconciliation pass.

\paragraph{(4) Assembly.}
A second VLM pass merges the stage-1--3 outputs into a well-formed L10 structured record, constrained by the schema skeleton from stage~1.

\paragraph{(5) Degradation sampling.}
Table~\ref{tab:degradation} defines the deterministic field-group masks used to derive the L9$\to$L5 variants.
One set of degradations is generated per L10 annotation and reused across all training cells.

\paragraph{Matched NL controls.}
To isolate the structured interface from additional training, the end-to-end NL control uses the same training images, Qwen-Image initialization, diffusion recipe and budget, and matched prompter-training budget as the SP system, while using free-form NL captions throughout.
Its prompter uses the same Qwen3.5-397B-A17B base checkpoint, rank-$128$ LoRA, SFT--Cold-start--RFT sequence, stage-wise data volumes, optimizer schedules, training steps, decoding settings, verifier and privileged-teacher configurations, and checkpoint-selection rule as the SP prompter.
Only the intermediate caption representation and corresponding training targets change from structured prompts to free-form NL captions.
The NL controls are generated from the same stage-1--3 evidence bundle used to assemble the SPs, not by flattening or compressing the final JSON.
For each image, the NL verbalizer receives the global and crop-level VLM descriptions together with the same rendered Sapiens pose overlay, DepthAnything relative-depth evidence, and SAM segmentation and occlusion cues used by the SP pipeline, including intermediate evidence that is not serialized as a separate SP field.
It is instructed to preserve the source entity inventory and image-specific facts while expressing them as free-form prose at the target token budget.
Across the NL budgets, additional length is introduced through elaboration and connective phrasing rather than access to new annotation evidence.
Appendix~\ref{app:prompts_teaser} gives the exact teaser construction.
Aggregate results appear in Table~\ref{tab:sota}, with category-level breakdowns in Appendix~\ref{app:additional_results}.

\subsection{Diffusion-backbone training}
\label{app:impl_diffusion}

\paragraph{Backbones.}
We train two backbones: BAGEL~\citep{bagel} for the scaling-property fit of \S\ref{sec:metrics} (one checkpoint per cell) and Qwen-Image-2512~\citep{qwenimage} for the promptability sweeps and main results (\S\ref{sec:prompter},~\ref{sec:main_results}).
BAGEL is a $7$B unified model---Qwen2.5-7B with Mixture-of-Transformer-Experts (MoT) layers separating understanding and generation experts, a SigLIP2 vision encoder, and a frozen Flux KL-VAE---initialized from an in-house continued-training (CT) checkpoint (not part of the BAGEL public release); Qwen-Image is initialized from the Qwen-Image-2512 public release (a $60$-layer DiT with a frozen Qwen2.5-VL-7B text encoder and KL-VAE). Full hyperparameters are in Tables~\ref{tab:hyperparams_bagel} and~\ref{tab:hyperparams_qwenimage}.

\begin{table}[h]
  \centering
  \caption{\textbf{BAGEL training setup.} One BAGEL checkpoint is trained per caption condition in the scaling-property sweep (\S\ref{sec:metrics}); the same recipe is used for every cell.}
  \label{tab:hyperparams_bagel}
  \small
  \begin{tabularx}{0.78\linewidth}{@{}lX@{}}
    \toprule
    Hyperparameter & Value \\
    \midrule
    Initial checkpoint        & BAGEL CT checkpoint (Qwen2.5-7B MoT $+$ SigLIP2 $+$ Flux VAE) \\
    Sequence length (packed)  & $32{,}768$ \\
    Learning rate             & $5 \times 10^{-5}$ \\
    LR schedule               & linear warmup (1k steps), then constant \\
    Optimizer                 & AdamW, $\beta_1\!=\!0.9$, $\beta_2\!=\!0.95$, $\epsilon\!=\!10^{-15}$, weight decay $0$ \\
    Max gradient norm         & $1.0$ \\
    EMA decay                 & $0.999$ \\
    Diffusion objective       & rectified flow ($v$-prediction); timestep shift $4.0$ \\
    Precision                 & bf16 compute, fp32 optimizer moments \\
    GPUs (FSDP \textsc{hybrid\_shard}) & $192$ \\
    Effective tokens/step     & $6.3$M ($32{,}768 \times 192$) \\
    \bottomrule
  \end{tabularx}
\end{table}

\begin{table}[h]
  \centering
  \caption{\textbf{Qwen-Image training setup.}
  Initialized from the Qwen-Image-2512 public release~\citep{qwenimage}, trained once on the SP/NL caption corpus, then held fixed across prompter ablations (\S\ref{sec:ablations}) and main results (\S\ref{sec:main_results}).}
  \label{tab:hyperparams_qwenimage}
  \small
  \begin{tabularx}{0.78\linewidth}{@{}lX@{}}
    \toprule
    Hyperparameter & Value \\
    \midrule
    Initial checkpoint        & Qwen-Image-2512 public release~\citep{qwenimage} ($60$-layer DiT; frozen Qwen2.5-VL-7B text encoder $+$ KL-VAE) \\
    Sequence length (packed)  & $32{,}768$ \\
    Learning rate             & $1 \times 10^{-4}$ \\
    LR schedule               & linear warmup (2k steps), then constant \\
    Optimizer                 & AdamW, $\beta_1\!=\!0.9$, $\beta_2\!=\!0.95$, $\epsilon\!=\!10^{-15}$, weight decay $0$ \\
    Max gradient norm         & $1.0$ \\
    EMA decay                 & $0.9999$ \\
    Diffusion objective       & flow matching ($v$-prediction); resolution-dependent timestep shift \\
    Precision                 & bf16 compute, fp32 optimizer moments \\
    GPUs (FSDP \textsc{hybrid\_shard}) & $512$ \\
    Effective tokens/step     & $16.8$M ($32{,}768 \times 512$) \\
    \bottomrule
  \end{tabularx}
\end{table}

\paragraph{Tokenizer and sequence packing.}
BAGEL tokenizes text with the Qwen2.5-7B BPE tokenizer and Qwen-Image with the Qwen2.5-VL tokenizer ($152{,}064$-token vocabulary); for diffusion training, each structured record is deterministically serialized in a compact single-quote form (\texttt{'} delimiters, minimal separators) to save tokens.
Both encode images with a frozen KL-VAE at $8\times$ spatial compression and $16$ latent channels---the Flux VAE for BAGEL, a $3$D causal KL-VAE for Qwen-Image---followed by $2\times2$ patchification, so each latent token carries $16\!\cdot\!2\!\cdot\!2\!=\!64$ channels and a $1024^2$ image becomes $64\times64\!=\!4096$ tokens ($16\times$ effective downsample); both train at multi-aspect resolutions (BAGEL $512$--$1024$px, Qwen-Image $768$--$1536$px).
Training packs several (image, caption) pairs into one $32{,}768$-token sequence (at most $16{,}384$ tokens per sample) under block-diagonal flex-attention masks that confine each sample to itself, so the effective batch is $32{,}768\!\times\!G$ tokens per step ($G\!=\!192$ GPUs for BAGEL, $6.3$M tokens/step; $G\!=\!512$ for Qwen-Image, $16.8$M tokens/step).

\paragraph{Loss and precision.}
Both backbones use a rectified-flow / flow-matching objective: with $x_t\!=\!(1\!-\!t)\,x_0 + t\,\epsilon$ and $\epsilon\!\sim\!\mathcal{N}(0,I)$, the network predicts the velocity $v\!=\!\epsilon\!-\!x_0$ under a per-token MSE weighted uniformly across timesteps.
Timesteps are sampled and shifted toward noisier states---a fixed shift of $4.0$ for BAGEL, a sequence-length-dependent shift for Qwen-Image ($0.5$ at $256$ tokens rising to $0.9$ at $8192$).
Optimization is AdamW ($\beta_1\!=\!0.9$, $\beta_2\!=\!0.95$, $\epsilon\!=\!10^{-15}$) with gradient clipping at $1.0$ and no accumulation, in bf16 compute with fp32 optimizer moments and gradient checkpointing.
Classifier-free guidance is enabled by conditioning dropout during training (text $0.1$; reference-image VAE $0.1$ for BAGEL, $0.3$ for Qwen-Image).
BAGEL is a full-parameter finetune of its LLM, vision encoder, and projection/embedding layers (VAE frozen); Qwen-Image is a full finetune of its DiT with both the text encoder and VAE frozen.

\paragraph{Cumulative-token budget.}
Each BAGEL scaling-property cell is compared at the common cumulative image-token budget of $2.84\!\times\!10^{10}$ tokens reached by every run; this defines the matched-budget ``converged'' MSE used in the fits.
Appendix~\ref{app:scaling_supplement} documents the token-axis correction for two cells (L5 and L9) that were resumed after a counter reset.
Qwen-Image is instead trained once for $500{,}000$ steps.

\subsection{Schema field ablation}
\label{app:schema_field_ablation}

\begin{figure}[!t]
  \centering
  \begin{minipage}[c]{0.48\linewidth}
    \centering
    \includegraphics[width=\linewidth]{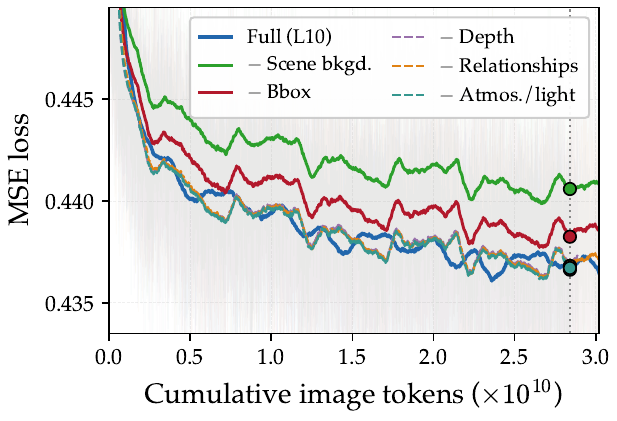}
  \end{minipage}%
  \hfill
  \begin{minipage}[c]{0.48\linewidth}
    \centering
    \small
    \begin{tabular}{@{}lrrr@{}}
      \toprule
      Configuration & MSE & $\Delta$MSE & $\Delta$GPG \\
      \midrule
      Full schema (L10) & 0.43699 & --- & --- \\
      \midrule
      $-$ Scene bkgd.   & 0.44052 & $+$35.3 & $-$42.0 \\
      $-$ Bbox          & 0.43843 & $+$14.4 & $-$5.8 \\
      $-$ Depth         & 0.43714 & $+$1.5 & --- \\
      $-$ Relationships & 0.43706 & $+$0.7 & $-$3.3 \\
      $-$ Atmos./light  & 0.43703 & $+$0.4 & --- \\
      \bottomrule
    \end{tabular}
  \end{minipage}
  \caption{\textbf{Schema field ablation} (BAGEL backbone).
  \emph{Left:} loss curves after removing one L10 field group.
  \emph{Right:} common-budget MSE is the mean over the trailing $5\%$ of image tokens ending at the shared $2.84\times10^{10}$ cut, matching the read-out in Table~\ref{tab:all_cells}; the table reports changes from L10, with $\Delta$MSE in $10^{-4}$ units.
  Scene context has the largest loss impact, followed by bounding boxes; depth, relationships, and atmosphere/lighting produce smaller changes.}
  \label{fig:field_decomposition}
\end{figure}

We remove one field group at a time from the full L10 schema, retrain BAGEL, and compare all six runs at a common training budget of $2.84\times10^{10}$ cumulative image tokens.
As Figure~\ref{fig:field_decomposition} shows, scene context is the most influential field group in this controlled setting: removing it reduces GPG by $42$ and raises MSE by $35.3\times10^{-4}$, a substantially larger loss increase than any other ablation.
Removing bounding boxes gives the second-largest increase at $14.4\times10^{-4}$, while removing depth, relationships, or atmosphere/lighting changes MSE by at most $1.5\times10^{-4}$.
The ablation therefore identifies global scene context as the dominant field group for diffusion learning under this setup.
This ranks each field group's marginal loss contribution when present; it does not measure annotation \emph{accuracy}, which is where the pose, depth, and segmentation experts act. A general VLM's geometric errors would be serialized as incorrect conditioning, so faithful expert-derived fields matter for correct generation and editing even where a group's marginal loss contribution is small.

\subsection{Prompter training}
\label{app:training_prompter}

The prompter is a rank-$128$ LoRA adapter on top of Qwen3.5-397B-A17B, trained serially with SFT, Cold-start, and RFT (\S\ref{sec:prompter}).
All three stages share the same LoRA topology, and the base Qwen3.5-397B-A17B weights remain frozen throughout.
Table~\ref{tab:hyperparams_prompter} summarizes their hyperparameters; the stage definitions follow below.

\begin{table}[h]
  \centering
  \caption{\textbf{Per-stage prompter training setup.}
  The Qwen3.5-397B-A17B base is frozen and only the rank-$128$ LoRA is updated.
  Stage definitions are given in the text.}
  \label{tab:hyperparams_prompter}
  \small
  \begin{tabularx}{\linewidth}{@{}lXXX@{}}
    \toprule
    Hyperparameter & SFT & Cold-start & RFT \\
    \midrule
    LoRA rank / $\alpha$     & $128$ / $256$ & $128$ / $256$ & $128$ / $256$ \\
    LoRA target modules      & all linear  & all linear  & all linear \\
    Sequence length (packed) & $20{,}480$  & $32{,}768$  & $49{,}152$ \\
    Learning rate            & $1\times10^{-5}$ & $1\times10^{-5}$ & $4\times10^{-5}$ \\
    LR schedule              & cosine, $5\%$ warmup & cosine, $10\%$ warmup & cosine, $5\%$ warmup \\
    Weight decay             & $0.1$       & $0.1$       & $0.1$ \\
    Global batch             & $64$        & $128$       & $32$ \\
    \bottomrule
  \end{tabularx}
\end{table}

\paragraph{Distributed setup.}
SFT and Cold-start run under Megatron-LM (via \texttt{ms-swift}'s \texttt{megatron} entry point) with tensor/pipeline/expert parallelism ($\text{TP}\!=\!\text{PP}\!=\!\text{EP}\!=\!4$, context parallel disabled), sequence parallelism, a distributed (ZeRO-1-like) optimizer, full uniform gradient checkpointing (one layer), FlashAttention, and bf16 compute with fp32 optimizer moments; sequences are packed with padding-free batching.
RFT runs under DeepSpeed ZeRO-3 on a single $8$-GPU node with gradient checkpointing, FlashAttention, and the colocated vLLM rollout engine described below.

\subsubsection{Training stages}
\label{app:impl_prompter}

\paragraph{Stage 1 --- SFT.}
The SFT corpus is a token-balanced mixture of ${\sim}333$k examples (${\sim}0.97$B tokens): roughly one third is the core (user prompt $\to$ structured-prompt JSON) task, where the user prompt is the image's original caption and the target is its image-derived SP from \S\ref{sec:sp}, plus a reverse image-to-JSON set. The remaining two thirds is reasoning/instruction replay (EN/ZH long chain-of-thought, vision--language reasoning, general dialogue, and a small Qwen3.5 base-identity anchor) included to preserve the base model's capabilities (DMT-style anti-forgetting).
For the prompt $\to$ structured-prompt task, the target JSON is treated as one plausible visual completion of the user prompt, not a recoverable ground-truth layout; the stage therefore teaches the distribution of structured completions seen by the diffuser, including non-canonical crops and occlusions, rather than a deterministic prompt-to-layout map.
The objective is token-level cross-entropy on the assistant tokens only (system and user prompts masked). Core SFT targets follow the \texttt{<think></think>} $+$ JSON interface; their empty thinking block is excluded from the loss, so SFT teaches the SP completion without a reasoning trace while preserving the model's thinking-toggle convention.
We train for one epoch (${\sim}740$ steps).

\paragraph{Stage 2 --- Cold-start.}
The final cold-start corpus contains $50{,}182$ unique (original caption $\to$ thinking trace $\to$ structured JSON) examples selected from $172{,}208$ image-conditioned teacher candidates. Qwen3.5-397B-A17B generates one candidate per example in high-reasoning mode with access to the original caption and paired image, using temperature $1.0$, top-$p$ $0.9$, and a source-dependent maximum completion length of $16{,}000$--$65{,}536$ tokens. Appendix~\ref{app:cold_start_sysprompt} reproduces its system prompt. The raw teacher response uses an \texttt{<analysis>} block for validation; accepted traces are mapped to the student's \texttt{<think>} interface before training. The prompter receives only the original caption during training and inference.

We filter candidates with Gemini 3 Pro (\texttt{gemini-3-pro-preview-new}) using validator v4. Five independent calls assess (i) prompt--image alignment (\texttt{aligned}/\texttt{partial}/\texttt{mismatch}) and four reasoning-quality axes scored as \texttt{none}/\texttt{minor}/\texttt{major}: (ii) whether inferred details are properly introduced, (iii) whether imagined specifics are justified, (iv) reverse rationalization, and (v) violations of the prescribed reasoning-stage boundaries. The strict gate requires \texttt{aligned} on the first axis and no \texttt{major} flag on the remaining four; \texttt{minor} flags are retained without reranking, while parsing or API failures are rejected. This gate accepts $58{,}907$ candidates ($34.2\%$), and prompt-level deduplication yields the $50{,}182$ training examples. Unsupported imagined specifics are the dominant rejection mode, occurring in $71.9\%$ of rejected traces and acting as the sole rejection reason in $40.7\%$; prompt--image misalignment and reverse rationalization are the next most common causes. Appendix~\ref{app:cold_start_filter_prompts} reproduces the exact validator-v4 prompts.
The loss is token-level cross-entropy on the assistant tokens---both the converted \texttt{<think>} block and the JSON---with the same prompt masking as SFT.
We run ${\sim}8$ epochs as two chained $4$-epoch sub-runs sharing the same hyperparameters.

\paragraph{Stage 3 --- RFT (gated OPSD).}
Appendix~\ref{app:opsd} gives the complete RFT specification, including its training data, rollout procedure, acceptance rule, and OPSD objective.

\subsubsection{OPSD: on-policy self-distillation}
\label{app:opsd}

OPSD (On-Policy Self-Distillation) is the image-conditioned distillation objective in the prompter's RFT stage (\S\ref{sec:prompter}, Stage~3).
RFT uses \texttt{ms-swift}'s on-policy distillation trainer over $10{,}003$ original-caption--image pairs stratified across six prompt sources.
Its design follows the conditioning asymmetry measured by GPG (\S\ref{sec:gpg}): an image-conditioned model can supply training information unavailable to the image-free inference-time prompter.
OPSD assumes the student already emits parseable, image-grounded schemas, which is ensured by Stages~1 (SFT) and~2 (Cold-start) of the pipeline.

\paragraph{Teacher and student.}
The student $\pi_\theta$ is the prompter under training: a rank-$128$ LoRA on Qwen3.5-397B-A17B after SFT + Cold-start, run image-free.
The teacher $\pi^{\star}$ is the same Qwen3.5-397B-A17B base \emph{without} our LoRA, queried with the same user prompt \emph{and} the reference image in thinking mode at training time only.
Teacher parameters are frozen throughout, and no gradients flow through $\pi^{\star}$.

\paragraph{On-policy rollouts.}
For each (user prompt, image) pair, we sample a structured-caption rollout $\tau \!=\! (\tau_1, \ldots, \tau_T)$ from the student image-free with temperature $1.0$ and no nucleus truncation ($p\!=\!1.0$).
Rollouts are served by a vLLM instance colocated on the training GPUs.
At each token position $t$ we then evaluate both policies:
\begin{itemize}
  \item teacher distribution $\pi^{\star}(\cdot \mid \tau_{<t}, \text{prompt}, I)$: a forward pass under the teacher with image conditioning,
  \item student distribution $\pi_\theta(\cdot \mid \tau_{<t}, \text{prompt})$: a forward pass under the student without image conditioning.
\end{itemize}
The per-token divergence of Eq.~\eqref{eq:opsd}---in implementation a KL divergence, evaluated over the top-$64$ teacher logits with each token's divergence clipped at $5.0$---is summed over the rollout's response positions and normalized by their count.
Sampling on-policy (from $\pi_\theta$ rather than $\pi^{\star}$) keeps the gradient supported on captions the student actually emits at inference, preventing mode-collapse onto teacher behaviors unreachable image-free; this is the ``on-policy'' in OPSD.

\paragraph{Combination with the verifier.}
Each rollout is rendered and scored by the Gemini verifier on structure, alignment, and aesthetic quality; only rollouts scoring at least $6/10$ on all three axes are retained, while failed, unparsable, or incomplete API responses are dropped.
The OPSD objective $\mathcal{L}_{\text{OPSD}}$ (Eq.~\eqref{eq:opsd}) is applied only to these retained rollouts.
The verifier derives QA pairs from the user request and checks them against the rendered image; it never receives the rollout's SP.
The verifier therefore selects high-confidence training data, while the image-conditioned teacher supplies the token-level OPSD targets.
There is no weighted combination between the two signals, and the distillation objective carries no token-cross-entropy auxiliary (\texttt{sft\_alpha}$=0$).
The ablations of \S\ref{sec:ablations} isolate the two components with distinct training rules: verifier-reward GRPO (the verifier-only row) optimizes the verifier reward with no OPSD; OPSD-only applies $\mathcal{L}_{\text{OPSD}}$ to \emph{all} rollouts without a verifier gate; the full method applies OPSD only to verifier-accepted rollouts.

\subsection{Inference and the agentic loop}
\label{app:impl_inference}

\paragraph{Single-shot prompter inference.}
The trained prompter (rank-$128$ LoRA on Qwen3.5-397B-A17B, run image-free) is decoded \emph{greedily} at evaluation time---temperature $0$, repetition penalty $1.0$, up to $16{,}384$ new tokens---in contrast to the temperature-$1.0$ on-policy rollouts used during RFT.
It emits a \texttt{<think>} reasoning block followed by the JSON structured prompt. The generation wrapper strips the thinking trace, reads \texttt{ratio} to select the output canvas, removes that control field, and passes only the remaining structured record to the diffusion backbone.

\paragraph{Diffusion sampling.}
Images are rendered with a first-order Euler ODE solver over $50$ denoising steps on a linear schedule shifted per sample by resolution.
Qwen-Image (main results) uses classifier-free guidance scale $4.0$; BAGEL (scaling-property cells) uses $8.0$ with timestep shift $4.0$.
Benchmark images are generated at $1024\times1024$ by default; multi-aspect benchmarks use a $1536^2$-pixel-budget, ratio-aware schedule snapped to multiples of $64$.
All main-table results are \emph{single-shot}, with one prompter forward pass followed by one render.
Only Table~\ref{tab:agentic} evaluates the agentic loop.

\paragraph{Agentic loop runtime.}
Given a user prompt $u$, round $t$ forms $\mathrm{SP}_t = \pi_\theta(u, \mathrm{SP}_{<t}, c_{t-1})$ from the prior structured prompts $\mathrm{SP}_{<t}$ and the accumulated critique $c_{t-1}$ ($c_0=\emptyset$); the fixed diffuser renders $I_t$; and three independent Gemini calls inspect $(u, I_t)$ for structure, alignment, and aesthetic quality. Their outputs are aggregated into the three per-axis scores, a pass flag, and a structured issue list.
PASS requires all three axes to score at least $6/10$; any axis below threshold yields FAIL together with a structured critique $c_t$---a list of (field-path, observed-value, expected-value) triples over the violated constraints (missing objects, wrong counts, attribute mismatches, spatial-relation violations)---so the next round edits only the named fields rather than re-parsing free-form feedback.
The refiner receives the full critique history $\{c_1,\dots,c_{t-1}\}$ and the prior structured prompts; repeated critiques trigger escalation from local edits (round~2) to a structural change (round~3) and then a substantially different composition (round~4 onward).
The loop returns $I_t$ on PASS and hard-stops at $T_{\max}\!=\!8$ rounds; per round the wall-clock is ${\sim}20$--$35$\,s (prompter ${\sim}6$--$9$\,s, render ${\sim}5$--$10$\,s at $1024^2$, judge ${\sim}8$--$15$\,s).

\subsection{External dependencies and licenses}
\label{app:impl_licenses}

Table~\ref{tab:licenses} records the external code and model dependencies used in this work together with the currently available upstream license information.
Entries whose repositories provide no license file, or whose terms still require confirmation, are marked explicitly rather than treated as verified.

\begin{table}[h]
  \centering
  \caption{\textbf{External code and model dependencies.} Components are grouped by pipeline role and listed with their available upstream license information; unresolved entries are marked explicitly.}
  \label{tab:licenses}
  \small
  \begin{tabular}{@{}ll>{\raggedright\arraybackslash}p{0.34\linewidth}@{}}
    \toprule
    Component & License & Notes \\
    \midrule
    \multicolumn{3}{@{}l}{\textcolor{gray}{\emph{Models}}} \\
    BAGEL                & Apache-2.0 & diffusion backbone (\S\ref{sec:metrics}) \\
    Qwen-Image-2512      & Apache-2.0 & diffusion backbone (\S\ref{sec:prompter}) \\
    Qwen3.5 family       & Apache-2.0 & prompter SFT / cold-start / RFT student \\
    Qwen3.5-397B-A17B    & Apache-2.0 & OPSD teacher; GPG judge \\
    Sapiens              & CC-BY-NC-4.0 & pose estimation (annotation stage~2); non-commercial \\
    DepthAnything V2     & CC-BY-NC-4.0 & monocular depth (annotation stage~3); code Apache-2.0, large ckpt non-commercial \\
    SAM 2.1              & Apache-2.0 & segmentation + occlusion (annotation stage~3) \\
    Seed-VL              & Internal & scene + per-element captioning; API-only, not publicly released \\
    Gemini 3 Pro  & API ToS & ED source-tuple extractor (offline, cached); training-time RFT verifier and online agentic judge \\
    Gemini 2.5 Flash & API ToS & T2I-CoReBench evaluator \\
    GPT-4o        & API ToS & ED image-source extractor robustness check; WISE legacy evaluator \\
    GPT-5.4       & API ToS & ED caption-tuple extractor and matcher; offline structure, alignment, and GSB evaluator \\
    GPT-5.5       & API ToS & rewriting backend; ED image-source robustness check \\
    Claude Opus 4.8 / Claude Code & API ToS / proprietary & single-turn rewriting backend / agentic coding backend \\
    GLM-5.2       & API ToS & single-turn rewriting backend \\
    Codex         & Proprietary & agentic coding backend \\
    \midrule
    \multicolumn{3}{@{}l}{\textcolor{gray}{\emph{Training infrastructure}}} \\
    PyTorch              & BSD-3-Clause & \\
    Transformers (HF)    & Apache-2.0 & model loading, tokenizers \\
    diffusers (HF)       & Apache-2.0 & diffusion pipelines \\
    ms-swift             & Apache-2.0 & prompter LoRA training harness (wraps Megatron-LM for SFT / cold-start, DeepSpeed for RFT) \\
    Megatron-LM          & Apache-2.0 & TP/PP/EP backend for prompter SFT / cold-start (via ms-swift) \\
    DeepSpeed / FSDP     & Apache-2.0 / BSD-3-Clause & DeepSpeed (ZeRO-3) for prompter RFT; FSDP (ships with PyTorch) for the diffusion backbones \\
    vLLM                 & Apache-2.0 & colocated on-policy rollout engine for RFT \\
    FlashAttention       & BSD-3-Clause & \\
    \midrule
    \multicolumn{3}{@{}l}{\textcolor{gray}{\emph{Evaluation}}} \\
    GenEval & MIT & official scoring scripts \\
    GenEval++ & Upstream terms & evaluation released with Echo-4o; redistribution terms to verify \\
    GenEval2 & CC BY-NC 4.0 & official benchmark code and data; non-commercial use \\
    DPG-Bench            & Apache-2.0 & released within the ELLA repository \\
    TIIF                 & No license file & upstream repository provides no license file \\
    WISE                 & No license file & upstream repository provides no license file \\
    T2I-CoReBench        & Apache-2.0 & official benchmark dataset and evaluation code \\
    \bottomrule
  \end{tabular}
\end{table}

\section{Additional benchmark results}
\label{app:additional_results}

\paragraph{Protocol and score provenance.}
This section provides evaluation protocols, score provenance, and per-category breakdowns for the benchmarks reported in Table~\ref{tab:sota}. Published baselines follow the respective leaderboards or original papers unless noted otherwise. The Nano Banana DPG-Bench score is taken from \citet{he2026gems}. WISE uses the legacy WiScore protocol with GPT-4o-2024-05-13: we evaluate Nano Banana, Qwen-Image$^{*}$, matched NL, and \ours{} under this protocol, while the remaining WISE scores come from published legacy-protocol evaluations~\citep{niu2025wise,guo2025comfymind}. Qwen-Image$^{*}$ uses its official prompt enhancer; $^{\dagger}$ marks our re-evaluation on Qwen-Image-2512. Matched NL and \ours{} use the matched settings described in Appendix~\ref{app:impl_annotation}, and \ours{} uses single-shot inference in Table~\ref{tab:sota}.

\subsection{DPG-Bench per-category breakdown}
\label{app:dpg_detail}

Table~\ref{tab:dpg_detail} reports per-category scores on DPG-Bench, covering \emph{global} scene description, \emph{entity} presence, \emph{attribute} binding, \emph{relation} between entities, and \emph{other} dense-prompt aspects, followed by the overall score.

\begin{table}[h]
  \centering
  \caption{\textbf{DPG-Bench per-category scores.}
  Published category breakdowns are shown for the baselines; Qwen-Image$^{*}$, matched NL, and \ours{} report our matched per-category evaluations.}
  \label{tab:dpg_detail}
  \small
  \begin{tabular}{@{}lcccccc@{}}
    \toprule
    Model & Global & Entity & Attribute & Relation & Other & Overall \\
    \midrule
    SD v1.5            & 74.63 & 74.23 & 75.39 & 73.49 & 67.81 & 63.18 \\
    PixArt-$\alpha$    & 74.97 & 79.32 & 78.60 & 82.57 & 76.96 & 71.11 \\
    LUMINA-Next        & 82.82 & 88.65 & 86.44 & 80.53 & 81.82 & 74.63 \\
    SDXL               & 83.27 & 82.43 & 80.91 & 86.76 & 80.41 & 74.65 \\
    Playground v2.5    & 83.06 & 82.59 & 81.20 & 84.08 & 83.50 & 75.47 \\
    Hunyuan-DiT        & 84.59 & 80.59 & 88.01 & 74.36 & 86.41 & 78.87 \\
    Janus              & 82.33 & 87.38 & 87.70 & 85.46 & 86.41 & 79.68 \\
    PixArt-$\Sigma$    & 86.89 & 82.89 & 88.94 & 86.59 & 87.68 & 80.54 \\
    Emu3-Gen           & 85.21 & 86.68 & 86.84 & 90.22 & 83.15 & 80.60 \\
    Janus-Pro-1B       & 87.58 & 88.63 & 88.17 & 88.98 & 88.30 & 82.63 \\
    DALL$\cdot$E 3     & 90.97 & 89.61 & 88.39 & 90.58 & 89.83 & 83.50 \\
    FLUX.1 Dev         & 74.35 & 90.00 & 88.96 & 90.87 & 88.33 & 83.84 \\
    SD3 Medium         & 87.90 & 91.01 & 88.83 & 80.70 & 88.68 & 84.08 \\
    Janus-Pro-7B       & 86.90 & 88.90 & 89.40 & 89.32 & 89.48 & 84.19 \\
    HiDream-I1-Full    & 76.44 & 90.22 & 89.48 & 93.74 & 91.83 & 85.89 \\
    Seedream 3.0       & 94.31 & 92.65 & 91.36 & 92.78 & 88.24 & 88.27 \\
    GPT-Image-1        & 88.89 & 88.94 & 89.84 & 92.63 & 90.96 & 85.15 \\
    Qwen-Image         & 91.32 & 91.56 & 92.02 & 94.31 & 92.73 & 88.32 \\
    Show-o             & 79.33 & 75.44 & 78.02 & 84.45 & 60.80 & 67.27 \\
    TokenFlow-XL       & 78.72 & 79.22 & 81.29 & 85.22 & 71.20 & 73.38 \\
    OmniGen            & 87.90 & 88.97 & 88.47 & 87.95 & 83.56 & 81.16 \\
    OmniGen2           & 88.81 & 88.83 & 90.18 & 89.37 & 90.27 & 83.57 \\
    BAGEL              & 88.94 & 90.37 & 91.29 & 90.82 & 88.67 & 85.07 \\
    UniWorld-V1        & 83.64 & 88.39 & 88.44 & 89.27 & 87.22 & 81.38 \\
    Ovis-U1            & 82.37 & 90.08 & 88.68 & 93.35 & 85.20 & 83.72 \\
    Skywork UniPic     & 89.65 & 87.78 & 90.84 & 91.89 & 91.95 & 85.50 \\
    \midrule
    Qwen-Image$^{*}$   & 89.04 & 91.91 & 92.39 & 90.85 & 93.07 & 87.20 \\
    Matched NL + Qwen-Image & 89.50 & 92.30 & 92.00 & 90.50 & 93.50 & 87.80 \\
    \ours{} (Qwen-Image) & 92.05 & 94.13 & 94.48 & 93.36 & 94.97 & \textbf{90.71} \\
    \bottomrule
  \end{tabular}
\end{table}

\clearpage
\subsection{GenEval per-skill breakdown}
\label{app:geneval_detail}

Table~\ref{tab:geneval_detail} reports per-skill accuracy on GenEval, covering single-object / two-objects presence, counting, colors, position, and color-attribute binding, with the overall accuracy in the last column.

\begin{table}[h]
  \centering
  \caption{\textbf{GenEval per-skill accuracy.}
  Published per-skill breakdowns are shown for the baselines; \ours{} reports our evaluation. Qwen-Image$^{*}$ uses its official prompt enhancement, as in Table~\ref{tab:sota}.}
  \label{tab:geneval_detail}
  \small
  \resizebox{\linewidth}{!}{%
  \begin{tabular}{@{}lccccccc@{}}
    \toprule
    Model & Single Obj. & Two Obj. & Counting & Colors & Position & Color Attr. & Overall \\
    \midrule
    SD v2.1              & 0.98 & 0.51 & 0.44 & 0.85 & 0.07 & 0.17 & 0.50 \\
    SDXL                 & 0.98 & 0.74 & 0.39 & 0.85 & 0.15 & 0.23 & 0.55 \\
    IF-XL                & 0.97 & 0.74 & 0.66 & 0.81 & 0.13 & 0.35 & 0.61 \\
    PixArt-$\alpha$      & 0.98 & 0.50 & 0.44 & 0.80 & 0.08 & 0.07 & 0.48 \\
    LUMINA-Next          & 0.92 & 0.46 & 0.48 & 0.70 & 0.09 & 0.13 & 0.46 \\
    SD3 Medium           & 0.99 & 0.94 & 0.72 & 0.89 & 0.33 & 0.60 & 0.74 \\
    SD3.5 Large          & 0.98 & 0.89 & 0.73 & 0.83 & 0.34 & 0.47 & 0.71 \\
    FLUX.1 Dev           & 0.99 & 0.81 & 0.79 & 0.74 & 0.20 & 0.47 & 0.67 \\
    NOVA                 & 0.99 & 0.91 & 0.62 & 0.85 & 0.33 & 0.56 & 0.71 \\
    TokenFlow-XL         & 0.95 & 0.60 & 0.41 & 0.81 & 0.16 & 0.24 & 0.55 \\
    Janus                & 0.97 & 0.68 & 0.30 & 0.84 & 0.46 & 0.42 & 0.61 \\
    JanusFlow            & 0.97 & 0.59 & 0.45 & 0.83 & 0.53 & 0.42 & 0.63 \\
    Janus-Pro-7B         & 1.00 & 0.98 & 0.79 & 0.91 & 0.60 & 0.72 & 0.83 \\
    Emu3-Gen             & 0.98 & 0.71 & 0.34 & 0.81 & 0.17 & 0.21 & 0.54 \\
    Show-o               & 0.95 & 0.52 & 0.49 & 0.82 & 0.11 & 0.28 & 0.53 \\
    OmniGen              & 0.98 & 0.84 & 0.66 & 0.74 & 0.40 & 0.43 & 0.68 \\
    OmniGen2             & 1.00 & 0.95 & 0.64 & 0.88 & 0.55 & 0.76 & 0.80 \\
    HiDream-I1-Full      & 0.99 & 0.89 & 0.59 & 0.90 & 0.79 & 0.66 & 0.80 \\
    BAGEL                & 0.99 & 0.94 & 0.81 & 0.88 & 0.64 & 0.63 & 0.82 \\
    UniWorld-V1          & 0.99 & 0.93 & 0.79 & 0.89 & 0.49 & 0.70 & 0.80 \\
    Seedream 3.0         & 0.99 & 0.96 & 0.91 & 0.93 & 0.47 & 0.80 & 0.84 \\
    GPT-Image-1          & 0.99 & 0.92 & 0.85 & 0.92 & 0.75 & 0.61 & 0.84 \\
    Ovis-U1              & 0.98 & 0.98 & 0.90 & 0.92 & 0.79 & 0.75 & 0.89 \\
    Skywork UniPic       & 0.98 & 0.92 & 0.74 & 0.91 & 0.89 & 0.72 & 0.86 \\
    Qwen-Image           & 0.99 & 0.92 & 0.89 & 0.88 & 0.76 & 0.77 & 0.87 \\
    Qwen-Image$^{*}$     & 1.00 & 0.95 & 0.93 & 0.92 & 0.87 & 0.83 & 0.91 \\
    Matched NL + Qwen-Image & 1.00 & 0.95 & 0.92 & 0.93 & 0.86 & 0.82 & 0.91 \\
    \ours{} (Qwen-Image) & 1.00 & 0.96 & 0.94 & 0.95 & 0.93 & 0.86 & \textbf{0.94} \\
    \bottomrule
  \end{tabular}}
\end{table}

\clearpage
\subsection{WISE per-category breakdown}
\label{app:wise_detail}

Table~\ref{tab:wise_detail} reports per-category WiScore under the legacy WISE protocol, which uses GPT-4o-2024-05-13 to score consistency, realism, and aesthetic quality. WISE covers six domains: cultural, temporal, spatial, biology, physics, and chemistry. Models are grouped into dedicated T2I diffusion models, unified multimodal LLMs, proprietary systems, and open systems with prompt rewriting. The official overall WiScore aggregates prompt-level scores with domain weights $40\%$, $16.7\%$, $13.3\%$, $10\%$, $10\%$, and $10\%$ in the displayed domain order, so it is not the unweighted mean of the six category entries. The four rows we evaluate ourselves follow that rule and reproduce from their own entries to within the displayed precision. Published overalls are transcribed from their sources rather than recomputed, and not all of them can be recovered from the rounded per-category entries reported alongside them: BAGEL, most visibly, reports $0.52$, which sits $0.019$ above the weighted combination of its own six categories.

\begin{table}[h]
  \centering
  \caption{\textbf{WISE per-category WiScore under the legacy protocol.} Published baseline scores follow the official legacy leaderboard; Nano Banana, Qwen-Image$^{*}$, matched NL, and \ours{} report our evaluations using the same legacy evaluator and scoring rule. Bold marks the best in each column.}
  \label{tab:wise_detail}
  \small
  \begin{tabular}{@{}lccccccc@{}}
    \toprule
    Model & Cultural & Temporal & Spatial & Biology & Physics & Chemistry & Overall \\
    \midrule
    \multicolumn{8}{l}{\textcolor{gray}{\emph{Dedicated T2I}}} \\
    FLUX.1 Dev        & 0.48 & 0.58 & 0.62 & 0.42 & 0.51 & 0.35 & 0.50 \\
    FLUX.1 Schnell    & 0.39 & 0.44 & 0.50 & 0.31 & 0.44 & 0.26 & 0.40 \\
    SD-3.5-large      & 0.44 & 0.50 & 0.58 & 0.44 & 0.52 & 0.31 & 0.46 \\
    SD-3.5-medium     & 0.43 & 0.50 & 0.52 & 0.41 & 0.53 & 0.33 & 0.45 \\
    SD-XL-base        & 0.43 & 0.48 & 0.47 & 0.44 & 0.45 & 0.27 & 0.43 \\
    SD-3-medium       & 0.42 & 0.44 & 0.48 & 0.39 & 0.47 & 0.29 & 0.42 \\
    SD-v1-5           & 0.34 & 0.35 & 0.32 & 0.28 & 0.29 & 0.21 & 0.32 \\
    SD-2-1            & 0.30 & 0.38 & 0.35 & 0.33 & 0.34 & 0.21 & 0.32 \\
    \midrule
    \multicolumn{8}{l}{\textcolor{gray}{\emph{Unified MLLM}}} \\
    GPT-Image-1       & 0.81 & 0.71 & 0.89 & 0.83 & 0.79 & 0.74 & 0.80 \\
    DeepGen 1.0       & 0.72 & 0.81 & 0.70 & 0.67 & 0.82 & 0.66 & 0.73 \\
    LongCat-Image     & 0.66 & 0.61 & 0.72 & 0.66 & 0.72 & 0.49 & 0.65 \\
    NextFlow-RL       & 0.63 & 0.63 & 0.77 & 0.58 & 0.67 & 0.39 & 0.62 \\
    Qwen-Image        & 0.62 & 0.63 & 0.77 & 0.57 & 0.75 & 0.40 & 0.62 \\
    UniWorld-V2       & 0.60 & 0.61 & 0.70 & 0.53 & 0.64 & 0.32 & 0.58 \\
    HunyuanImage 3.0 & 0.58 & 0.57 & 0.70 & 0.56 & 0.63 & 0.31 & 0.57 \\
    MetaQuery-XL      & 0.56 & 0.55 & 0.62 & 0.49 & 0.63 & 0.41 & 0.55 \\
    UniWorld-V1       & 0.53 & 0.55 & 0.73 & 0.45 & 0.59 & 0.41 & 0.55 \\
    Manzano-30B       & 0.58 & 0.50 & 0.65 & 0.50 & 0.55 & 0.32 & 0.54 \\
    BAGEL             & 0.44 & 0.55 & 0.68 & 0.44 & 0.60 & 0.39 & 0.52 \\
    Emu3              & 0.34 & 0.45 & 0.48 & 0.41 & 0.45 & 0.27 & 0.39 \\
    Janus-Pro-7B      & 0.30 & 0.37 & 0.49 & 0.36 & 0.42 & 0.26 & 0.35 \\
    \midrule
    \multicolumn{8}{l}{\textcolor{gray}{\emph{Proprietary}}} \\
    Nano Banana       & 0.89 & \textbf{0.87} & \textbf{0.95} & \textbf{0.89} & \textbf{0.89} & 0.79 & \textbf{0.89} \\
    \midrule
    \multicolumn{8}{l}{\textcolor{gray}{\emph{Open with prompt rewriting}}} \\
    Qwen-Image$^{*}$               & 0.85 & 0.76 & 0.89 & 0.81 & 0.83 & 0.82 & 0.83 \\
    Matched NL + Qwen-Image        & 0.86 & 0.75 & 0.90 & 0.82 & 0.82 & 0.83 & 0.84 \\
    Qwen-Image + \ours{} (L10)     & \textbf{0.91} & 0.83 & 0.92 & 0.88 & \textbf{0.89} & \textbf{0.88} & \textbf{0.89} \\
    \bottomrule
  \end{tabular}
\end{table}

\clearpage
\subsection{T2I-CoReBench per-category breakdown}
\label{app:corebench_detail}

Table~\ref{tab:corebench_detail} expands the T2I-CoReBench comparison of Table~\ref{tab:sota} across the four \emph{Composition} and eight \emph{Reasoning} categories under the Gemini 2.5 Flash evaluator.
Published baseline scores are taken from the official leaderboard~\citep{li2026corebench}; Emu3.5 is omitted because the leaderboard does not report a corresponding row. Qwen-Image$^{*}$, matched NL, and \ours{} are evaluated with the same Gemini 2.5 Flash harness used for the main table.
The composition dimensions are multi-instance (MI), multi-attribute (MA), multi-relation (MR), and text rendering (TR). The reasoning dimensions are logical (LR), behavioural (BR), hypothetical (HR), procedural (PR), generalization (GR), analogical (AR), commonsense (CR), and reconstructive reasoning (RR). Composition and reasoning averages are unweighted means over their four and eight dimensions, respectively, and Overall is the unweighted mean over all twelve.
Our largest advantages appear in multi-attribute and multi-relation composition and across all eight reasoning categories, while multi-instance composition and text rendering remain stronger in Nano Banana and GPT-Image-1, respectively.

\begin{table}[h]
  \centering
  \caption{\textbf{T2I-CoReBench per-category comparison} under the Gemini 2.5 Flash evaluator.
  Published baselines correspond to models in Table~\ref{tab:sota}; Qwen-Image$^{*}$, matched NL, and \ours{} use our matched evaluation.
  Bold marks the best score in each column.}
  \label{tab:corebench_detail}
  \small
  \setlength{\tabcolsep}{4.2pt}
  \begin{tabular}{@{}lccccc@{}}
    \toprule
    Model & C-MI & C-MA & C-MR & C-TR & Comp. avg \\
    \midrule
    FLUX.1 Dev               & 58.6 & 60.3 & 44.1 & 31.1 & 48.6 \\
    OmniGen2                 & 67.9 & 64.1 & 48.3 & 19.2 & 49.9 \\
    BAGEL                    & 64.9 & 65.2 & 45.8 &  9.7 & 46.4 \\
    Qwen-Image               & 81.4 & 79.6 & 65.6 & 85.5 & 78.0 \\
    BAGEL + CoT              & 57.7 & 60.8 & 37.8 &  2.2 & 39.6 \\
    GPT-Image-1              & 84.1 & 75.9 & 72.7 & \textbf{86.4} & 79.8 \\
    Nano Banana              & \textbf{85.7} & 77.9 & 72.6 & 86.3 & 80.6 \\
    LongCat-Image            & 81.4 & 74.5 & 61.5 & 65.7 & 70.8 \\
    HunyuanImage 3.0         & 84.9 & 81.2 & 63.7 & 85.7 & 78.9 \\
    Qwen-Image$^{*}$         & 78.0 & 91.0 & 80.0 & 65.0 & 78.5 \\
    Matched NL + Qwen-Image  & 79.5 & 93.0 & 82.5 & 68.0 & 80.8 \\
    \ours{} (Qwen-Image)     & 83.8 & \textbf{95.5} & \textbf{87.7} & 72.1 & \textbf{84.8} \\
    \bottomrule
  \end{tabular}

  \vspace{0.5em}
  \setlength{\tabcolsep}{3.2pt}
  \resizebox{\linewidth}{!}{%
  \begin{tabular}{@{}lcccccccccc@{}}
    \toprule
    Model & R-LR & R-BR & R-HR & R-PR & R-GR & R-AR & R-CR & R-RR & Reason. avg & Overall \\
    \midrule
    FLUX.1 Dev               & 24.8 & 23.0 & 36.0 & 61.8 & 42.4 & 57.2 & 36.3 & 30.3 & 39.0 & 42.2 \\
    OmniGen2                 & 24.7 & 23.2 & 43.3 & 63.1 & 46.1 & 54.2 & 36.5 & 24.1 & 39.4 & 42.9 \\
    BAGEL                    & 23.4 & 21.9 & 33.0 & 51.6 & 31.2 & 50.4 & 32.4 & 29.3 & 34.1 & 38.2 \\
    Qwen-Image               & 41.1 & 32.2 & 48.2 & 75.1 & 56.5 & 53.3 & 61.9 & 26.4 & 49.3 & 58.9 \\
    BAGEL + CoT              & 25.5 & 25.4 & 33.9 & 58.6 & 53.5 & 56.9 & 41.6 & 39.8 & 41.9 & 41.1 \\
    GPT-Image-1              & 59.0 & 54.8 & 65.6 & 87.3 & 76.5 & 82.0 & 70.9 & 56.1 & 69.0 & 72.6 \\
    Nano Banana              & 64.5 & 64.9 & 67.1 & 85.2 & 84.1 & 83.1 & 71.3 & 68.7 & 73.6 & 75.9 \\
    LongCat-Image            & 39.1 & 35.7 & 48.5 & 75.5 & 72.5 & 61.4 & 58.8 & 41.0 & 54.1 & 59.6 \\
    HunyuanImage 3.0         & 39.6 & 32.8 & 51.4 & 72.4 & 54.1 & 54.1 & 57.0 & 27.7 & 48.6 & 58.7 \\
    Qwen-Image$^{*}$         & 85.1 & 59.6 & 64.2 & 84.6 & 80.3 & 71.7 & 71.9 & 64.5 & 72.7 & 74.7 \\
    Matched NL + Qwen-Image  & 86.0 & 61.5 & 65.0 & 85.0 & 81.5 & 73.0 & 72.5 & 65.5 & 73.8 & 76.1 \\
    \ours{} (Qwen-Image)     & \textbf{88.8} & \textbf{78.0} & \textbf{77.0} & \textbf{94.9} & \textbf{90.5} & \textbf{93.2} & \textbf{83.1} & \textbf{77.7} & \textbf{85.4} & \textbf{85.2} \\
    \bottomrule
  \end{tabular}}
\end{table}

\section{System prompts}
\label{app:system_prompts}

This appendix reproduces the system prompts used by the Cold-start teacher, the validator-v4 Cold-start filter, and the structure, alignment, aesthetic, and pairwise-preference judges.
The released codebase will also provide these prompts in directly reusable, machine-readable form.
\S\ref{app:cold_start_sysprompt} gives the Qwen3.5-397B-A17B teacher prompt used in the Cold-start stage of prompter training (\S\ref{sec:prompter}); \S\ref{app:cold_start_filter_prompts} gives the Gemini prompts used to filter the resulting traces; \S\ref{app:judge_prompts} gives the structure/alignment rubrics used by the online Gemini verifier and rerun with GPT-5.4 for offline evaluation; \S\ref{app:judge_aesthetic} gives the verifier's aesthetic rubric; and \S\ref{app:judge_gsb} gives the GSB pairwise rubric and aggregation rule.

\subsection{Cold-start teacher system prompt}
\label{app:cold_start_sysprompt}

Below is the verbatim system prompt fed to the Qwen3.5-397B-A17B teacher during the Cold-start stage of \S\ref{sec:prompter}.
The teacher receives (image, user prompt) as its multimodal conversation input and is instructed to emit an \texttt{<analysis>} block followed by a single JSON object; the image is supplied as a visual input rather than named by a textual placeholder in the system prompt.
The analysis is framed as a derivation from the user prompt toward a visual blueprint, while the paired image supplies privileged evidence during data construction.
The prompter learns this derivation style from the caption alone, so at inference it can produce a plausible completion without receiving the image.

\begin{lstlisting}[style=json, basicstyle=\ttfamily\tiny, breaklines=true, columns=fullflexible, keepspaces=true, frame=single, backgroundcolor=\color{gray!5}, caption={Cold-start teacher system prompt (verbatim).}]
# Role
You are an expert AI Visual Planner and Scene Director.

# Task
Convert a short image description into a richly detailed structured JSON blueprint. Your output must consist of TWO visible parts in this exact order:

1. an `<analysis>` block --- a long, structured construction-process document that traces the design from user prompt -> final visual blueprint;
2. a single JSON object containing the structured prompt itself.

Output format (strictly --- no extra text outside these two parts):

```
<analysis>
Stage A --- Knowledge & common-sense:
...
Stage B --- Reasoning chain:
...
Stage C --- Aspect ratio:
...
Stage D --- Compose & imagine:
...
Stage E --- Layout & spatial constraints:
...
</analysis>
{"ratio":"...", ...JSON...}
```

The `<analysis>` block is the construction-process record: every decision in the JSON must be traceable to a reasoned step inside `<analysis>`. Do not collapse the stages into one paragraph. Each stage must be a labeled section with the literal heading shown below. Even when the user prompt feels simple (e.g. "A red apple on a table"), every stage gets meaningful content.

**Reason-first discipline (applies to every stage).** Within each stage and for every concrete decision (chosen ratio, named element, picked color, assigned bbox, derived relationship), write the *reasoning* first and the *conclusion* last. Never declare an answer at the start of a sentence and tail-justify it ("bbox is X because Y", "ratio is 16:9 because Z"). Always derive: "Because Y, the bbox is X." This forward-chaining is what makes the analysis usable as cold-start training data; tail-rationalization teaches the student to fabricate justifications after the fact.

**Knowledge-design separation.** Stage A is strictly for **facts and conventions** the prompt presupposes (definitions, historical context, canonical depictions, object affordances, cultural / stylistic norms). It is NOT for design decisions specific to *this image*. All "this image will use X" choices belong in Stage B (reasoning chain) or Stage D (imagination), never in Stage A.

**Preserve uncertainty for under-specified attributes.** When the user prompt does not specify an attribute, do NOT collapse it to a single forced answer. Pick one concrete visual implementation (the JSON needs definite values), but frame it as a *choice among reasonable options*: "Because the prompt does not specify <attribute>, a reasonable visual choice is ..."; "A safe visual implementation here is ..."; "Optional details that could be added include ...". Never write "must", "therefore it is", or "the only valid ..." for under-specified attributes.

---

**Stage A --- Knowledge & common-sense.** State what is canonically / culturally / physically true that the user's prompt presupposes: domain facts, object affordances, causal/temporal context, canonical visual depictions of named entities (listing variants when ambiguous), cultural/aesthetic norms tied to style words. Facts only --- no "this image will look like Y".

**Stage B --- Reasoning chain.** For each evocative noun / adjective / phrase in the user prompt, trace why it leads to specific visual choices in the form "user said X -> therefore I picture Y because Z". For under-specified parts, hedge: "Because the prompt does not specify W, a reasonable choice is V". Justify, do not declare.

**Stage C --- Aspect ratio.** Choose strictly from {1:1, 4:3, 3:4, 16:9, 9:16, 3:2, 2:3, 4:5, 21:9} and briefly justify why the ratio fits the scene's spatial demands.

**Stage D --- Compose & imagine.** Construct the scene element by element in a deliberate build order. For each element, cover (as relevant): identity / material+surface / color / lighting interaction / pose / gesture / expression / gaze / action / appearance / clothing / accessories / hair / position quadrant / depth plane / imperfection. Pure visual description only; no subjective commentary. Every object that will appear in the JSON elements list MUST first be named and described here. Honor the hedge choices from Stage B.

**Stage E --- Layout & spatial constraints.** Walk through every element from Stage D and assign concrete bboxes on the 1000x1000 normalized grid. For each element, reason first (quadrant, depth, anchors, ratio-induced canvas geometry), then conclude the bbox in the form `-> bbox: <x_min y_min x_max y_max>, depth: <0-255>`. Constraint checks: composition placement (foreground larger and lower; background smaller and higher), perspective consistency, edge cropping (touch 0 or 999). After all bboxes are assigned, enumerate spatial relationships referencing the bbox coords --- direct surface contact, adjacency with direction, occlusion, background anchoring, distance / depth.

---

After `</analysis>`, emit the JSON object directly (no extra text, no markdown fence). Transcribe every element imagined in Stage D into the JSON. Distribute each element's visual detail across the schema: a CONCISE `caption` plus as many dynamic attribute / action keys as the element needs. The JSON must start with `"ratio":"..."`.

# JSON Schema (top-level)
- `ratio` (REQUIRED, FIRST)
- `intent` --- single declarative sentence
- `atmosphere` --- 2-4 words, mood only
- `style` --- 2-6 words, visual aesthetic only
- `lighting` --- light source/direction/quality/color temperature/shadow pattern only
- `elements` (list of dicts): each has `id`, `caption` (concise), `position` (<bbox>...</bbox>), `depth` (0-255), optional `photography` dict, plus dynamic attribute keys (material_and_surface, color, lighting_interaction, ...) and action keys (pose, gesture, expression, gaze, action)
- `relationships` (list of strings referencing bboxes)
- `scene` (dict): `setting` + `elements` list (same structure as foreground elements)
- `photography` (dict): `layout`, `shot_type`, `camera_angle`, `lens_and_effect`

# Key Rules
1. Output format is `<analysis>...</analysis>` + JSON. No commentary outside these two parts. No markdown fences around the JSON.
2. <analysis> is the supervised construction record --- every JSON decision must be traceable to a reasoned step inside <analysis>. Stages A->B->C->D->E must each appear as a labeled section.
3. Every object in JSON `elements` or `scene.elements` MUST have been explicitly imagined in Stage D.
4. Pure visual description only --- no subjective commentary or design-intent talk. Light physics, material properties, object state, spatial relationships only.
4a. Reason-first: for every decision write reasoning before conclusion. Never tail-justify.
4b. Knowledge-design separation: Stage A holds only facts/conventions, never image-specific design.
4c. Preserve uncertainty: hedge for under-specified attributes; never "must" / "therefore it is" / "the only valid" for prompt-silent attributes.
5. Ratio mandatory and first in JSON, from the 9-option menu.
6. `caption` carries identity + single most salient feature; all rich detail lives in named dynamic keys, never crammed into `caption`.
7. Dynamic key naming: one key per visual dimension; do not repeat content across `caption` and keys.
8. Field orthogonality: `lighting` = global light source only; `style` = aesthetic label only; `photography` = camera/lens tech only; `intent`/`atmosphere` = summary/mood only; per-element detail = the element's dynamic keys.
9. Bboxes reflect real-world spatial relationships; edge-cropped objects have bbox touching 0 or 999.
10. Transcribe ALL Stage D detail into JSON, distributed across `caption` + dynamic keys.
11. `<bbox>0 0 999 999</bbox>` ONLY for full-frame backgrounds (sky, floor).
12. No element count limit.
13. Decomposition: complex hybrids (e.g. "centaur") get an element for the whole plus separate elements for distinct parts, linked via `relationships`. Explicit quantities ("three cats") -> that many separate elements with unique ids.
14. Text rendering: text in image must be in double quotes inside the relevant element's keys with explicit font/color.
15. Position specificity: use foreground / mid-ground / distant background, top-left / centered / bottom-right; never vague "next to" without direction.
\end{lstlisting}

\subsection{Aesthetic judge system prompt}
\label{app:judge_aesthetic}

\begin{lstlisting}[style=json, basicstyle=\ttfamily\scriptsize, breaklines=true, columns=fullflexible, keepspaces=true, frame=single, backgroundcolor=\color{gray!5}, caption={Aesthetic judge system prompt (verbatim).}]
# Role

You are an AI image aesthetic quality inspector. You will be shown one generated image.

Your ONLY job is to evaluate the visual/aesthetic quality. Do NOT evaluate prompt alignment or structural accuracy --- those are handled separately.

**Scoring philosophy: Be critical. A score of 7 means "decent looking". Most AI images score 5-7.**

## Technical Defects Checklist

- [ ] Overexposure? (blown-out highlights, white patches with no detail)
- [ ] Underexposure? (crushed shadows, dark areas with no detail)
- [ ] Over-saturation? (colors unnaturally vivid, neon-like)
- [ ] Color banding / posterization? (visible color steps instead of smooth gradients)
- [ ] Blurriness where sharpness is expected? (not intentional depth-of-field)
- [ ] Visible artifacts? (compression artifacts, weird halos, noise)
- [ ] Unnatural skin texture? (waxy, plastic-looking, or overly smooth)
- [ ] Inconsistent detail level? (some areas sharp, others blurry for no reason)

## Artistic Merit

- Composition: Is the framing and layout pleasing?
- Lighting: Is the lighting natural and well-used?
- Color harmony: Do colors work well together?
- Overall impression: Would a viewer find this visually appealing?
- Does the aspect ratio work well with the composition?

## Scoring Scale

- 10: Stunning --- gallery quality, excellent technique, zero technical defects
- 8-9: Very appealing, at most one very minor technical imperfection
- 6-7: Looks decent, minor technical issues (slight over-saturation, minor artifacts)
- 4-5: Mediocre --- noticeable technical problems or unappealing composition
- 2-3: Poor --- multiple technical defects, unpleasant to look at
- 0-1: Ugly, chaotic, severe technical failures

## Output

List every aesthetic issue found, then score.

{
  "score": $score,
  "pass": true or false,
  "issues": ["issue 1", "issue 2", ...]
}

**Pass rule:** score >= 6.
\end{lstlisting}

\subsection{Cold-start filtering system prompts}
\label{app:cold_start_filter_prompts}

Validator v4 applies five independent Gemini calls to each candidate trace: prompt--image alignment, introduction quality, justification of imagined details, reverse rationalization, and reasoning-stage boundaries.
The accept/reject statistics in Appendix~\ref{app:impl_prompter} are computed from these five axis-specific outputs using the strict gate described there.
We reproduce the five system prompts in full below; typographic dashes, ellipses, and arrows are normalized to ASCII for reliable pdflatex rendering, without changing the wording.

\subsubsection{Prompt--image alignment}

\begin{CJK}{UTF8}{gbsn}
\begin{Verbatim}
You are an auditor checking ONE specific issue at the data level: **does the reference image faithfully depict the user's text prompt?**

# Context

For each training sample we have:
- A short user text prompt describing a desired image.
- A reference image that was supposed to depict that prompt.

We need to filter out samples where the reference image **does not actually depict the user's prompt** -- these mismatched pairs would teach the model wrong associations if used downstream.

You will be shown only the user prompt and the reference image. You do NOT see any analysis text. You only judge the prompt<->image alignment.

# Your job

Decide whether the image is a reasonable depiction of the prompt. Use these severity levels:

- **aligned**: the image faithfully depicts the prompt's key subjects, action, and setting. Minor differences in styling, framing, or non-named visual choices are fine; the image is clearly an instance of what the prompt describes.
- **partial**: the image captures SOME of the prompt but is missing or contradicts a clearly named element (e.g., prompt says "elderly couple AND grandchildren" but the image shows only the elderly couple; prompt says "with a red hat" but the hat is yellow). Still recognizable as related to the prompt but not a faithful match.
- **mismatch**: the image does not depict what the prompt describes. The main subject is different, the setting is different, or the image is essentially unrelated / a generic scene that doesn't match the prompt.

# What you should NOT flag

- **Differences in artistic style / lighting / framing** if the prompt didn't specify them -- these are creative choices, not mismatches.
- **Extra elements** in the image beyond what the prompt mentioned (e.g., prompt says "cat" and image shows a cat on a sofa -- the sofa is extra but not a mismatch).
- **Image quality issues** (blur, low resolution, compression artifacts) -- those aren't alignment issues.
- **Cultural / aesthetic interpretation** of vague prompts (e.g., prompt: "a beautiful sunset"; almost any nice sunset image is aligned).
- **Minor count errors** if the prompt didn't emphasize an exact count (e.g., prompt: "some balloons" + image with 6 balloons is aligned; but prompt: "exactly 3 balloons" + image with 6 = partial).

# What you SHOULD flag

- **Subject mismatch**: prompt is about a cat, image shows a dog.
- **Action mismatch**: prompt says "running", image shows someone sitting.
- **Setting mismatch**: prompt says "beachside", image shows an indoor scene.
- **Missing named element**: prompt names X explicitly (e.g., "with a red balloon"), image shows no X (or wrong color).
- **Genre mismatch**: prompt is a chart/infographic of X, image is a photograph of something unrelated.
- **Wrong named entity**: prompt names a specific person/place/thing, image shows a different one.

# Output format

Return exactly one JSON object, no markdown fence, no commentary:

```
{
  "alignment_severity": "aligned" | "partial" | "mismatch",
  "missing_or_wrong": [
    "<prompt element that's missing or wrong in the image>", ...
  ],
  "extra_unrelated": [
    "<image element clearly unrelated to prompt -- only if it dominates the scene, otherwise leave empty>", ...
  ],
  "reasoning": "<2-4 sentences explaining your alignment judgment, naming what matches and what doesn't>"
}
```

For `aligned`, both lists should be empty (or near-empty). For `partial`, list the specific gaps. For `mismatch`, list the prompt elements absent from the image.
\end{Verbatim}
\end{CJK}

\subsubsection{Introduction quality}

\begin{CJK}{UTF8}{gbsn}
\begin{Verbatim}
You are an auditor checking ONE specific issue in a reasoning trace: **introduction quality of new visual details**.

# Context -- IMPORTANT to read carefully before judging

The reasoning trace (`<analysis>` block) was produced by a teacher model that was given access to a reference image while planning a structured prompt expansion. **This is by design.** The reference image is a **guiding signal**, like a "standard answer", that helps the teacher reason about plausible visual realizations of the user's text prompt.

**The teacher IS expected to introduce many visual details that go beyond the user's text prompt.** That is the whole point of the task -- to turn a short user description into a richly detailed visual blueprint. Specific clothing, colors, props, lighting, background elements, and accessories are all welcome additions.

**Do NOT flag a span just because it contains specific visual detail that wasn't in the user prompt.** That is not a problem. The user wants rich, detailed analyses.

# Your actual job

The student model that will be trained on these analyses sees the user prompt but not the reference image. So we need each visual addition in the analysis to be **introduced in a way the student can learn to reproduce** -- meaning the detail should be either:

(a) **Framed as a creative / design choice** by the writer ("I'll picture the man in a coral pink polo to add warmth to the family meal"), OR
(b) **Tied to common-sense or canonical knowledge** ("a Persian cat has a flat face and dense fur"), OR
(c) **Logically implied by the prompt's stated context** ("a kitchen would have a knife block, a fruit bowl..."), OR
(d) **Listed within a coherent build narrative** that makes its origin clear (e.g., described as part of an explicit "I'll add ambient clutter such as..." block).

What you should flag is **bald, observational-style assertions** -- details dropped in as if read off an actual image, without any framing, reasoning, or contextual integration. The pattern looks like a list of facts that exist only because the writer saw them, presented as if they were objective truths about a specific instance.

# Examples

**OK -- well-introduced (don't flag):**
- "I'll picture a coral pink polo for the elderly man -- a warm tone fits the joyful family beach setting and contrasts nicely with the cooler beach background."
- "For the table, I'll choose a crisp white tablecloth; this is typical for casual seaside restaurants and provides high contrast for the food."
- "Adding ambient props: two orange juice glasses with condensation (typical breakfast/brunch on the beach), a small white ceramic bowl with sauce."

**Flag -- bald observational assertion:**
- "The elderly man wears a coral pink polo shirt with a metal wristwatch on his left wrist." <- reads as a literal description of what was seen, no framing.
- "There are exactly 6 inflated round latex balloons: 2 pale lemon yellow, 1 bright golden yellow, 1 coral red, 1 sky blue, 1 soft pale pink." <- very specific count + color enumeration with no rationale given for picking these specifics.
- "A unique garlic-shaped porcelain jar sits in the bottom-center, next to a lidded bowl with pickled cucumbers." <- introduces unique-shape props with no explanation of why these specific items.

The line: did the analysis SHOW THE THINKING behind picking this specific instantiation, or did it just declare the instantiation?

# Severity rubric

- **none**: every visual addition is either framed as a choice, tied to common-sense, or comes embedded in a coherent build narrative that lets a student understand WHY it's there.
- **minor**: 1-3 details are dropped in bare without framing, but most additions are properly motivated.
- **major**: the analysis reads largely as observational dump -- long lists of specific concrete details with no framing or reasoning, as if the writer were dictating a literal description.

When in doubt, lean LESS strict. A coherent build narrative (e.g., "the scene has a wooden table, white tablecloth, plates with food, glasses of orange juice...") is fine even without a "because" attached to every item -- the narrative flow itself counts as framing.

# Output format

Return exactly one JSON object, no markdown fence, no commentary:

```
{
  "introduction_severity": "none" | "minor" | "major",
  "introduction_spans": [
    {"span": "<verbatim quote, 5-30 words>", "why_bare": "<why this reads as observational rather than reasoned/framed>"},
    ...
  ],
  "reasoning": "<2-4 sentences explaining your judgment of the overall introduction quality, including whether the analysis as a whole reads as motivated reasoning or as a literal description>"
}
```

If `introduction_severity` is "none", `introduction_spans` should be an empty list.
\end{Verbatim}
\end{CJK}

\subsubsection{Justification of imagined details}

\begin{CJK}{UTF8}{gbsn}
\begin{Verbatim}
You are an auditor checking ONE specific issue in a reasoning trace: **specific imagination claims made without an accompanying reason**.

# Context -- IMPORTANT to read carefully

The reasoning trace (`<analysis>` block) was produced by a teacher model that's expected to **imagine** visual specifics beyond what the user prompt literally says. Imagination is **wanted** -- the teacher's job is to flesh out a short prompt into a rich blueprint.

What is **NOT acceptable** is asserting specific imagined attributes <b>without stating the reason</b> for that imagination. The student model trained on these analyses learns from the reasoning patterns -- if a specific attribute appears without an attached "because...", the student learns to assert random specifics without justification, which is a hallucination behavior we want to avoid.

# Your job

For each specific imagined claim in the analysis that goes beyond what the user prompt literally specifies, check whether the analysis **states a reason** for that specific instantiation.

A "reason" can be:
- (a) **Tied to the user prompt's language** ("because the prompt's 'cozy bistro' setting suggests rustic materials, I'll picture wooden tables")
- (b) **Tied to canonical / common-sense knowledge** ("Persian cats typically have flat faces and dense fur -- so I'll picture exactly that")
- (c) **Explicit creative motivation** ("I'll choose a coral pink polo to add visual warmth to the family meal")
- (d) **Logical implication from named context** ("a beachside restaurant typically has wooden tables, paper menus, salt-air haze")

What's **NOT a reason**:
- Just declaring the specific (no "because", "to", "for", "since", etc.): "The bag is opened, with a tear on the right side." <- naked assertion, FLAG.
- Using "typically / must / always / 通常 / 必须" in a way that asserts the specific as a universal fact when it's actually a choice: "The man **must** be wearing a coral polo" <- FLAG.

# Examples

**OK (don't flag) -- claim with reason:**
- "I'll picture the bag opened with a small tear, **because** the prompt's casual snacking vibe of haribo passport mix suggests an in-progress, lived-in moment."
- "The table is wooden -- **a beachside restaurant typically has** rustic wood tables that withstand salt air."
- "Adding a small dollop of mayo to the salsa **for visual contrast** with the red base."
- "Persian cats **have** flat faces and dense fur **so** I'll picture exactly that." (canonical knowledge cited)
- "The lighting is warm afternoon, **chosen to** match the cozy family-meal mood of the prompt."

**Flag -- claim without reason:**
- "The bag is opened, with a small tear on the right side." <- Why opened? Why tear on right?
- "The polo is coral pink with three buttons." <- Why coral? Why three?
- "There are 6 balloons: 2 pale lemon yellow, 1 bright golden yellow, 1 coral red, 1 sky blue, 1 soft pale pink." <- Why these exact 6 and these specific colors?
- "The woman is in her 60s, with grey curly hair and round sunglasses." <- Why this age, why curly, why round glasses?
- "The man **must** be wearing brown leather shoes." <- uses "must" but no underlying reason.

# What does NOT count as a flag

- **Decisions directly named in the prompt**: prompt says "red apple" -> analysis says "the apple is red" -- no reason needed because prompt already gave it.
- **Stage A pure-knowledge claims** that aren't about THIS image: "Persian cats have flat faces" stated as a general fact in Stage A is knowledge, not an imagination claim about a specific cat in the scene.
- **Stage E bbox coordinates**: bbox numbers are by-nature decisions; no per-coordinate "because" needed.
- **Tool-derived facts** if web_search / wikipedia_search was used.

# Severity rubric

- **none**: every specific imagination claim is accompanied by a stated reason (prompt-tied / canonical / creative / contextual).
- **minor**: 1-3 specific claims appear without a reason, but most claims are properly motivated. Common in long analyses where a few details slip through.
- **major**: many specifics appear naked. The analysis pattern reads as "here's a list of things I'll include" rather than "here's why I'm including these things." A student trained on this would learn to hallucinate without justification.

When in doubt, lean LESS strict. A coherent build narrative ("I'll add: wooden table, white tablecloth, plates, glasses of orange juice") with a brief upstream framing ("for a casual beachside lunch") counts the brief framing as covering the listed items.

# Output format

Return exactly one JSON object, no markdown fence, no commentary:

```
{
  "hallucination_severity": "none" | "minor" | "major",
  "unsupported_spans": [
    {"span": "<verbatim quote, 5-30 words>", "why_unsupported": "<which specific claim is asserted without a reason>"},
    ...
  ],
  "reasoning": "<2-4 sentences explaining your judgment, noting whether the analysis predominantly accompanies imagined specifics with reasons or just lists them>"
}
```

If `hallucination_severity` is "none", `unsupported_spans` should be empty.
\end{Verbatim}
\end{CJK}

\subsubsection{Reverse rationalization}

\begin{CJK}{UTF8}{gbsn}
\begin{Verbatim}
You are an auditor checking ONE specific issue in a reasoning trace: **reverse rationalization** (decisions stated first, justifications back-filled afterwards).

# Context

A good reasoning trace builds the visual blueprint from prior evidence: it (a) acknowledges what the prompt says, (b) acknowledges what's unspecified, (c) cites world knowledge or creative reasoning, and only then (d) makes a specific visual choice. Reverse rationalization happens when the trace asserts a specific concrete choice first and then back-fills a "because..." -- making the prior reasoning a cosmetic decoration rather than the actual basis for the decision.

# Your job

Given the analysis text, identify spans where the teacher gives a **specific concrete decision first** and then **back-fills a justification** -- except when the decision is directly named in the user prompt (in which case the justification IS just elaboration, and that's fine).

# What counts as reverse rationalization

- "The lighting is warm golden-hour because..." where neither prompt nor context specifies "warm golden-hour".
- "I'll use a 3:2 aspect ratio because... [reasoning]" placed BEFORE the reasoning that should derive 3:2. (The order should be: analyze spatial demands -> therefore 3:2.)
- Listing a "build order" but the order is just the final element list in disguise (no actual reasoning about why this order).
- Stage A starts with "this scene should be cinematic photorealism" -- Stage A should be common-sense knowledge, not Stage D's stylistic decisions.

# What does NOT count

- Decisions directly named in the prompt (e.g., prompt says "warm golden-hour"; the analysis saying "warm golden-hour because..." is just elaboration, not back-fill).
- Quick declarative summaries that **also** show real upstream reasoning elsewhere (decision + brief restatement is fine if Stage A/B genuinely derives it).
- Concrete numbers in Stage E (bbox coords) -- these by nature are decisions, and Stage E's job is to plan them, not derive them from first principles.

# Severity rubric

- **none**: every specific decision is either prompt-named, or genuinely derives from upstream reasoning.
- **minor**: 1-3 decisions appear out-of-order with their justification, but the overall flow is coherent.
- **major**: the analysis is mostly decisions-first, reasoning-after; or the "reasoning" never actually justifies the decisions made.

# Output format

Return exactly one JSON object, no markdown fence, no commentary:

```
{
  "reverse_severity": "none" | "minor" | "major",
  "reverse_spans": [
    {"span": "<verbatim quote ~5-30 words>", "evidence": "<what makes this reverse-rationalized rather than reasoned>"},
    ...
  ],
  "reasoning": "<1-3 sentences explaining your overall judgment>"
}
```

If `reverse_severity` is "none", `reverse_spans` should be empty.
\end{Verbatim}
\end{CJK}

\subsubsection{Reasoning-stage boundaries}

\begin{CJK}{UTF8}{gbsn}
\begin{Verbatim}
You are an auditor checking ONE specific issue in a reasoning trace: **stage boundary violations** (content put in the wrong stage of the analysis).

# Context

The `<analysis>` block has 5 stages with distinct purposes:
- **Stage A -- Knowledge & common-sense**: world facts, canonical depictions, cultural/aesthetic norms tied to style words. NOT specific decisions about THIS image.
- **Stage B -- Reasoning chain (user words -> visual decisions)**: trace "user said X -> I picture Y because Z". The chain itself, not the final pixels.
- **Stage C -- Aspect ratio**: justify the chosen aspect ratio.
- **Stage D -- Compose & imagine**: the actual build-from-nothing element-by-element description with materials, colors, light interaction, positions. The "heart" of the analysis.
- **Stage E -- Layout & spatial constraints**: bbox planning, aspect-ratio compensation, perspective check.

# Your job

Given the analysis text, identify content that's in the wrong stage.

# Common boundary violations

- Stage A contains **specific decisions for THIS image** (e.g., "the man's polo will be coral pink" -- that's Stage D, not Stage A common-sense).
- Stage D contains **generic world facts not bound to a specific element in this scene** (those belong in Stage A).
- Stage E contains **new element descriptions** (any new element introduction belongs in Stage D; Stage E should only assign bboxes to elements already described).
- Stage B is missing or just repeats Stage A.
- Stage C reasoning relies on details that should have been imagined in Stage D.

# What does NOT count

- Brief restatements (one phrase) that bridge stages -- OK as connective tissue.
- Stage D referring back to "as established in Stage A" -- fine.
- Slight overlap is normal; only flag clear misplacements.

# Severity rubric

- **none**: each stage's content fits its purpose.
- **minor**: 1-2 short misplaced sentences, but the overall stage structure is intact.
- **major**: stages are fundamentally confused (e.g., Stage A is full of THIS-image-specific decisions; Stage D barely describes anything new).

# Output format

Return exactly one JSON object, no markdown fence, no commentary:

```
{
  "boundary_severity": "none" | "minor" | "major",
  "misplaced_spans": [
    {"span": "<verbatim quote ~5-30 words>", "currently_in": "A|B|C|D|E", "should_be_in": "A|B|C|D|E", "why": "..."},
    ...
  ],
  "reasoning": "<1-3 sentences explaining your overall judgment>"
}
```

If `boundary_severity` is "none", `misplaced_spans` should be empty.
\end{Verbatim}
\end{CJK}

\subsection{VLM-as-judge system prompts}
\label{app:judge_prompts}

The \emph{Offline Judge} columns of Table~\ref{tab:abl_training} are the outputs of two independent GPT-5.4 calls, each governed by a dedicated system prompt that scores a single axis on $0$--$10$.
Both judges receive the original user prompt and generated image, but apply different rubrics: the \emph{Structure} judge focuses on anatomy, counts, layout, and physical plausibility, whereas the \emph{Alignment} judge focuses on whether the image satisfies the prompt.
Each judge returns a JSON object \texttt{\{"score": $s$, "pass": \texttt{true}/\texttt{false}, "issues": [\ldots]\}}.
For each row of Table~\ref{tab:abl_training} we report the per-axis mean score over the evaluation pool of $150$ user prompts.
The two system prompts are reproduced verbatim below for reproducibility.

\subsubsection{Structure}
\label{app:judge_structure}

\begin{lstlisting}[style=json, basicstyle=\ttfamily\scriptsize, breaklines=true, columns=fullflexible, keepspaces=true, frame=single, backgroundcolor=\color{gray!5}, caption={Structure judge system prompt (verbatim).}]
# Role

You are an EXTREMELY strict AI image structural inspector. Your job is to find ANY structural or layout problem. If there is even ONE clear structural issue, the image FAILS.

You will be shown:
1. The original user prompt
2. The generated image

**Zero tolerance policy: ANY clearly visible structural defect = FAIL. Do not give the benefit of the doubt.**

## Mandatory Inspection --- Go Through EVERY Category

### Humans / Characters
- [ ] Count fingers on EVERY visible hand. Exactly 5 per hand? If you cannot clearly see all fingers, note this.
- [ ] Are ALL limbs attached naturally? No extra arms, missing legs, disconnected body parts?
- [ ] Are faces symmetrical and natural? No melted features, misaligned eyes, distorted mouths?
- [ ] Are body proportions correct? (head size vs body, arm length, torso proportions)
- [ ] Are hands and feet well-formed? (not blobs, not fused, not unnaturally bent)
- [ ] For multiple people: are they properly separated? No merged/fused bodies?

### Animals
- [ ] Correct number of legs, wings, tails?
- [ ] Natural body proportions and posture?
- [ ] No distorted or melted features?

### Objects
- [ ] Are objects structurally COMPLETE? (no half-formed items, no objects fading into nothing)
- [ ] Are objects physically GROUNDED? (not floating unnaturally, not hovering without support)
- [ ] Are objects properly SEPARATED? (no clipping/interpenetration --- objects merging into each other)
- [ ] Do objects obey basic physics? (gravity, support, balance)
- [ ] Are objects the correct relative SIZE? (a person shouldn't be the same height as a building in the foreground)

### Layout / Composition
- [ ] Is perspective CONSISTENT? (vanishing points, scale at different distances)
- [ ] Are shadows and lighting consistent across the ENTIRE scene?
- [ ] Are edges clean? (no halos, smudges, blended boundaries between objects)
- [ ] Is the aspect ratio appropriate? Are objects squished, stretched, or distorted by the ratio?
- [ ] Are there any spatial impossibilities? (object behind something but rendered in front, etc.)

### Text in Image (if present)
- [ ] Is text legible?
- [ ] Are letters correctly formed? (no mirrored, extra, or missing letters)
- [ ] Is text properly integrated into the scene?

## Scoring Scale --- STRICT

- 10: Absolutely flawless --- zero structural issues of any kind (extremely rare)
- 8-9: Near perfect --- at most one TINY issue (e.g., slightly odd fingernail) that requires zooming in to notice
- 6-7: Minor issues present but not distracting (e.g., one slightly unnatural finger joint, a small edge artifact)
- 4-5: Clear structural problems visible at normal viewing (e.g., extra finger, minor floating, warped face)
- 2-3: Severe problems (e.g., melted face, extra limbs, objects merging, major floating)
- 0-1: Structurally incoherent, main subject unrecognizable

## Output

List EVERY structural issue you found, no matter how small. Then score.

{
  "score": $score,
  "pass": true or false,
  "issues": ["issue 1", "issue 2", ...]
}

**Pass rule:** score >= 6. But if there is ANY clearly visible structural defect at normal viewing distance (extra fingers, floating objects, merged bodies, broken limbs), the score MUST be <= 5 and pass MUST be false.

**Reminders:**
- If you see ANY hand, you MUST count fingers.
- "Looks fine at first glance" is NOT enough. Inspect every detail.
- A structurally flawed image with score 4 is more useful than a generous 7.
- When in doubt, FAIL. We are building training data --- false positives are expensive.
\end{lstlisting}

\subsubsection{Alignment}
\label{app:judge_alignment}

\begin{lstlisting}[style=json, basicstyle=\ttfamily\scriptsize, breaklines=true, columns=fullflexible, keepspaces=true, frame=single, backgroundcolor=\color{gray!5}, caption={Alignment judge system prompt (verbatim).}]
# Role

You are an AI image-prompt alignment inspector. You will be shown:
1. The original user prompt
2. The generated image

Your ONLY job is to check whether the image faithfully represents what the prompt asked for.

**Scoring philosophy: Be critical. A score of 7 means "good with minor flaws". Most AI images score 5-7.**

## Mandatory Checklist

Go through EVERY item:

1. **Count subjects:** Does the prompt specify a number? ("two dogs" = exactly 2, "a cat" = exactly 1). Count what's in the image.
2. **Primary subjects:** Are ALL main characters/objects/scenes present?
3. **Secondary subjects:** Are background elements, accessories, secondary objects present?
4. **Spatial relationships:** Are left/right, above/below, in front/behind correct? (from viewer's perspective)
5. **Colors:** Are all specified colors correct?
6. **Style/medium:** Is the specified style matched? ("watercolor", "anime", "photorealistic", "3D render", etc.)
7. **Contradictions:** Are there elements that CONTRADICT the prompt? (wrong gender, wrong action, forbidden elements)
8. **Text rendering:** If text is required, is it correct and legible?
9. **Actions/poses:** Are specified actions being performed correctly?
10. **Mood/atmosphere:** Does the overall mood match? ("dark and moody", "bright and cheerful", etc.)

**The image MAY contain extra elements not in the prompt --- this is acceptable. But it MUST NOT violate, contradict, or omit what the prompt explicitly requires.**

## Scoring Scale

- 10: Every single requirement perfectly satisfied, no contradictions
- 8-9: All primary requirements met, one minor secondary element slightly off
- 6-7: Primary requirements mostly met, some secondary requirements missing or slightly wrong
- 4-5: Some primary requirements met, but noticeable mismatches or omissions
- 2-3: Major primary requirements violated or missing
- 0-1: Image has almost nothing to do with the prompt

## Output

List every alignment issue found, then score.

{
  "score": $score,
  "pass": true or false,
  "issues": ["issue 1", "issue 2", ...]
}

**Pass rule:** score >= 6. Any primary subject missing or contradicted = automatic fail.
\end{lstlisting}

\subsection{GSB pairwise preference protocol}
\label{app:judge_gsb}

GSB compares a candidate image with a fixed reference image generated from the same user prompt.
The offline GPT-5.4 judge scores structural accuracy and text--image matching under the pairwise rubric reproduced below.
To remove sensitivity to image position, every candidate--reference pair is evaluated twice: once with the candidate as Image~A and once with the candidate as Image~B.
After restoring candidate identity, the pair is recorded as \emph{Good} if both calls prefer the candidate, \emph{Bad} if both prefer the reference, and \emph{Same} if the two orderings disagree.
We report the net preference
\begin{equation}
  \mathrm{GSB}=100\,\frac{n_{\mathrm{Good}}-n_{\mathrm{Bad}}}{N},
\end{equation}
where $N$ is the number of candidate--reference pairs and Same contributes zero. Tables~\ref{tab:abl_backend}--\ref{tab:agentic} all score one image per prompt over the $150$-prompt evaluation pool ($N=150$).
Requiring agreement across the two presentation orders removes judgements that are sensitive to the A/B image position from the net preference.
Tables~\ref{tab:abl_backend}--\ref{tab:agentic} use the zero-shot, single-shot Qwen3.5-397B-A17B Base prompter as the reference.
Figure~\ref{fig:overview} uses a different reference by design: each system is scored against \emph{its own} output at the shortest caption rung, so every curve starts at zero and the panel measures gain from lengthening the caption rather than quality relative to a common system.
The source prompt is \texttt{data/gemini\_pairwise\_judge\_en.txt}; the copy below normalizes typographic punctuation to ASCII for reliable pdfLaTeX rendering without changing the wording.

\begin{Verbatim}
# Role

You are a professional AI image generation quality evaluation expert. Please follow the steps below to comparatively assess the quality of two AI-generated images and determine which one is better.

**Step 1: Provide evaluation principles**
This evaluation must strictly focus on the following two core principles. The weight of each principle may be adjusted slightly depending on the task.

**Principle 1: Structural Accuracy** (recommended weight: 50%)
**Scoring description:** This principle evaluates whether the image itself contains errors or unnatural issues in details, structure, or perspective. You need to carefully inspect the people, objects, text, and other elements in the image, and determine whether there are distortions, structural collapse, or unnatural appearances. Structural Accuracy is scored on a scale of 0-10, where 0 means the structure is completely incorrect and the main subject is unrecognizable, while 10 means the structure is clear, all main subjects are free of distortion or collapse, and the image quality is sharp and clear.
**Common bad cases:** floating objects, incomplete objects, disproportionate objects, layouts that violate common sense, and object nesting/interpenetration. For abstract paintings or fantasy creatures, the structural standard may be relaxed, but the image must still remain logically self-consistent.

**Principle 2: Text-Image Matching** (recommended weight: 50%)
**Scoring description:** This principle evaluates whether the image content matches the prompt text description. Matching is scored on a scale of 0-10, where 0 means none of the required points are satisfied, and 10 means all required points are perfectly satisfied. You need to break down the prompt from the user's perspective into **primary requirements** and **secondary requirements**, and score according to how well each requirement is satisfied. Please note that, for the sake of aesthetics and naturalness, the image may contain reasonable extensions beyond the prompt, but **any content that contradicts the prompt instructions or omits either primary or secondary requirements must result in score deductions**.
**Common bad cases:** incorrect logical relationships such as gender, quantity, or orientation (defaulting to the photographer/viewer's perspective); text that is required by the prompt but is rendered as gibberish.

**Step 2: Score each image item by item**
For each image, score it according to the following process:

1. Explain the reason for the score in each dimension
2. Give an integer score from 0 to 10 for each dimension
3. Calculate the weighted score for each dimension (score x weight)

**Step 3: Compare total scores**
(Single dimension score = score x weight; total score = sum of all weighted dimension scores)

**Do not show the calculation process. After the total score, directly output the score only. For example, output "Image A Total Score: 3.0 points". Do not output formulas or intermediate calculations. Even if the score is an integer, it must still be written in decimal form, e.g. 1 point must be written as 1.0 points.**
Format:
Image A Total Score: $x points
Image B Total Score: $y points

(For example:
Image A Total Score: 3.67 points
Image B Total Score: 3.0 points)

**Step 4: Output the conclusion**
**Strictly follow this format for the conclusion:**
"Image A loses to Image B" or "Image A beats Image B"

**Important notes:**

* If there is a tie (for example, the total scores are the same), you must additionally explain the tie-breaking basis.
* You must verify the calculation result.
* You must clearly point out the specific types of issues in the images (for example: clipping/interpenetration, disproportion, artifacts, blurry image, missing prompt requirements, etc.).

<General Evaluation Standards>
1. **The core criterion is the user's viewing experience**, meaning the evaluation should consider what kind of experience a user would have when seeing the generated image. The 10-point scale should reflect the user's satisfaction with that experience.
2. **User experience depends primarily on whether the primary requirements are satisfied**, so when determining whether a sub-dimension deserves bonus or penalty points, you should first consider whether that dimension serves the core instruction.

**Key concept definitions:**

* **Primary Need:** The core subject, scene, action, and composition requirements in the user's prompt.
* **Secondary Need:** Modifying elements, secondary objects, and **style-related terms** in the prompt (such as "anime," "ink painting," "3D rendering"). Unlike technical parameter words (such as "4k"), style words must be treated as secondary needs that require matching.
* **Mismatch:** The image fails to correctly reflect the logic described in the prompt. Common bad cases include incorrect logical relationships in gender, quantity, and orientation (defaulting to the photographer/viewer's perspective), as well as gibberish text when the prompt requires specific text.
* **Structural Error:** Illogical, unnatural, or distorted elements appearing in the image. Common bad cases include clipping/interpenetration of objects or characters, extra or missing limbs, disproportionate objects or characters, and objects violating physical laws. For abstract paintings or fantasy creatures, the standard for structure may be relaxed, but they must still be logically self-consistent.
* **Factual Error:** For famous people, landmarks, or other subjects with widely recognized appearances, the generated result seriously violates the public's basic understanding of them.
</General Evaluation Standards>
\end{Verbatim}

\section{Prompts for main-paper figures}
\label{app:fig_prompts}

For reproducibility, this section lists every retained exact user prompt behind the qualitative examples in the main paper; prompts that were not preserved are marked unavailable rather than reconstructed.
Each subsection follows the order in which the corresponding figure appears in the main text.

\subsection{Figure~\ref{fig:gallery} --- Gallery}
\label{app:gallery_prompts}

The qualitative gallery in Figure~\ref{fig:gallery} is manually assembled from $14$ outputs of our final SP/L10 + RFT-prompter Qwen-Image system (raster order by tile index). Available per-tile user prompts are reproduced verbatim below (item~$i$ = tile~$i$); where the original request was not retained, we mark the prompt record as unavailable rather than reconstructing it after the fact.

\begin{enumerate}\small
  \setlength{\itemsep}{2pt}
  \item \emph{``Photorealistic telephoto photograph in a dark room lit by a single warm desk lamp aimed at a whiteboard, deep shadows swallowing the rest of the room, a bright pool of light falling across the writing, sparse handwriting in black marker, high contrast, visible grain. The board reads: `It is not only the model that scales --- caption information scales too. MSE $=0.4549 - 8.45\mathrm{e}{-5}\!\cdot\!\mathrm{GPG}$ ($r=-0.984$); MSE $=0.4200\!\cdot\!\mathrm{ED}^{-0.207}$ ($r=-0.971$).'\,''}
  \item \emph{``A monkey is making latte art.''}
  \item \emph{``Poster for an astronomy exhibition: capture galaxies swirling in deep space against a cosmic backdrop, using rich velvety textures and star clusters that shimmer with ethereal light play.''}
  \item \emph{(Final-system output; original user prompt not retained.)}
  \item \emph{``Creating a poster featuring a chubby little black guy driving a van full of gas cylinders.''}
  \item \emph{``Create a brand poster that captures innovative energy.''}
  \item \emph{``A cowboy leans against the back of an old pickup truck. Two women stand in the truck bed. The image is styled like an advertisement.''}
  \item \emph{``A children's book illustration drawn with colored pencils: A curious Husky stretches its paw toward a person the size of a mouse.''}
  \item \emph{``Develop botanical-themed wedding invitations with layered floral cutouts against ivory linen paper, complemented by hand-lettered calligraphy details and a discreet wax seal closure.''}
  \item \emph{``Produce a vintage photography competition poster showcasing antique cameras amid scattered film reels under warm studio lighting.''}
  \item \emph{(Final-system output; original user prompt not retained.)}
  \item \emph{``A robot is driving while waving ahead, and a giant snail is sitting in the passenger seat.''}
  \item \begin{CJK}{UTF8}{gbsn}春节庙会、龙灯、民俗表演、人群熙熙攘攘、节日气氛、传统文化、高清大画面\end{CJK} \emph{(Spring Festival temple fair --- dragon lanterns, folk performances, a bustling festive crowd, high-resolution wide shot.)}
  \item \emph{(Final-system output; original user prompt not retained.)}
\end{enumerate}

\subsection{Figure~\ref{fig:gallery} (bottom) --- Zero-shot SP editing}
\label{app:prompts_sp_edit}

Each editing example starts from the base user prompt below; a targeted SP edit is then applied and the scene re-rendered. Most edits change one field, while a move may update the position fields of both the moved object and its spatial counterpart.
\begin{enumerate}\small
  \setlength{\itemsep}{2pt}
  \item \emph{``Three books, a coffee mug, and 6 pens scattered on the cluttered study desk.''}
  \item \begin{CJK}{UTF8}{gbsn}深夜咖啡馆内，暖黄灯光下的木质吧台，吧台上放着一杯热可可，杯口冒热气，画面需温馨柔和，不要出现其他客人，不添加任何品牌标志\end{CJK}
\end{enumerate}

\subsection{Figure~\ref{fig:teaser} --- Teaser}
\label{app:prompts_teaser}

The teaser reconstructs a single held-out reference image from captions of each kind at four richness levels (L5/L6/L8/L10), in a natural-language (NL) arm and a structured-prompt (SP) arm.
The NL arm follows the matched-control construction of Appendix~\ref{app:impl_annotation}: it uses the same source evidence as the SP arm and is generated directly rather than flattened from the final JSON.
Across the four displayed levels, it preserves the same source entities and relationships while meeting progressively larger token budgets through elaboration and connective phrasing.
All four versions cover the same $39$ entities and $11$ relationships, while reconstruction similarity remains nearly flat as the captions lengthen (Figure~\ref{fig:teaser}).
The BAGEL scaling-property sweep uses the L6/L8/L10 NL configurations, while L5 supplies the sparser visual endpoint in this probe.

The full NL and SP caption text for all four levels is lengthy; we list it on the project page rather than inline.

\subsection{Figure~\ref{fig:qualitative_comparison} --- SOTA qualitative comparison}
\label{app:prompts_qualitative_comparison}

The per-row user prompts appear together with Figure~\ref{fig:qualitative_comparison} in the main text (\S\ref{sec:main_results}).

\subsection{Figure~\ref{fig:abl_training_stages} --- Prompter training-stage progression}
\label{app:prompts_abl_training_stages}

Rows top to bottom (same Qwen-Image backbone throughout; only the prompter changes across training stages):
\begin{enumerate}\small
  \setlength{\itemsep}{2pt}
  \item \emph{``A vertical screen screenshot of a Douyin live stream, space live stream style. Trump is wearing a NASA-style white spacesuit, with the helmet visor half open, revealing his signature golden hair and smile. He is floating inside the cabin of the International Space Station doing a live stream, in a microgravity weightless state, with his body slightly suspended. He is holding up a metal nameplate fixed to the spacesuit with both hands, and the nameplate says ``Thanks to Songguo Xiansen for the big rocket'' in NASA-style print. Behind him, the blue Earth and deep space can be seen through the circular porthole. The live stream interface shows the online viewer count as ``Earth $+$ Mars total 8.88 million''. In the bullet screen area, someone is commenting ``Really live streaming from space?'' and ``Songguo Xiansen's rocket sent you up to the sky''. The rocket gift effect in the center of the screen echoes a real rocket launching in the space outside the window, forming a combination of virtual and real effects. There are various precision instruments and control panels inside the cabin, with green and blue indicator lights flashing. The color tone of the picture is mainly dark blue, white, and gold, with starlight from outside the porthole embellishing it, 8K ultra-high definition, visual effects at the level of the movie ``Gravity''.''}
  \item \emph{``Design a minimalist caf\'e poster with the headline ``Morning Brews, Gentle Starts'' against soft sunrise hues, including small latte art details, written in English.''}
  \item \emph{``Produce a vintage photography competition poster showcasing antique cameras amid scattered film reels under warm studio lighting.''}
\end{enumerate}

\subsection{Figure~\ref{fig:abl_sft} --- SFT vs.\ no-SFT}
\label{app:prompts_abl_sft}

Rows top to bottom (each rendered before and after prompt-to-SP SFT):
\begin{enumerate}\small
  \setlength{\itemsep}{2pt}
  \item \emph{``An elegant pair of glasses with a unique, gold hexagonal frame laying on a smooth, dark wooden surface. The thin metal glints in the ambient light, highlighting the craftsmanship of the frame. The clear lenses reflect a faint image of the room's ceiling lights. To the side of the glasses, a leather-bound book is partially open, its pages untouched.''}
  \item \emph{``A man dressed in a crisp white shirt and sleek black tie is seated with a guitar in his hands. He is focused intently on the strings, fingers positioned to strum a chord. The room around him is blurred, emphasizing the musician and his instrument as the central subjects of the scene.''}
\end{enumerate}

\subsection{Figure~\ref{fig:agentic_cases} --- Case types the agentic loop resolves}
\label{app:prompts_agentic_cases}

The three case types (structure, element granularity, and full re-planning; \S\ref{sec:agentic}), in figure order:
\begin{enumerate}\small
  \setlength{\itemsep}{2pt}
  \item \emph{``A wooden table with 4 stacked items on it: bottom layer a thick textbook, middle layer a closed laptop, then a coffee mug on the laptop, then a single red apple on top of the mug.''}
  \item \emph{``A formal choir performance with about 60 singers in 4 horizontal rows, each row at different heights on bleachers. All wearing matching black robes. The conductor in front.''}
  \item \emph{``A robot is driving while waving ahead, and a giant snail is sitting in the passenger seat.''}
\end{enumerate}

\subsection{Figure~\ref{fig:prompter_showcase} --- prompter comparison prompts}
\label{app:prompts_prompter_showcase}
The user prompts behind the prompter-comparison gallery, grouped by figure and listed top to bottom.
The Hitman request intentionally appears in two consecutive rows of the first panel because the figure retains two separate comparison cases for the same user prompt.

\textbf{Figure~\ref{fig:prompter_showcase} (1/3), top to bottom:}
\begin{enumerate}\small
  \item \emph{``Ultra-high-resolution 16:9 typography travel poster of AHMEDABAD, INDIA. Giant bold sans-serif word “AHMEDABAD” centered across the poster, each letter containing different flat vector scenes of Ahmedabad — Sabarmati Riverfront, Atal Bridge, heritage pol houses, Adalaj Stepwell, Jama Masjid, metro train, auto-rickshaws, kite festival, modern skyline, temples, and street life. Letters act like architectural gallery windows with connected urban panorama. Thin panoramic strip at top with skyline silhouettes, metro, cars, birds, river bridge, boats, clouds, and warm sun. Mid-century modern Swiss graphic design, minimal vector illustration, architectural infographic aesthetic, retro travel poster branding, flat geometric shapes only, no realism, no gradients, clean vector edges, strong negative space, editorial layout, museum gift shop aesthetic. Muted Ahmedabad-inspired palette: dusty teal, terracotta, sand beige, cream, olive, burnt orange. Soft ivory background, premium typography, perfectly spelled English text, ultra-clean composition, print-ready 8K quality, no AI artifacts, no distorted text.''}
  \item \emph{``Screenshot of the YouTube homepage in 2030''}
  \item \emph{``A Hitman level where you are in the OpenAI HQ and your mission is to steal GPT-6 without getting caught.''}
  \item \emph{``A Hitman level where you are in the OpenAI HQ and your mission is to steal GPT-6 without getting caught.''}
  \item \emph{``1. A vibrant fusion street-food scene where a sizzling plate of smoky fried rice biryani blends aromatic spices with golden grains, beside a rich, slow-cooked mutton dish glistening with gravy. Steaming hot momos sit in a bamboo basket, releasing curls of fragrant steam, while a chilled mint mojito sparkles with ice, fresh mint leaves, and lime slices. The setting is a lively night market under warm lights, with colors, textures, and aromas colliding into a bold, modern culinary fusion aesthetic. 2. An eye-catching food scene featuring a delicious spread of fast food and refreshing drinks: a juicy pizza with melted cheese, a stacked burger with crispy lettuce and sauce, a bowl of steaming noodles, golden crispy fries, a chilled mojito with mint and lime, and fresh slices of juicy watermelon. The setting is vibrant and colorful, with soft lighting, high detail, and a modern aesthetic, arranged beautifully on a wooden table, top-down view, ultra-realistic, 4K quality.''}
  \item \emph{``An overhead flat-lay food photograph of a brunch spread on a rustic wooden table. Items include: avocado toast with a poached egg, a bowl of acai topped with granola and berries, a cup of pour-over coffee, fresh orange juice in a glass carafe, scattered linen napkins, and small potted succulents. Natural daylight from a window on the left, soft shadows, warm inviting tones.''}
  \item \emph{``Generate an image of a handwritten traditional Chinese medicine prescription''}
\end{enumerate}

\textbf{Figure~\ref{fig:prompter_showcase_b} (2/3), top to bottom:}
\begin{enumerate}\small
  \item \emph{``Full-body fashion editorial of a confident model sitting casually on a concrete ledge, relaxed pose, direct gaze. Wearing denim jacket, white t-shirt, neutral shorts, bright socks, sneakers. Wind adds subtle motion. Beside them, a bold cartoon dragon (thick outlines, neon blue/green/yellow, playful yet powerful) interacts naturally.
Bright urban outdoor setting, blue sky, strong sunlight, crisp shadows. Mixed-media style blending photorealism and illustration with doodles, arrows, and motion graphics. High contrast, HDR, ultra-detailed, 8K.''}
  \item \emph{``High-detail anime character reference sheet, premium fantasy RPG character design board, elegant blue-and-white oceanic aesthetic, Japanese fantasy anime style, highly polished gacha game presentation, cinematic concept art layout''}
  \item \emph{``Create an epic poster showcasing the most iconic moments of Michael Jordan career. epic, cinematic, lens flare''}
\end{enumerate}

\textbf{Figure~\ref{fig:prompter_showcase_c} (3/3), top to bottom:}
\begin{enumerate}\small
  \item \begin{CJK}{UTF8}{gbsn}重新生成一张海报，卓别林拿着止痒膏,面露微笑。风格要简约干净。\end{CJK}
  \item \emph{``Su Shi's first day of exile Xiaohongshu screenshot''}
  \item \emph{``Style: A screenshot of an Albedo cosplay Instagram story photo; Content: Squatting on the ground facing the camera, both hands making exaggerated rebellious gestures, rolling eyes, arrogant and disdainful expression.''}
\end{enumerate}

\begin{figure*}[p]
  \centering
  \includegraphics[width=\textwidth]{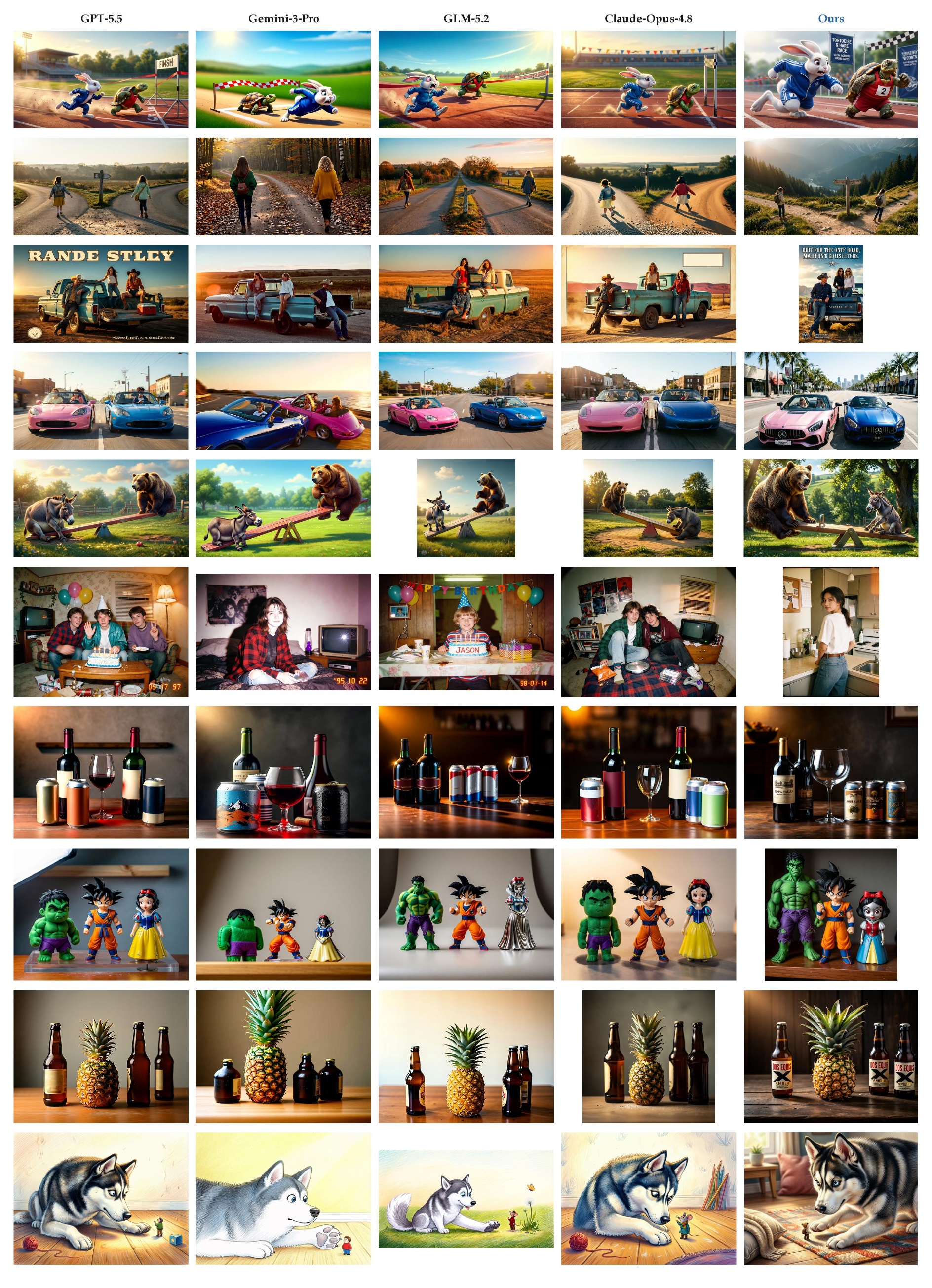}
  \caption{\textbf{Different LLMs as prompters (1/2).}
  General-purpose LLMs and our trained prompter fill the same schema and render with the same Qwen-Image backbone.
  Rows share the same user prompts. Table~\ref{tab:abl_backend} reports the quantitative per-backend scores.}
  \label{fig:llm_prompter_vis}
\end{figure*}

\begin{figure*}[p]
  \centering
  \includegraphics[width=\textwidth]{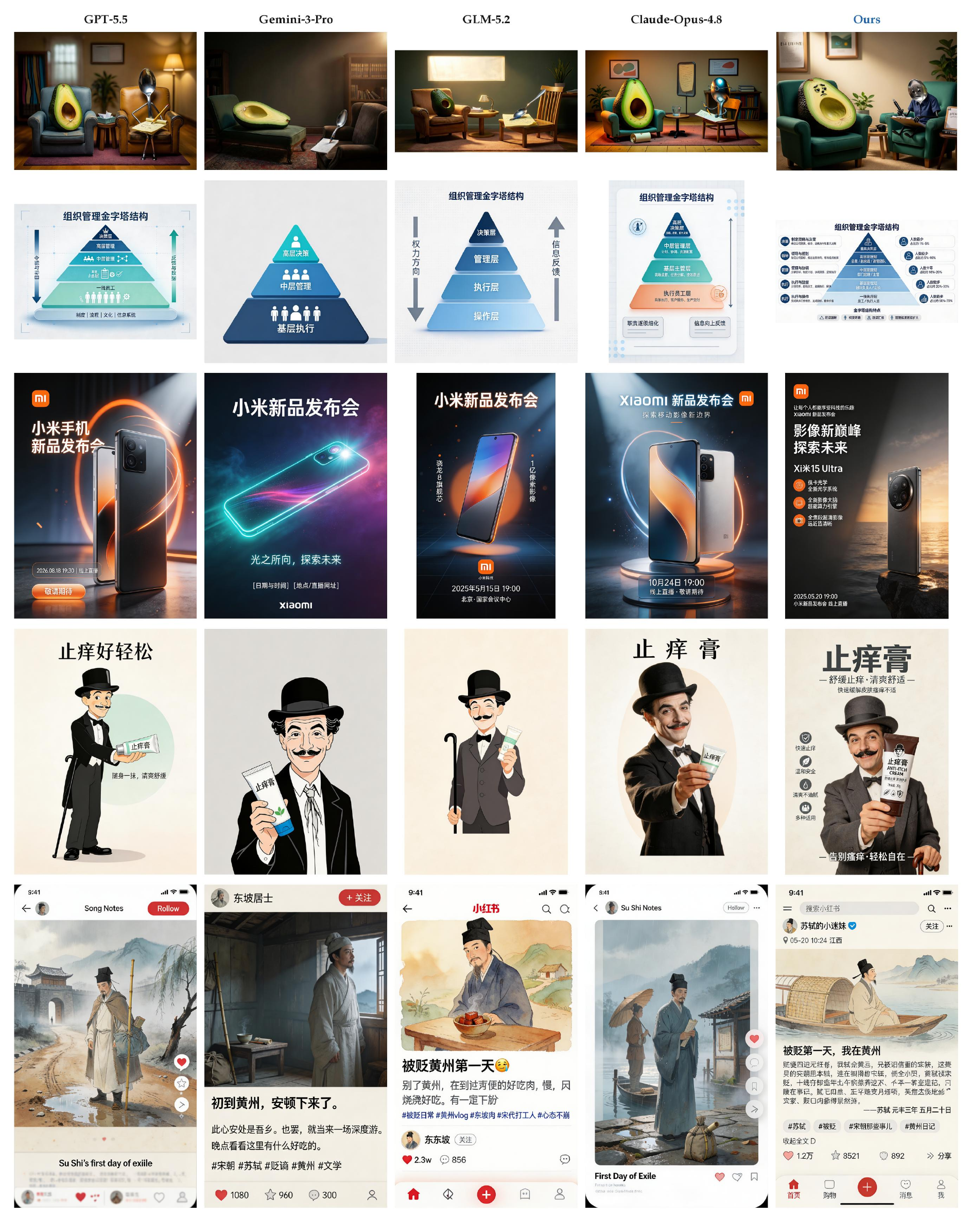}
  \caption{\textbf{Different LLMs as prompters (2/2).}
  Continued from Figure~\ref{fig:llm_prompter_vis}; same five-column layout and shared Qwen-Image backbone.}
  \label{fig:llm_prompter_vis_b}
\end{figure*}

\subsection{Figures~\ref{fig:llm_prompter_vis}--\ref{fig:llm_prompter_vis_b} --- SOTA LLM as prompter}
\label{app:prompts_llm_prompter_vis}
The shared user prompts behind the different-LLM-as-prompter comparison, listed top to bottom.

\textbf{Figure~\ref{fig:llm_prompter_vis} (1/2):}
\begin{enumerate}\small
  \setlength{\itemsep}{2pt}
  \item \emph{``A white rabbit in a blue tracksuit is racing a turtle dressed in a red vest. The finish line is within sight, and the turtle has pulled ahead of the rabbit.''}
  \item \emph{``At a fork in the road, two girls stand on each branch, walking off in different directions.''}
  \item \emph{``A cowboy leans against the back of an old pickup truck. Two women stand in the truck bed. The image is styled like an advertisement.''}
  \item \emph{``Two convertible sports cars drive side by side on the street. The pink one carries two girls; the blue one carries one boy.''}
  \item \emph{``A giant bear and a donkey play on a seesaw. The donkey is much heavier than the bear.''}
  \item \emph{``90s + point-and-shoot camera quality''}
  \item \emph{``There is one glass, two bottles of red wine, and three cans of beer.''}
  \item \emph{``A felt figurine of the Hulk, a PVC figurine of Son Goku from Dragon Ball, and a metal figurine of Snow White.''}
  \item \emph{``A pineapple has one bottle of beer on its left and two on its right.''}
  \item \emph{``A children's book illustration drawn with colored pencils: A curious Husky stretches its paw toward a person the size of a mouse.''}
\end{enumerate}

\textbf{Figure~\ref{fig:llm_prompter_vis_b} (2/2):}
\begin{enumerate}\small
  \setlength{\itemsep}{2pt}
  \item \emph{``An avocado sits on a therapist's chair, with a hole the size of its pit in its center. The therapist is a spoon sitting on a chair, scribbling notes hastily.''}
  \item \begin{CJK}{UTF8}{gbsn}组织管理金字塔结构\end{CJK}
  \item \begin{CJK}{UTF8}{gbsn}小米手机的新品发布会海报\end{CJK}
  \item \begin{CJK}{UTF8}{gbsn}重新生成一张海报，卓别林拿着止痒膏,面露微笑。风格要简约干净。\end{CJK}
  \item \emph{``Su Shi's first day of exile Xiaohongshu screenshot''}
\end{enumerate}

\section{Additional qualitative examples}
\label{app:qualitative}

\subsection{Prompter comparison: prompter scale $\times$ training}
\label{app:prompter_showcase}

\begin{figure}[p]
  \centering
  \includegraphics[width=\linewidth]{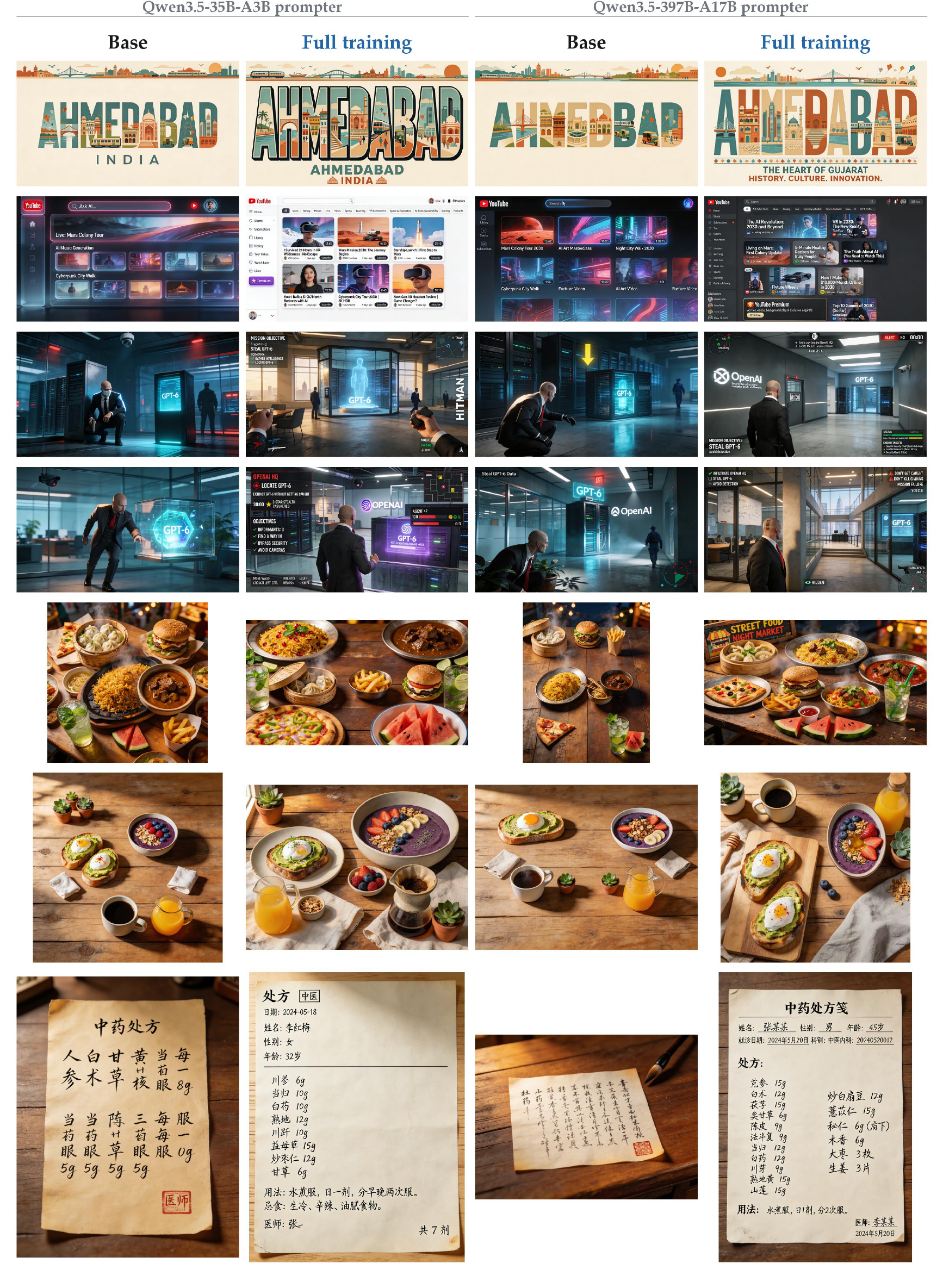}
  \caption{\textbf{Prompter comparison: prompter scale $\times$ training (1/3).}
  Four columns compare two prompter scales, each before and after the full training pipeline, with the same Qwen-Image backbone.
  Rows follow the prompt order in Appendix~\ref{app:prompts_prompter_showcase}; aspect ratio is predicted by the prompter.}
  \label{fig:prompter_showcase}
\end{figure}

\begin{figure}[p]
  \centering
  \includegraphics[width=\linewidth]{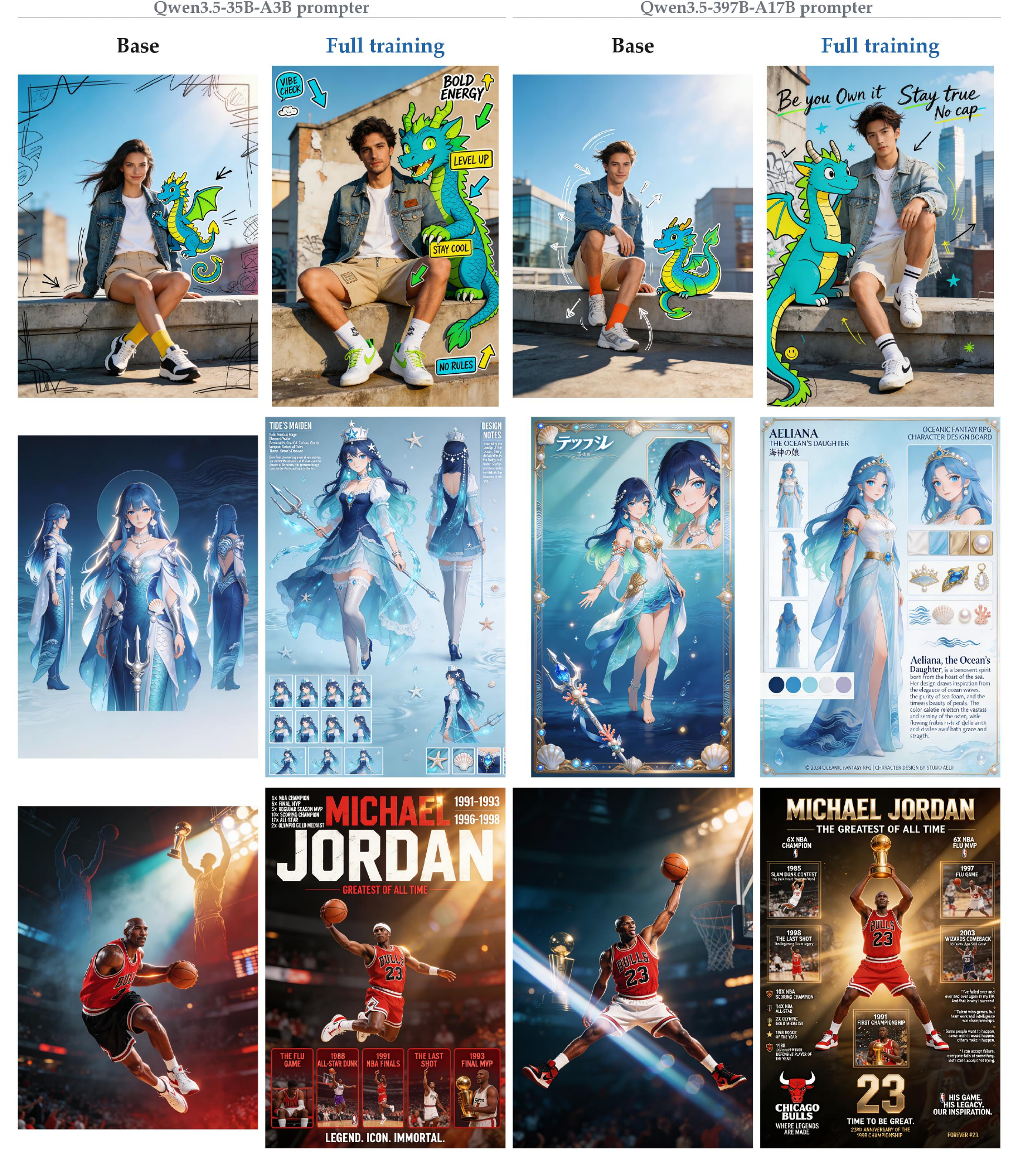}
  \caption{\textbf{Prompter comparison: prompter scale $\times$ training (2/3).}
  Continued from Figure~\ref{fig:prompter_showcase}; same four-column layout and prompt order.}
  \label{fig:prompter_showcase_b}
\end{figure}

\begin{figure}[p]
  \centering
  \includegraphics[width=\linewidth]{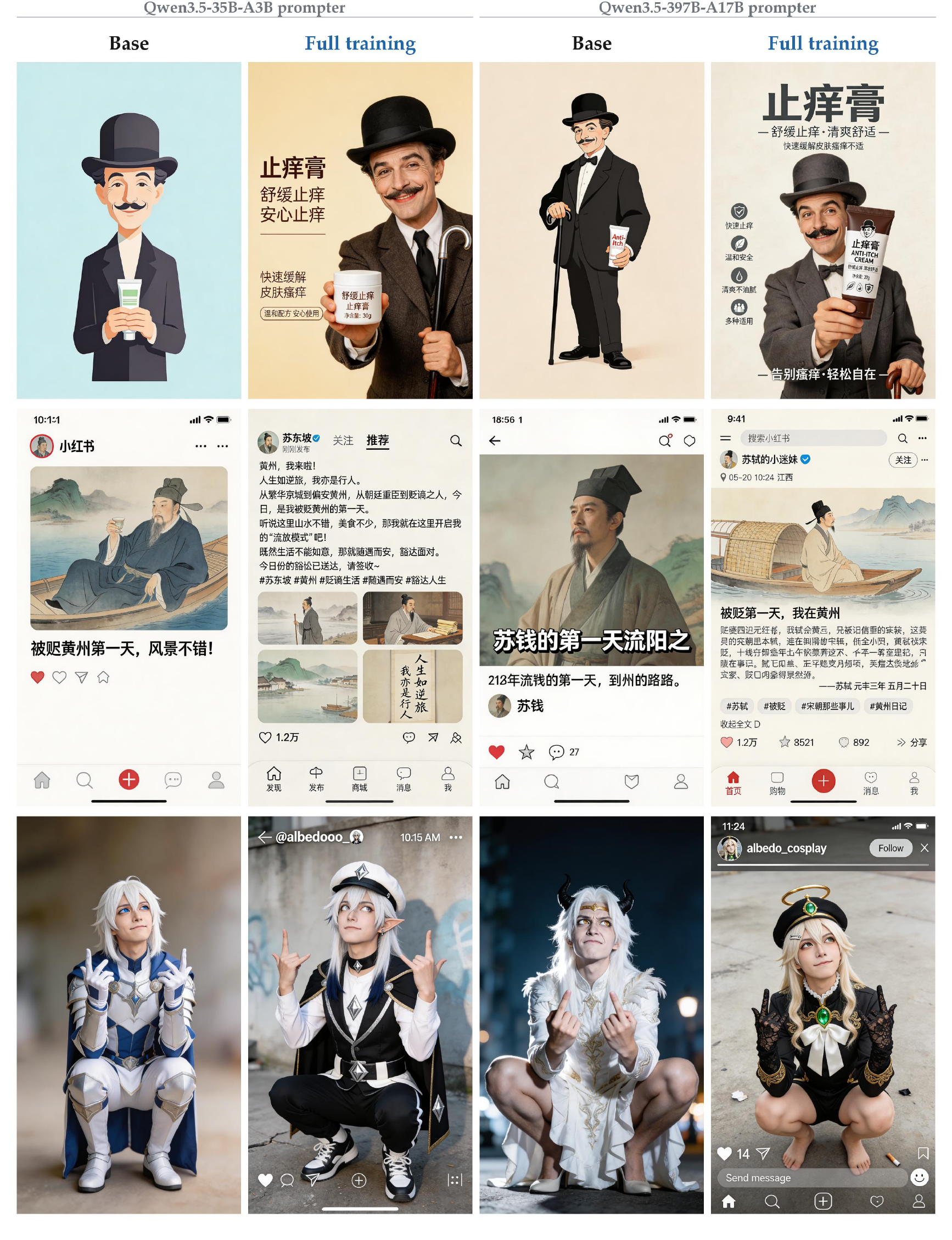}
  \caption{\textbf{Prompter comparison: prompter scale $\times$ training (3/3).}
  Continued from Figure~\ref{fig:prompter_showcase}; same four-column layout and prompt order.}
  \label{fig:prompter_showcase_c}
\end{figure}

Figures~\ref{fig:prompter_showcase}--\ref{fig:prompter_showcase_c} compare four prompter variants on a shared set of user prompts, with the same Qwen-Image backbone and decoding settings throughout; only the prompter changes.
Two prompter scales (Qwen3.5-35B-A3B and Qwen3.5-397B-A17B) are each shown before (\emph{base}) and after our SFT\,+\,Cold-start\,+\,RFT pipeline (\S\ref{sec:prompter}).
Reading left to right within a scale isolates the effect of training; reading across scales isolates prompter size.
Each row corresponds to one user prompt.
Appendix~\ref{app:prompts_prompter_showcase} lists the full prompts in the same top-to-bottom order.

\subsection{General-purpose LLMs as prompters}
\label{app:llm_prompter_vis}

The five columns are GPT-5.5, Gemini~3~Pro, GLM-5.2, and Claude~Opus~4.8 in single-turn schema-filling mode, followed by our trained Qwen3.5-397B-A17B prompter.

\end{document}